%% file: main.tex
\documentclass[acmsmall]{acmart}

\AtBeginDocument{%
  }

\setcopyright{acmlicensed}
\copyrightyear{2018}
\acmYear{2018}
\acmDOI{XXXXXXX.XXXXXXX}

\acmJournal{JACM}
\acmVolume{37}
\acmNumber{4}
\acmArticle{111}
\acmMonth{8}

\usepackage{multirow}
\usepackage{subfigure}
\usepackage{tablefootnote}
\usepackage{adjustbox}
\usepackage{appendix}
\usepackage{amsmath}
\usepackage{xspace}
\usepackage{bbding}
\usepackage{makecell}
\usepackage{ulem}
\begin{document}

\title{CodeApex: A Bilingual Programming Evaluation Benchmark for Large Language Models}
\newcommand \model{CodeApex\xspace}
\newcommand \web{https://}

\author{Lingyue Fu}
\author{Huacan Chai}
\author{Kounianhua Du}
\author{Weiming Zhang}
\author{Shuang Luo}
\author{Jianghao Lin}
\author{Yuchen Fang}
\author{Renting Rui}
\author{Hao Guan}
\author{Jianxing Liu}
\author{Siyuan Qi}
\author{Longteng Fan}
\author{Jiayi Lei}
\author{Yifan Liu}
\author{Jingkuan Wang}
\author{Kangning Zhang}
\author{Weinan Zhang}
\author{Yong Yu}

\affiliation{
\institution{ Apex Data \& Knowledge Management Lab, Shanghai Jiao Tong University}
  \state{Shanghai}
  \country{China}\\
  \texttt{https://apex.sjtu.edu.cn/codeapex/}
}
\authornote{The leading author is Lingyue Fu <fulingyue@sjtu.edu.cn>. Shuang Luo has graduated from SJTU and is now at KTH. Correspondence to Weinan Zhang <wnzhang@sjtu.edu.cn> and Yong Yu <yyu@sjtu.edu.cn>.}
\begin{abstract}
   With the emergence of Large Language Models (LLMs), there has been a significant improvement in the programming capabilities of models, attracting growing attention from researchers.
   Evaluating the programming capabilities of LLMs is crucial as it reflects the multifaceted abilities of LLMs, and it has numerous downstream applications.
  In this paper, we propose \model, a bilingual benchmark dataset focusing on the programming comprehension, code generation, and code correction abilities of LLMs.
   Programming comprehension task tests LLMs on multiple-choice exam questions covering conceptual understanding, commonsense reasoning, and multi-hop reasoning.
   The code generation task evaluates LLMs through completing C++ functions based on provided descriptions and prototypes.
   The code correction task asks LLMs to fix real-world erroneous code segments with different error messages. 
   We evaluate 12 widely used LLMs, including both general-purpose and specialized models. 
 GPT-4 exhibits the best programming capabilities, achieving approximate accuracy of 69\%, 54\%, and 66\% on the three tasks, respectively.
  Compared to human performance, there is still significant room for improvement in LLM programming.
  We hope that \model can serve as a reference for evaluating the coding capabilities of LLMs, further promoting their development and growth.
 
\end{abstract}



\received{20 February 2007}
\received[revised]{12 March 2009}
\received[accepted]{5 June 2009}

\maketitle
\input{introduction}
\input{relatedWork}

\section{Evaluation Protocol}\label{methodology}
\begin{figure}[t]
    \centering
    \includegraphics[width=0.9\textwidth]{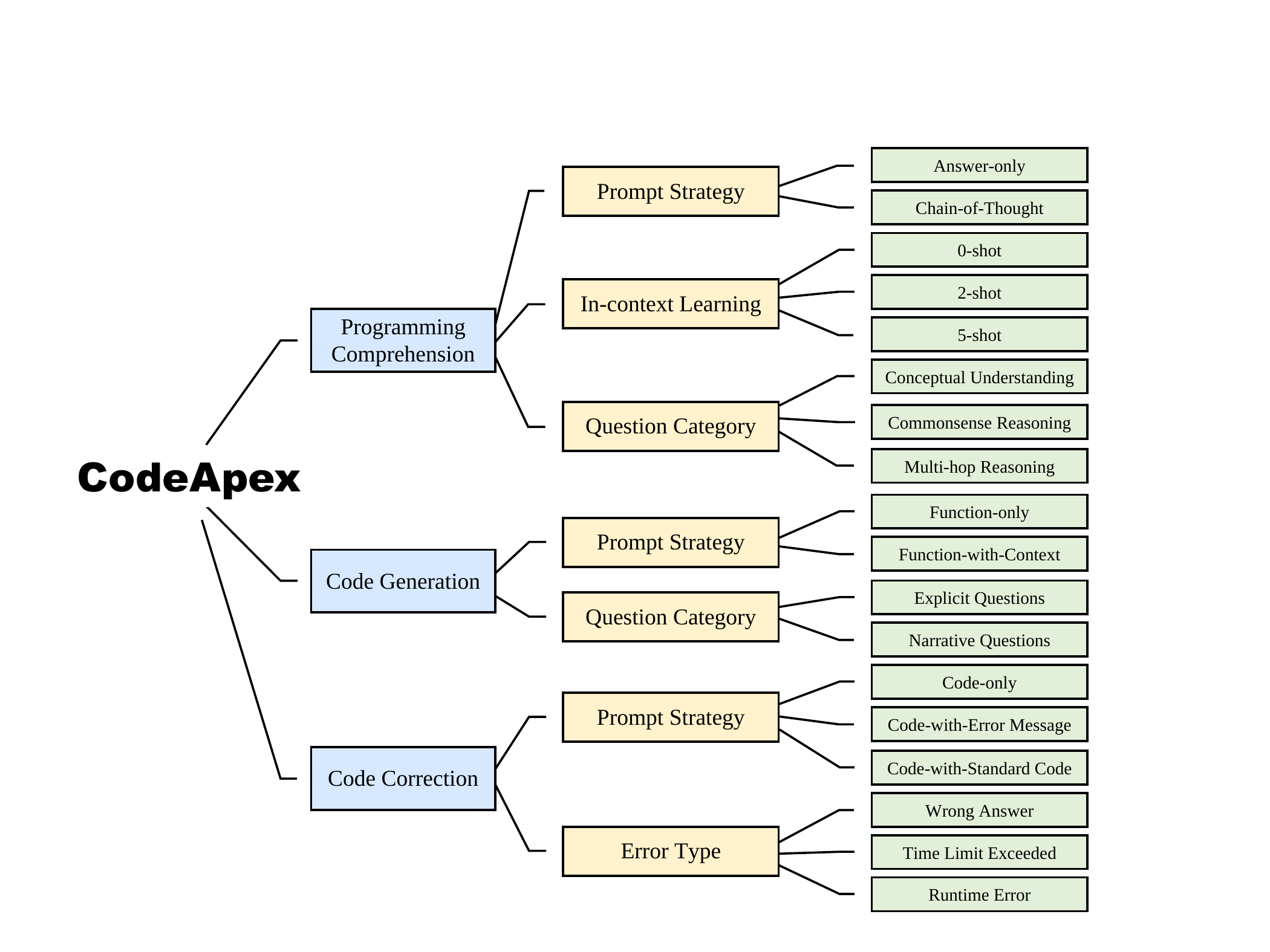}
    \caption{Overview diagram of \model benchmark.}
    \label{fig:intro}
\end{figure}

In this section, we introduce the evaluation framework of CodeApex for three code tasks: (1) programming comprehension, (2) code generation, and (3) code correction. 
The overall experiment scenarios of \model are listed in Figure \ref{fig:intro}.
\begin{itemize}
    \item The first task is \textit{programming comprehension}, whose test set includes $250$ multiple-choice exam questions, which are categorized into conceptual understanding, commonsense reasoning, and multi-hop reasoning questions.
The questions are selected from the final exams of different classes (Programming, Data Structure, Algorithm) at the university, which reduces the risk that the test data is already in the training corpus of LLMs.
LLMs are asked to choose the correct option under the 0-shot, 2-shot, and 5-shot in-context learning scenarios.
Due to the effective application of Chain-of-Thought (CoT), we also compare the performances of LLMs under the answer-only and CoT settings.
\item The second task is \textit{code generation}, whose test set includes $476$ C++-based coding questions, covering explicit questions and narrative questions.
The description of the question and the prototype of the function to implement the question are given, and LLMs are asked to complete the main part of the function.
We also provide both the function-only and function-with-context scenarios, which indicate whether the calling code of the objective function is given in addition to the description of the objective function (i.e., the context information of the code).
\item The third task is \textit{code correction}, whose test set includes 1330 erroneous code segments with three types of error messages (Wrong Answer, Time Limit Exceeded, and Runtime Error).
We introduce three types of prompts to simulate code-only, code-with-error message, and code-with-standard code scenarios.
\end{itemize}
For each task, we first provide an overview and then discuss about the data processing, prompting strategy, and evaluation metric.
To facilitate a fine-grained comparison of the differences in the abilities of LLMs across different natural languages, we provide \textit{aligned Chinese and English versions} for all test data in the three tasks.

\input{comprehension}

\input{generation}

\input{debug}
\input{experiemnt}

\input{conclusion}

\bibliographystyle{ACM-Reference-Format}
\bibliography{reference}

\section*{Acknowledgment}
We would like to thank Qinxiang Cao and Pengfei Liu for helpful discussions.
We also thank Shanghai Boyu Education Technology Co., Ltd for granting us data authorization and providing valuable support in establishing the testing framework.
The work is partially supported by National Natural Science Foundation of China (62177033). 
\setcounter{section}{0}
\setcounter{table}{0}  
\setcounter{figure}{0}
\renewcommand{\thesection}{\Alph{section}}

\clearpage
\appendix

\input{appendix}

\end{document}

%% file: introduction.tex
\section{Introduction}
Due to the widespread adaption of the Transformer~\cite{vaswani2017attention} architecture and advancements in computational power, Large Language Models (LLMs) have been widely employed in various tasks, including recommender systems~\citep{lin2023can,lin2023rella}, dialogue systems \citep{zhang2019dialogpt,adiwardana2020meena}, summarization \citep{zhang2020pegasus,liu2019bertsum}, sentiment analysis \citep{ghosal2020sail,basiri2021abcdm}, etc.
This trend highlights the need for robust evaluation frameworks to measure the effectiveness of LLMs in these diverse applications such as education~\cite{chen2023teaching,de2023can,hellas2023exploring}, social science~\cite{deroy2023ready, frank2023baby} and medical~\cite{lahat2023evaluating, article}.
Establishing these benchmarks can provide quantitatively thorough assessments for the capabilities of LLMs from various perspectives, thereby guiding model optimization, validating model performance, and analyzing model differences.  
Among these diverse benchmarks, we argue that the code and programming capability is one of the most crucial aspects for the evaluation of LLMs for the following two reasons.

\textit{Firstly}, the performance of LLMs on code and programming tasks largely reflects their inherent capabilities of logical reasoning, instruction following, and structural data understanding.
Code and programming serve as the bridge between humans and computers, and translate high-level human need into executable steps, logical consistency, featuring standard syntax, and functional modularity~\cite{yang2024if}.
Various works~\cite{yang2024if,nam2024using,ma2023training} have identified the importance of code data for model tuning to help unlock the reasoning ability of LLMs and steer LLMs to produce structured and precise intermediate steps.
As a result, various benchmarks and leaderboards~\cite{huang2023ceval} that focus on general reasoning ability evaluation of LLMs would include code-related tasks in addition to traditional tasks like arithmetic~\cite{arora2023llms, DBLP:journals/corr/abs-2110-14168} and symbolic reasoning~\cite{wei2022chain}, which draws a thorough assessment over the fundamental capabilities of large language models.

\textit{Secondly}, apart from the potential relationship with basic reasoning abilities of LLMs, code-related tasks also play crucial roles in a wide range of downstream applications, such as education~\cite{educhat2023} and software engineering~\cite{copilot}. 
A variety of code foundation models~\cite{copilot,zheng2023codegeex,alphacode,codet5} have been proposed for different programming tasks.
Copilot~\cite{copilot} achieves code auto-completion and auto-correction during the programming process, therefore greatly promoting the coding experience of human programmers.
CodeGeeX~\cite{zheng2023codegeex}, as an interactive programming assistant, can solve various programming problems, including code interpretation, code translation, code correction, document generation, etc.
Google DeepMind team proposes AlphaCode~\cite{alphacode}, which aims to solve programming problems that require deep reasoning abilities to find novel solutions. 
CodeT5~\cite{codet5} employs a unified framework to seamlessly support both code understanding and generation tasks, with the allowance of multi-task learning.
Such a emergence of code foundation models with different programming goals (e.g., code completion, code debugging) draws demand on the corresponding evaluation to ensure their performance in real-world production scenarios.

\begin{table}[t]
\centering
\caption{Comparison between commonly used programming benchmarks in the era of LLMs and \model. P.C. stands for programming comprehension. C.G. stands for code generation. C.C. stands for code correction. The number of plus symbols (i.e., $+$) in the \textit{Data Scale} column represents the order of magnitude of the test samples.}
\label{tab:comparation}
    \resizebox{0.9\textwidth}{!}{
\renewcommand{\arraystretch}{1.1}

\begin{tabular}{l|ccc|c|c|c|c}
\toprule\hline
\multirow{2}{*}{Benchmark} & \multicolumn{3}{c|}{Programming Task} & \multicolumn{1}{c|}{\multirow{2}{*}{\makecell[c]{Fine-grained \\Categorization} }} & \multirow{2}{*}{Data Scale} & \multicolumn{1}{c|}{\multirow{2}{*}{Bilingual}} & \multirow{2}{*}{Human Expert} \\ \cline{2-4}
                           & P.C.         & C.G.        & C.C.        & \multicolumn{1}{c|}{}                             & \multicolumn{1}{c|}{}                            & \multicolumn{1}{c|}{}                           &                             \\ \hline
MBPP~\cite{DBLP:journals/corr/abs-2108-07732} & \XSolidBrush & \Checkmark & \XSolidBrush & \XSolidBrush & +++ & \XSolidBrush &\XSolidBrush \\
CoderEval~\cite{10.1145/3597503.3623316} & \XSolidBrush &\Checkmark  & \XSolidBrush &\XSolidBrush & ++ & \XSolidBrush & \XSolidBrush \\
HumanEval~\citep{humaneval} & \XSolidBrush &\Checkmark  & \XSolidBrush &\XSolidBrush & +  & \XSolidBrush & \XSolidBrush \\
HumanEval-X~\citep{zheng2023codegeex} & \XSolidBrush &\Checkmark  & \XSolidBrush& \XSolidBrush & ++ &\XSolidBrush  & \XSolidBrush \\
 DebugBench~\cite{tian2024debugbench} & \XSolidBrush & \XSolidBrush & \Checkmark & \Checkmark & +++ & \XSolidBrush & \Checkmark \\\hline
 CodeApex (Ours) & \Checkmark & \Checkmark & \Checkmark & \Checkmark & +++ & \Checkmark & \Checkmark
\\\hline\bottomrule
\end{tabular}
}
\vspace{-10pt}
\end{table}

These factors discussed above yield an urgent need for a comprehensive benchmark to thoroughly evaluate the code and programming capabilities of large language models. 
In Table~\ref{tab:comparation}, we illustrate the important properties of the commonly used programming evaluation benchmarks~\cite{humaneval,zheng2023codegeex,tian2024debugbench, DBLP:journals/corr/abs-2108-07732, 10.1145/3597503.3623316}, and discuss their potential limitations as follows: 
(1) Existing benchmarks only focus on a single programming task for LLMs within uni-lingual test cases (i.e., English), which makes the evaluation incomprehensive. 
(2) Existing benchmarks, excluding DebugBench~\cite{tian2024debugbench}, generally lack fine-grained categorization over the test data and human-expert evaluation, which are crucial to derive deeper insights and analysis for different aspects of programming, as well as the thorough comparison between LLMs and human-level abilities.
To this end, we propose the \model benchmark, which gives a comprehensive code and programming evaluation of large language models. 
As shown in Table~\ref{tab:comparation}, CodeApex is a pioneering bilingual (English and Chinese) programming benchmark over three different code-related tasks (i.e., programming comprehension, code generation, and code correction) with fine-grained categorization, large test data scale, as well as human-expert evaluation.
We have comprehensively evaluated 12 different LLMs using \model, including both API-based and open-source models.
Our work analyzes the overall code capabilities of LLMs by comparing their performance across different tasks.
Fine-grained categorization experiments provide an analysis of LLMs across different strategies and data types.
Experimental results demonstrate the varying performance of different models in code-related tasks, with GPT models showcasing exceptional competitiveness and distinct advantages. 
Additionally, the experiment compares the performance of LLMs in bilingual and different prompt strategy scenarios.
We also organize human testing on code comprehension and code generation tasks, comparing performance between humans and LLMs.
Overall, within the leaderboard of the \model, there remains significant room for improvement in LLM accuracy, indicating an untapped potential for LLMs in code-related tasks.

The rest of the paper is organized as follows. In Section \ref{relatedwork}, we review previous work on evaluating the code-related capabilities of LLMs. 
We present the evaluation protocol for three programming tasks in Section \ref{methodology}.
We present and discuss the evaluation results across three tasks and multiple categorizations in Section \ref{exp}.
Finally, we conclude this paper and discuss future work in Section \ref{conclusion}.


%% file: relatedWork.tex
\section{Related Work}\label{relatedwork}
\subsection{Code Foundation Models}
The implementation of programming comprehension heavily relies on the alignment of code space and natural language space through encoding.
Graph2Code \citep{graph2code} employs graph neural networks to convert source code into a graph structure, capturing structural information within the code and thereby enhancing the accuracy of programming comprehension. 
Code2Vec \citep{code2vec} represents the Abstract Syntax Tree (AST) as token sequences along paths, enabling code transformation into fixed-length vector representations and facilitating the learning of code semantics.
The introduction of the Transformer architecture \citep{transformer} has provided novel approaches for code comprehension tasks.
One prominent model in this regard is CodeT5 \citep{codet5}, an extension of the Text-to-Text Transfer Transformer (T5) \citep{t5} specifically designed for natural language processing tasks on source code. 
CodeBERT \citep{codebert} maps both natural language and code into a shared vector space and leverages attention mechanisms to capture semantic relationships between them.
Furthermore, there are models \citep{cnntree, attentionxml} that focus on specific programming tasks such as code comment generation and API documentation generation within the source code context.

The task of code generation has garnered significant attention after the emergence of LLMs. 
These language models are pre-trained on massive text datasets, enabling them to learn rich language representations. 
General-purpose LLMs, such as GPT and Llama~\cite{llama}, have a certain ability to generate code.
Some LLMs are specifically designed training schemes for the programming tasks, aimed at improving their coding performance.
One common approach to code generation is fine-tuning existing large-scale language models. 
In this method, a pre-trained language model is used as the initial model and further trained on specific code generation datasets.
Codex~\cite{humaneval}, PaLMCoder\cite{palm}, CodeGeeX2\citep{zheng2023codegeex} are further trained on GPT-3, PaLM\cite{palm}, ChatGLM2\cite{zeng2022glm, du2022glm} with extensive public code datasets, demonstrating better comprehension and generation abilities on programming languages. 
With the development of instruction tuning techniques, a series of models, like WizardCoder\citep{luo2023wizardcoder}, Code Llama-Instruct\citep{codellama}, PanguCoder\citep{christopoulou2022pangu} demonstrate powerful capabilities in multilingual code generation and debugging tasks.
Another prevalent approach is prompt engineering, which involves designing suitable prompts or guiding statements to direct the model in completing specific code generation tasks. By providing insightful inputs, models can generate more accurate and expected code outputs. 
CodeGen \citep{nijkamp2022codegen} converts natural language questions into code generation tasks and improves generation effectiveness through appropriate prompt engineering techniques.
DocPrompting \citep{zhou2022docprompting} leverages keyword retrieval to integrate information from code repositories into prompt phrases, guiding the LLMs to enhance the performance of LLMs in code generation tasks.

\subsection{Programming Evaluation}
Evaluating large language models' (LLMs) capabilities can be effectively achieved through multi-choice formats, where models are tasked with selecting the correct answer from given options.
Multi-choice RACE transforms the original reading comprehension dataset RACE \citep{lai2017race} into a set of questions with answer choices, facilitating the evaluation of the model's score.
AI2 Reasoning Challenge (ARC) \citep{clark2018thinkarc} serves as a benchmark to measure how well language models can reason and draw conclusions within the domain of science.
C-Eval \citep{huang2023ceval} provides multiple-choice questions from multiple subjects, enabling us to evaluate the competency of LLMs in different academic domains.
\model employs multiple choice question formats to evaluate LLM's programming comprehension ability.

Another crucial dimension in assessing LLMs' capabilities involves their performance on code-related tasks. The development of specialized benchmark datasets offers nuanced insights into LLMs' programming comprehension and generation skills.
In the earlier days, FlashFill++ \citep{flashfill} provides a benchmark dataset for inferring string transformation programs from input-output examples. 
CodeSearchNet \citep{husain2020codesearchnet} serves code search tasks by offering a vast collection of code snippets for code retrieval and documentation generation.
With the development of deep learning, NLP tasks have become more diverse. 
Benchmark datasets have started combining code-related tasks with natural language. 
CodeXGLUE \citep{code_xglue} is a benchmark that encompasses natural language understanding of code, covering various tasks such as code summarization, code translation, code completion, and more.
APPS \citep{hendrycks2021measuring} is a benchmark designed to evaluate program synthesis systems, with a focus on probabilistic programs and their ability to generalize to new tasks.
With the emergence of LLMs, \cite{humaneval} proposed the HumanEval benchmark, specifically designed to assess the quality of generated code. 
HumanEval-X\cite{zheng2023codegeex} benchmark introduces support for multiple programming languages, including C++, Python, Go, and Java, aiming to facilitate research in cross-language code generation and understanding. 
CoderEval\cite{yu2024codereval} benchmark specifically focuses on evaluating large language models' abilities to tackle complex, industrial-level coding challenges through a series of practical programming tasks.
DebugBench~\cite{tian2024debugbench} implants bugs into standard code by GPT4 and evaluates five LLMs in a zero-shot scenario.
\model focuses on evaluating the reasoning capabilities of LLMs by presenting them with algorithmic challenges and associated test cases.

%% file: comprehension.tex
\subsection{Programming Comprehension Task}
\subsubsection{Overview}
In the programming comprehension task, LLMs have to answer multiple-choice questions that are closely related to code \& programming topics.
This evaluates the basic capabilities of LLMs in understanding the code from various dimensions such as syntax, execution flow and code structure.
The ability of programming comprehension lays the foundation for LLMs to accomplish other code-related tasks, and is therefore indispensable to thoroughly integrate LLMs into real-world code production scenarios. 

\subsubsection{Data Processing}
We evaluate the programming comprehension capability of LLMs via multiple-choice questions, which generally cover the key knowledge points of programming, data structure and algorithms. 
The multiple-choice question data is obtained from the final exams of real-world college courses under strict confidentiality, greatly decreasing the possibility of the test data overlapping with the training data of LLMs crawled online (i.e., test data leakage).
Questions with multiple correct answers are manually modified into single-choice questions to facilitate testing consistency.
Similar to C-Eval \citep{huang2023ceval}, we format the options of the multiple-choice questions for alignment. 
The question prompts are provided in Markdown format, i.e., code blocks are marked using triple quotes. 
All questions are manually cross-validated to ensure the unambiguity of the question descriptions and the correctness of the answers.

\begin{figure}[t]
    \centering
    \includegraphics[width=\textwidth]{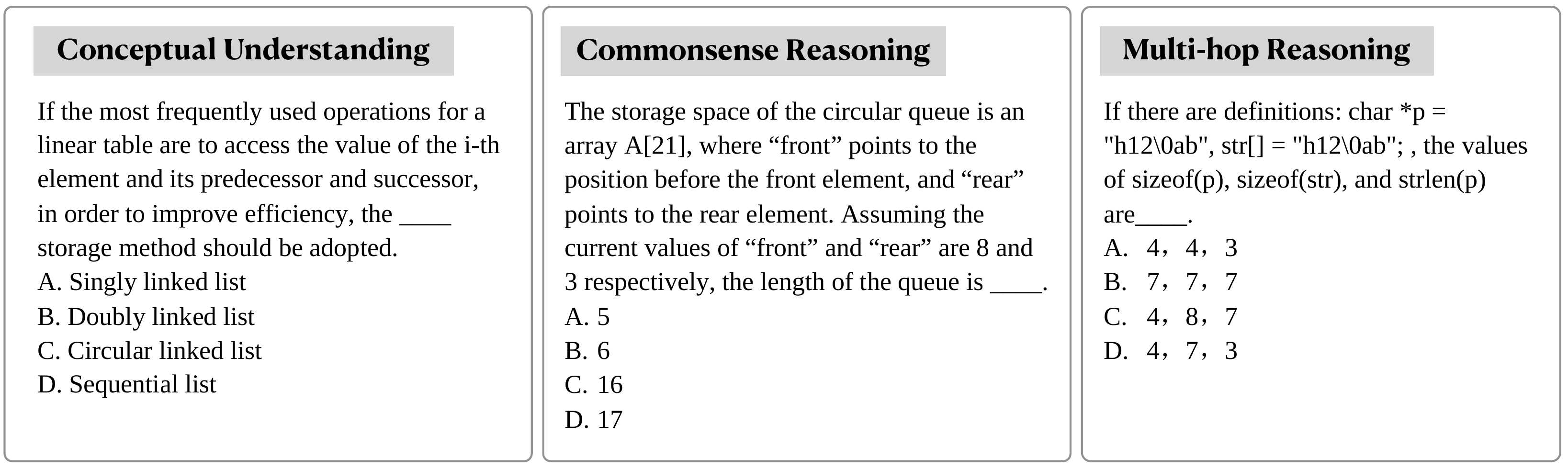}
    \caption{Examples of the three categories of questions in the programming comprehension task.}
    \label{fig:category}
\end{figure}

In order to facilitate further exploration of the fine-grained abilities of LLMs, we manually classify the questions into three categories: (1) conceptual understanding, (2) commonsense reasoning, and (3) multi-hop reasoning.
Example questions of three categories are demonstrated in Figure \ref{fig:category}.
Conceptual understanding questions assess LLMs' basic understanding of programming concepts, typically involving selecting the correct concept or natural language description without requiring specific code writing or comprehension. Commonsense Reasoning questions primarily evaluate LLMs' ability to read and understand code. 
These two types of questions focus on a single knowledge point, with relatively simple problem-solving steps that can be deduced through single-step reasoning.
Multi-hop reasoning questions evaluate LLMs' ability to reason about code. These questions often involve chains of thought and require multiple steps of inference to get the correct answer. 
These three categories of questions correspond to LLMs' elementary grasp of programming concepts, basic understanding of code, and advanced comprehension of code, respectively. 
The categorized results could provide us with deeper insights into the fine-grained levels of programming comprehension capabilities of LLMs.

\begin{figure}[t]
    \centering
    \subfigure[Answer-only setting (Chinese).]{\includegraphics[width=0.48\textwidth]{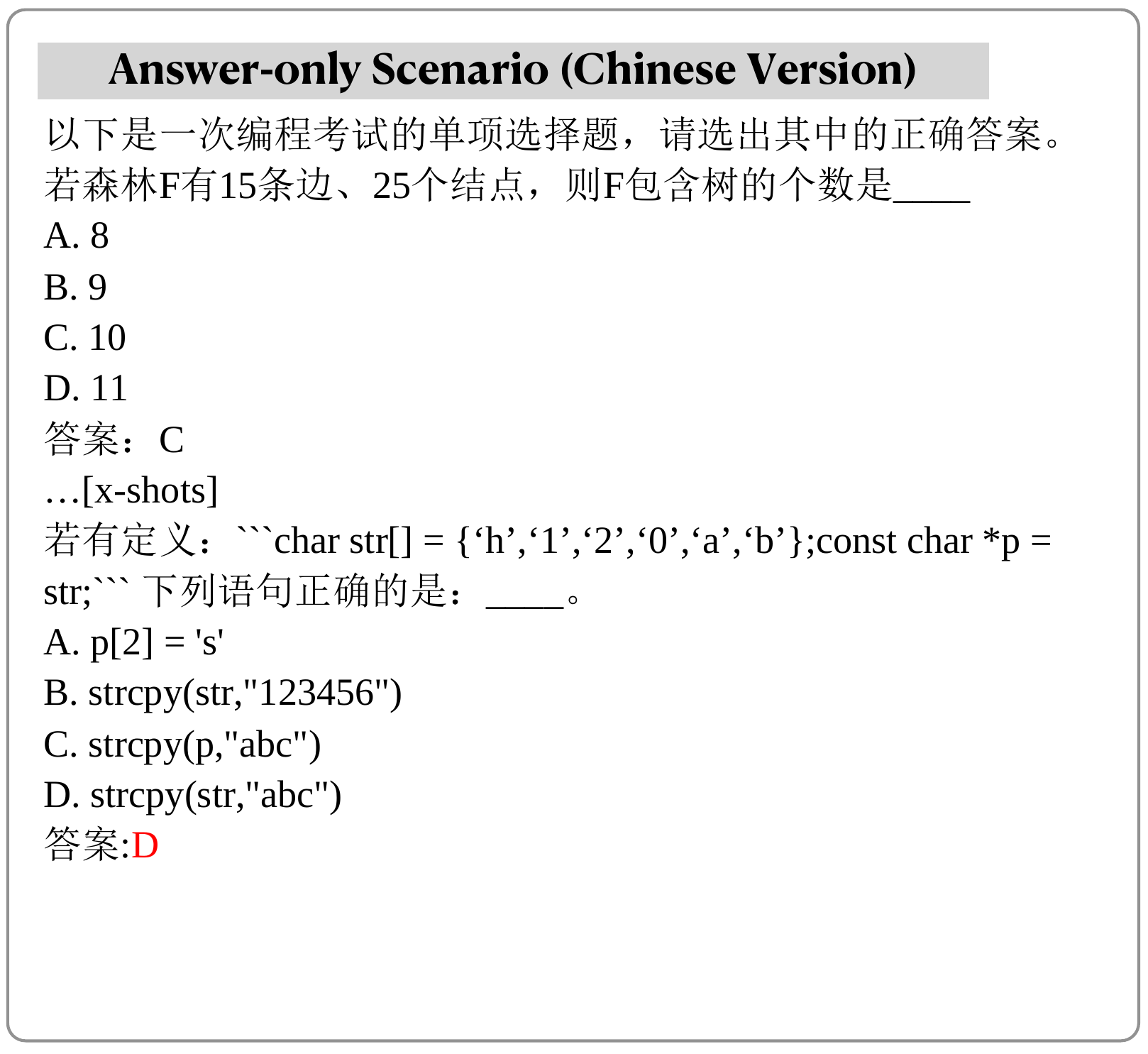}}
    \hspace{0.05cm} 
    \subfigure[Answer-only setting (English).]{\includegraphics[width=0.48\textwidth]{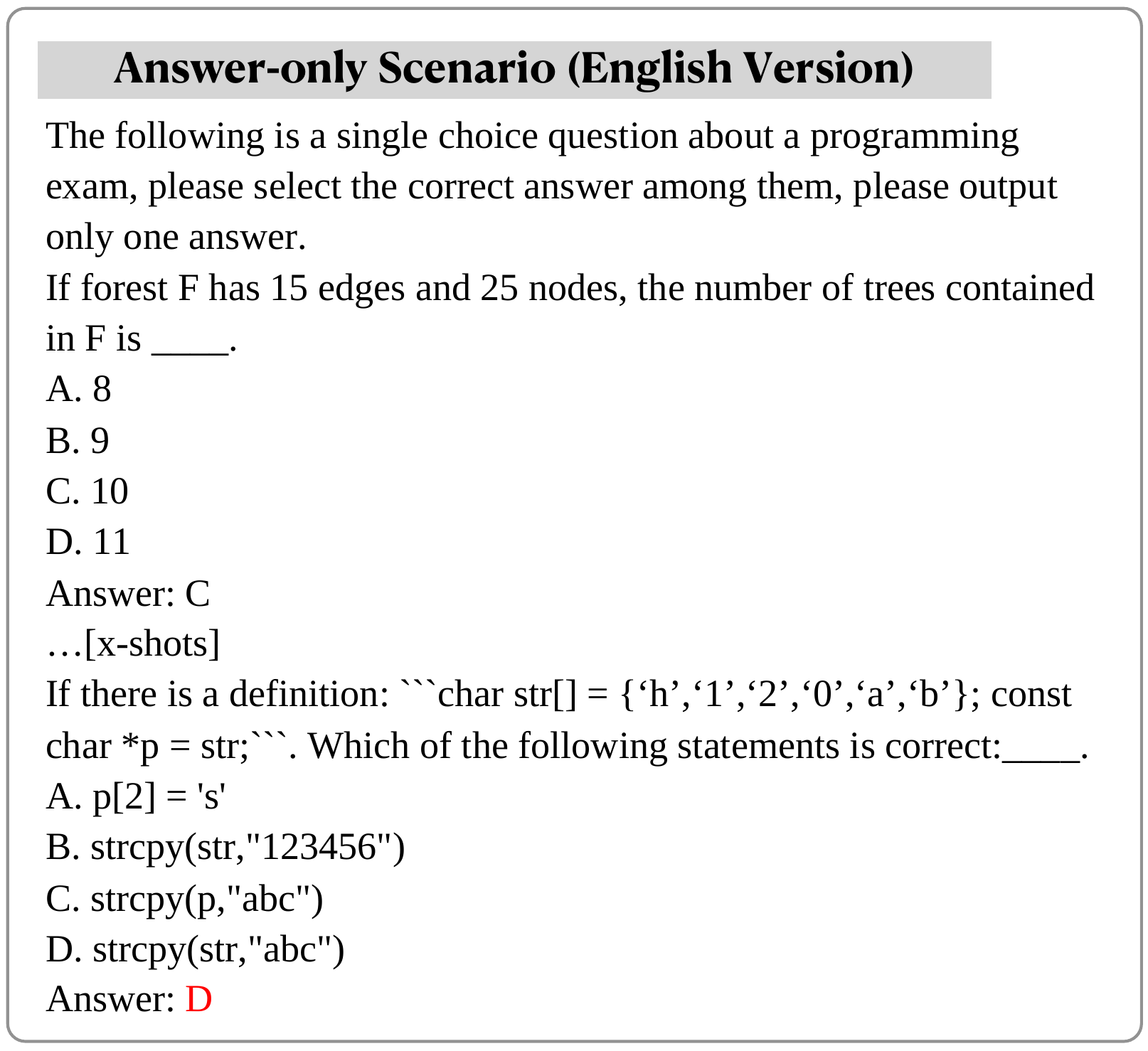}}

    \subfigure[Chain-of-Thought setting (Chinese).]{\includegraphics[width=0.48\textwidth]{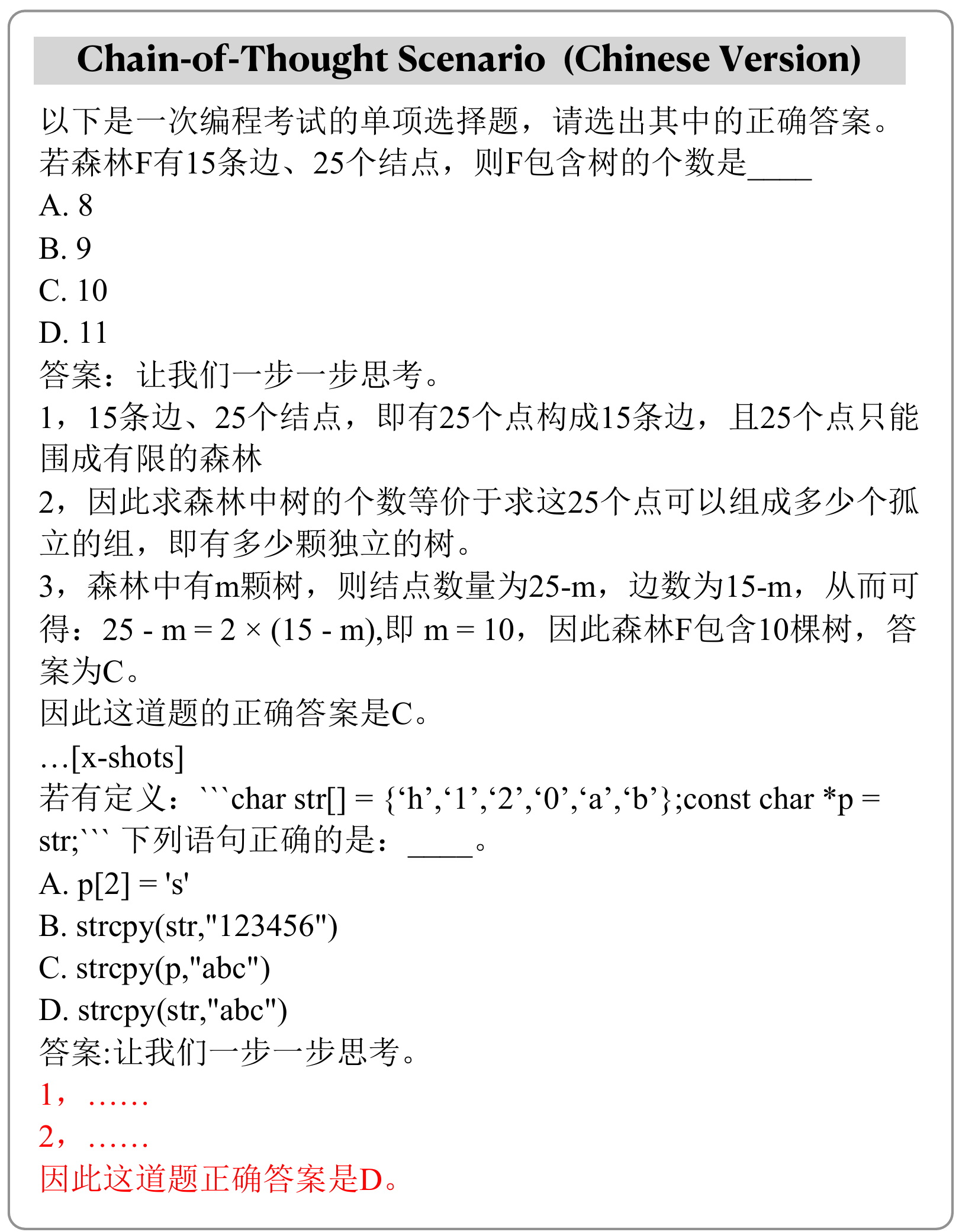}}
    \hspace{0.05cm} 
    \subfigure[Chain-of-Thought setting (English).]{\includegraphics[width=0.48\textwidth]{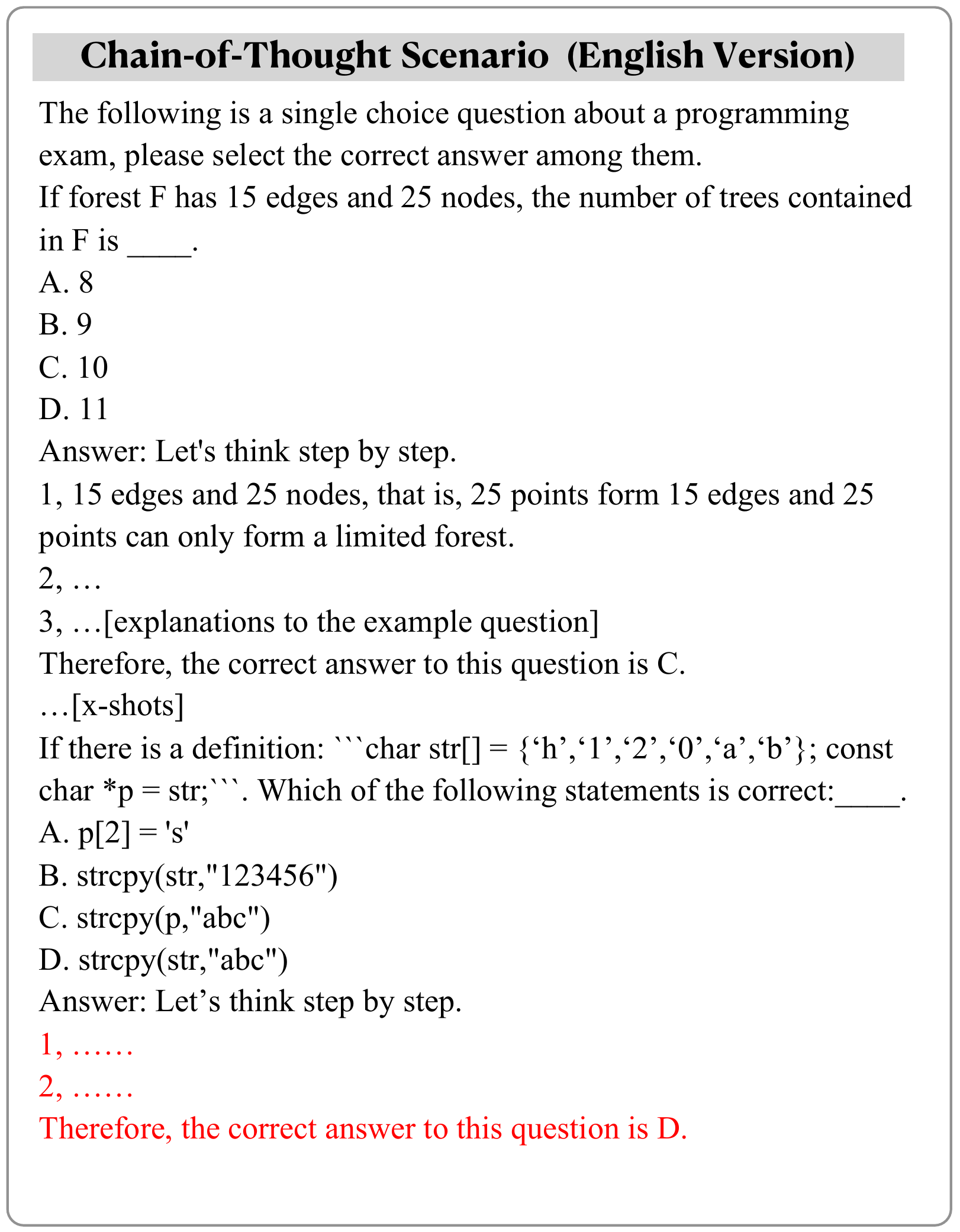}}

    \caption{Examples illustrating the programming comprehension task in answer-only and chain-of-thought scenarios, presented in both English and Chinese versions. The completed responses from LLMs are highlighted in red, while the input prompts are shown in black text.}
    \label{fig:programming-comprehension}
 
\end{figure}
\subsubsection{Prompting Strategy}
We evaluate the code comprehension abilities of LLMs under two different prompting strategies, i.e., answer-only and chain-of-thought.
In the answer-only setting, LLMs are required to generate only a single option (i.e., the selected choice), and the correctness of the generated option contributes to the final accuracy score. 
In the chain-of-thought setting, LLMs would first generate a piece of analytical text followed by a final answer, and the accuracy of the final answer serves as the measure of model performance. 
We illustrate the prompt examples under both settings in Figure~\ref{fig:programming-comprehension}.

\subsubsection{Evaluation Metric}
We adopt the accuracy (ACC) as the evaluation metric for the programming comprehension task. 
After LLMs provide a response under either answer-only or chain-of-thought settings, we extract the final choice using regular expressions. If the choice cannot be extracted via regular expressions, it is regarded as an incorrect answer. 
The final accuracy of the LLM is calculated as:
$$
\text{ACC } = \frac{\# \text{Correct Answers}}{\# \text{Questions}}.
$$


%% file: generation.tex
\subsection{Code Generation Task}
\subsubsection{Overview}
In the code generation task, LLMs have to extract relevant algorithm requirements from natural language descriptions, and then, in conjunction with a given code function framework, generate executable and accurate codes to pass the test cases. 
Such a task demands that large language models possess the comprehension ability for both natural languages and code structures, as well as the capability to analyze, design, and ultimately generate  functional code snippets.

\subsubsection{Data Processing}
\begin{figure}[t]
    \centering
    \includegraphics[width=0.5\textwidth]{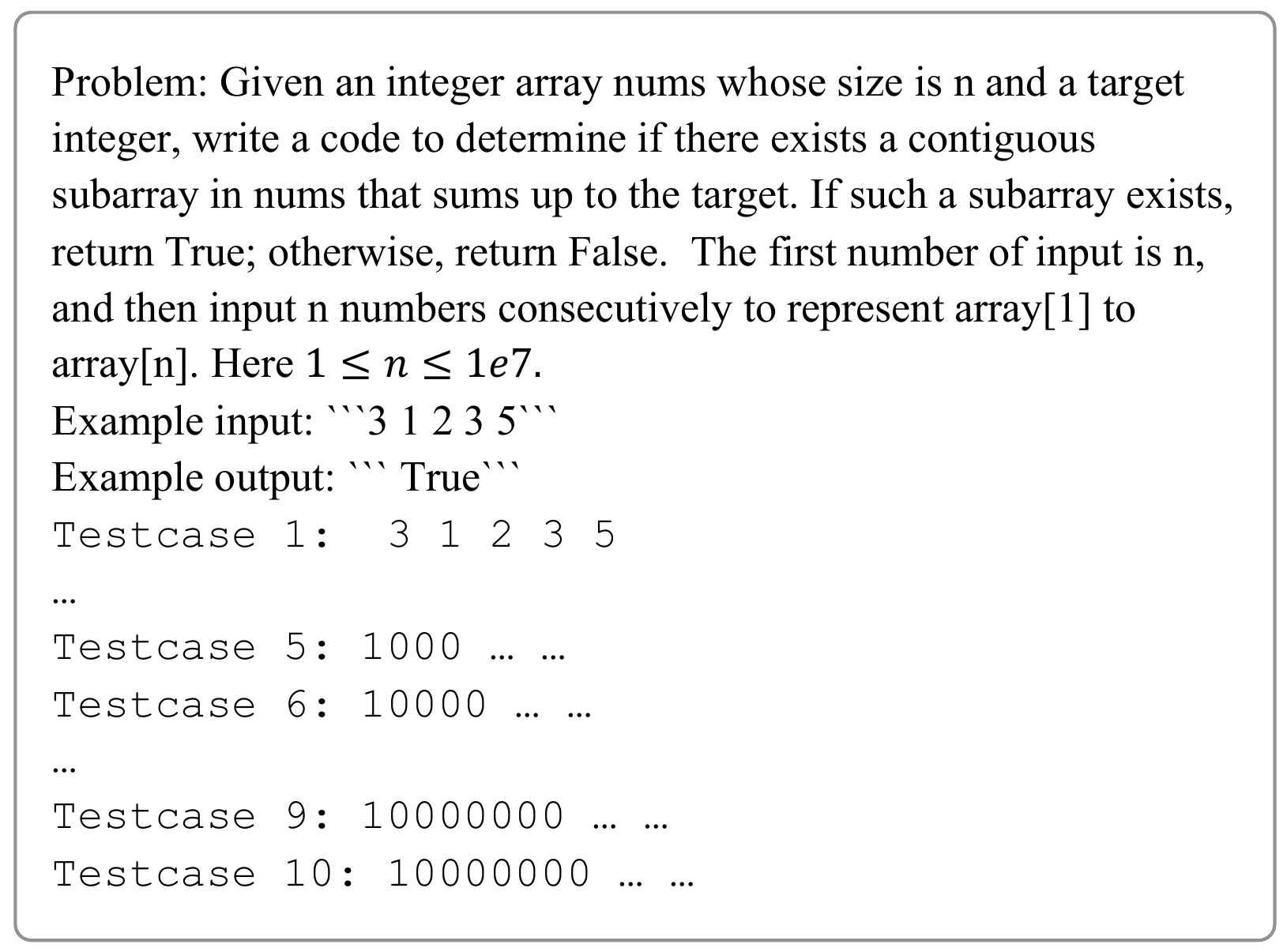}
    \caption{Illustration of testcases of different difficulty levels. If the complexity of the code is $O(n^2)$, it could pass at least the first six testcases. If the tester uses the prefix sum algorithms with $O(n)$ complexity, he can pass all test points.}
    \label{fig:unit-test}
\end{figure}

The test data is sourced from basic exercises on an internal Online Judge platform, which provides question descriptions, sample inputs/outputs, code templates, and test cases of varying difficulty levels. 
The question descriptions are typically provided in natural language, some of which also contain scenario-based descriptions. 
The sample inputs/outputs illustrate the format of the test case inputs and outputs, represented in Markdown format. The code templates represent fully functional code that passes all the tests and is used to generate test examples for \model. 
The difficulty levels of the test cases are illustrated in Figure \ref{fig:unit-test}.
Taking binary search as an example, the first five test cases have small input sizes $n$ and can be solved using a regular loop-based search.
However, the last five test cases have larger input sizes, and if a regular loop-based search is used, it will exceed the time limit. 
The privacy of data sources and the diversity of question descriptions decrease the possibility of test data appearing in the training corpus of LLMs. This promotes the fairness of evaluation and contributes to rational cognition of the programming abilities of different LLMs.

\begin{figure}[t]
    \centering
    \includegraphics[width=0.4\textwidth]{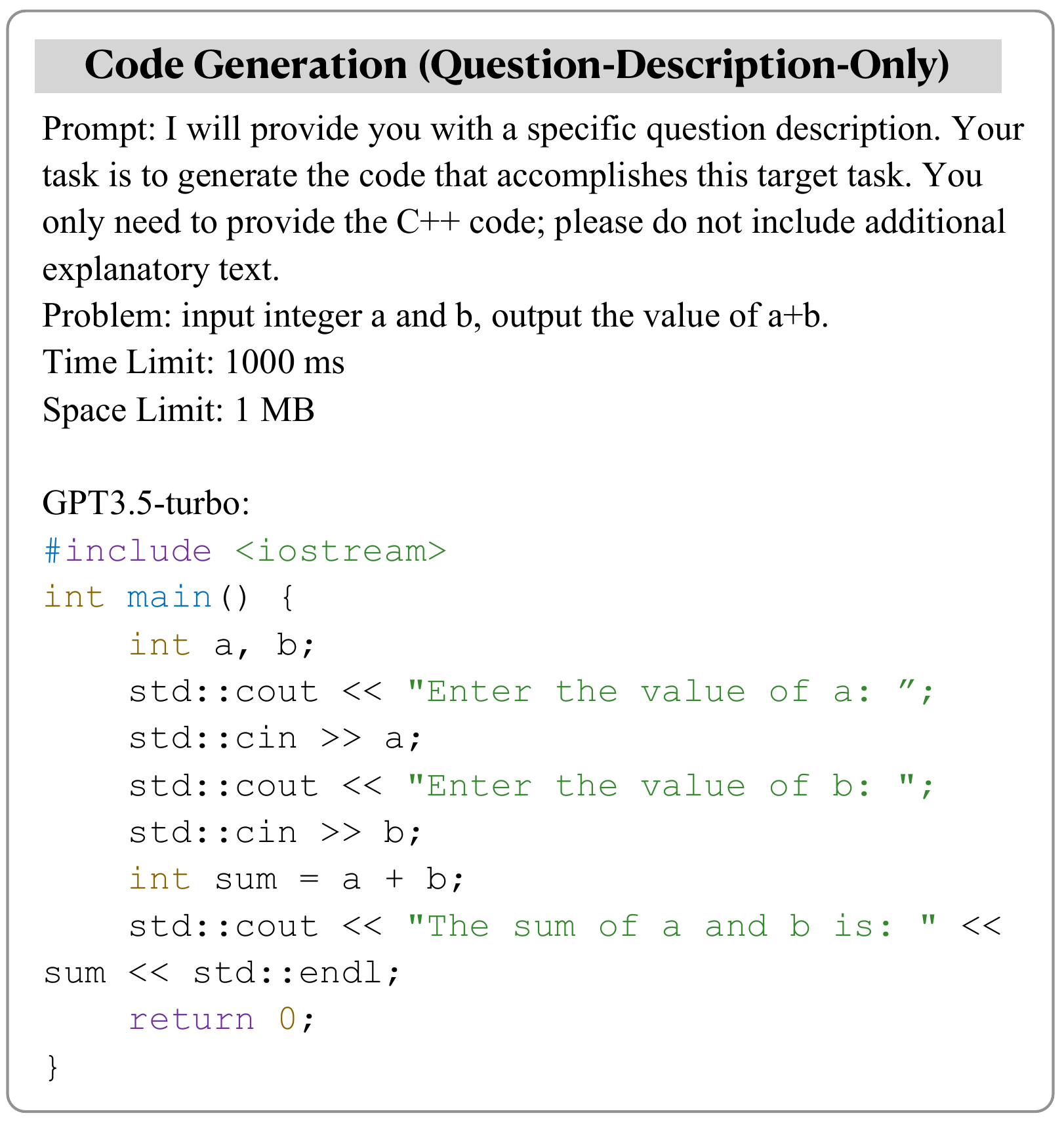}
    \caption{Illustration of the Question-Description-Only mode, where only the code question description is fed into LLMs for generation. The generated code by GPT3.5-turbo is demonstrated below. The code contains additional boot outputs, causing this logically correct code to fail the testcases.}
    \label{fig:QDO}
\end{figure}

Most LLMs are unable to generate code that meets the given input-output conditions based solely on question descriptions. 
When we directly input the original question description into LLMs, it is highly likely that the generated code will contain additional prompt-related output. 
Although the logic of the code itself may be correct, these extra outputs are not included in the original test cases, causing the generated code to fail the tests. 
Figure \ref{fig:QDO} illustrates an example of this situation, which occurs during code generation with GPT-3.5-turbo.
In such cases, in order to more accurately evaluate code generation capability of LLMs itself, we utilize GPT-4 to assist in generating functional form test examples.
With an in-context learning approach, we guide the LLMs in rewritting the template code and dividing it into a main function and some auxiliary functions. The main function includes (i) library references, (ii) the main function itself (with input and output code), and (iii) calls to auxiliary functions. The auxiliary functions contain the main algorithmic logic with input parameters and return the output to the main function.

To further explore the different types of code generation abilities of LLMs, we  divide the questions into two categories: explicit questions and narrative questions.
Explicit questions directly specify the functionality that the program needs to implement, such as "Implement a quicksort" or "Calculate the number of characters in the input." 
Narrative questions consist of more complex natural language descriptions, with the code requirements hidden within a scenario. 
For example, given a class of students, you need to inquire about the grade of the k-th student. 
This question requires storing the grades of the students in an array, sorting them, and then outputting the k-th largest element.
Previous benchmarks~\cite{humaneval} focus on explicit questions, and do not pay attention to the importance of evaluating narrative questions.

\subsubsection{Prompting Strategy}
\begin{figure}[t]
    \centering
    \subfigure[Function-only scenario (Chinese).]{\includegraphics[width=0.48\textwidth]{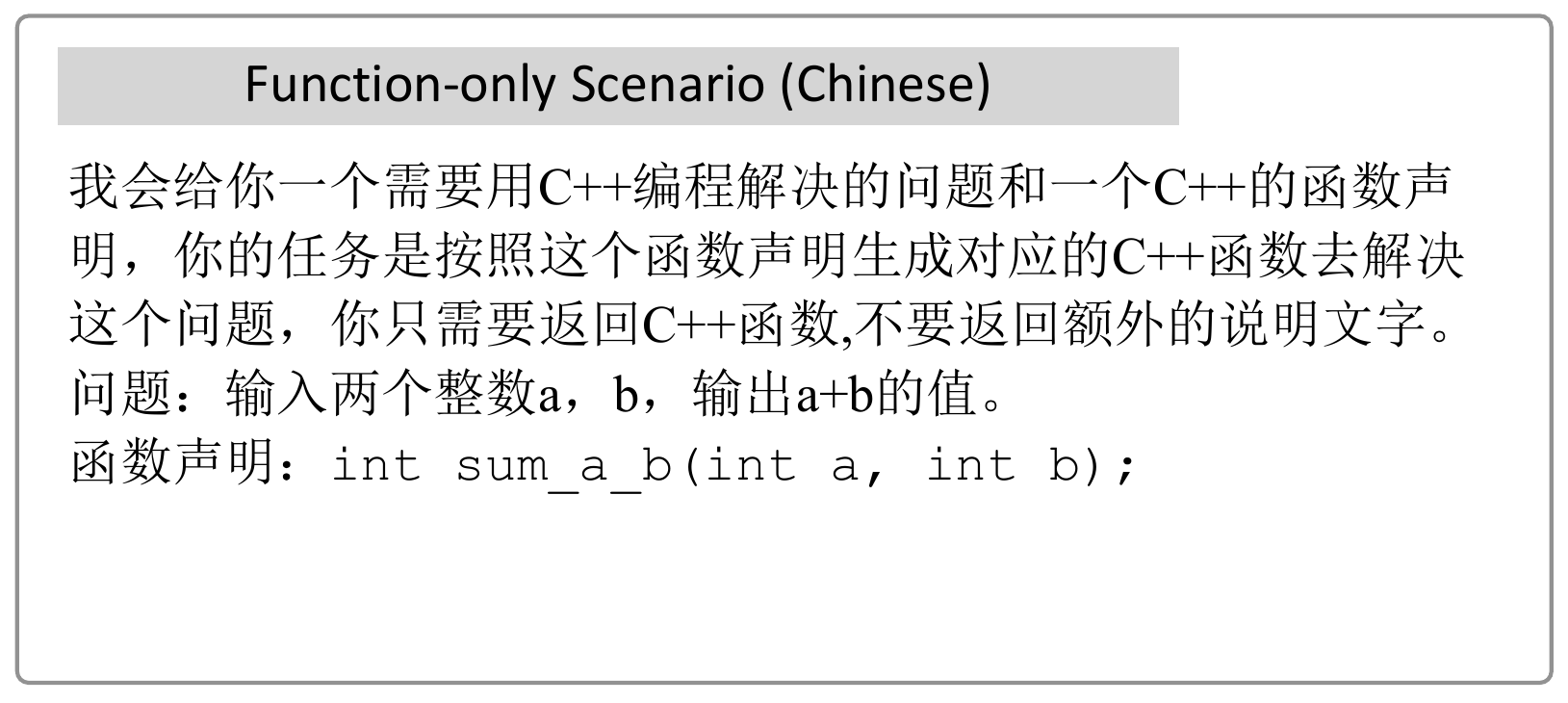}}
    \hfill
    \subfigure[Function-only scenario (English).]{\includegraphics[width=0.48\textwidth]{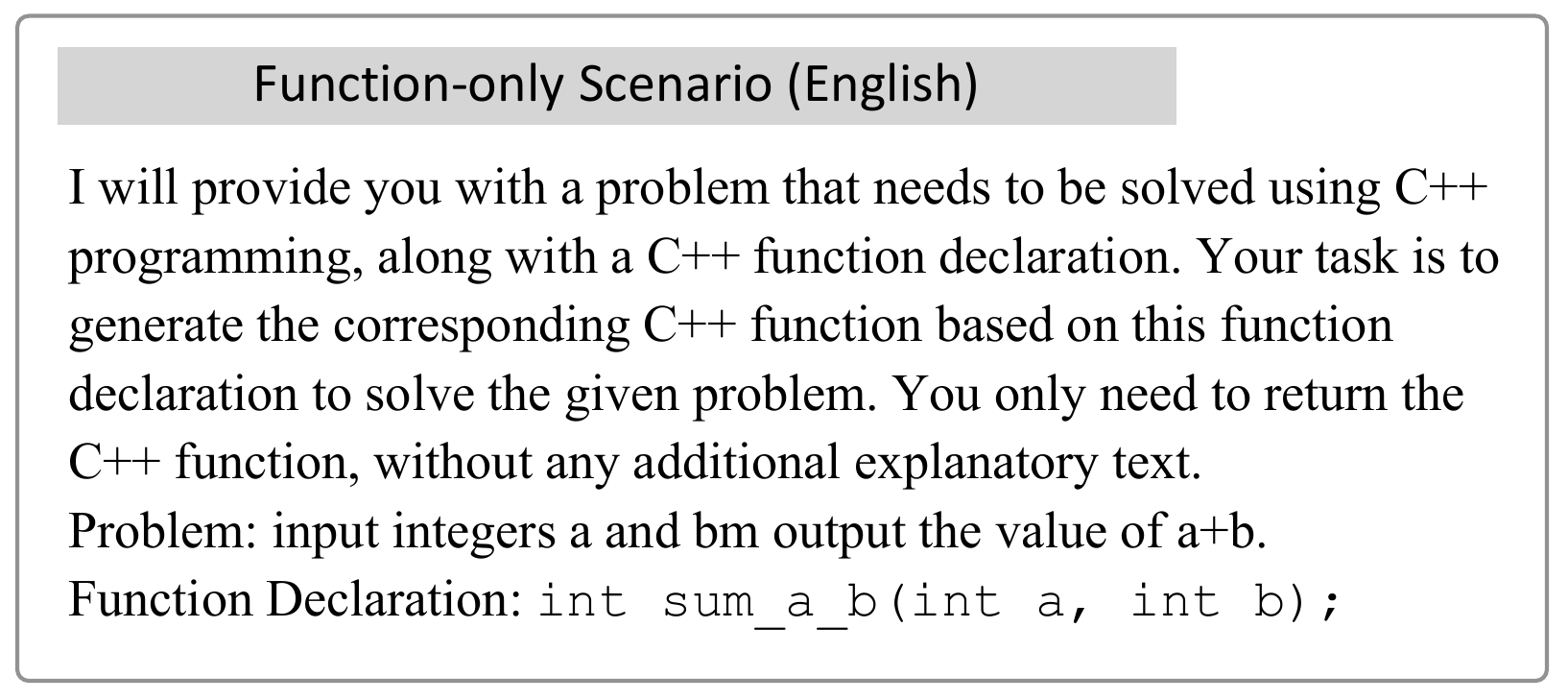}}

    \subfigure[Function-with-Context scenario (Chinese).]{\includegraphics[width=0.48\textwidth]{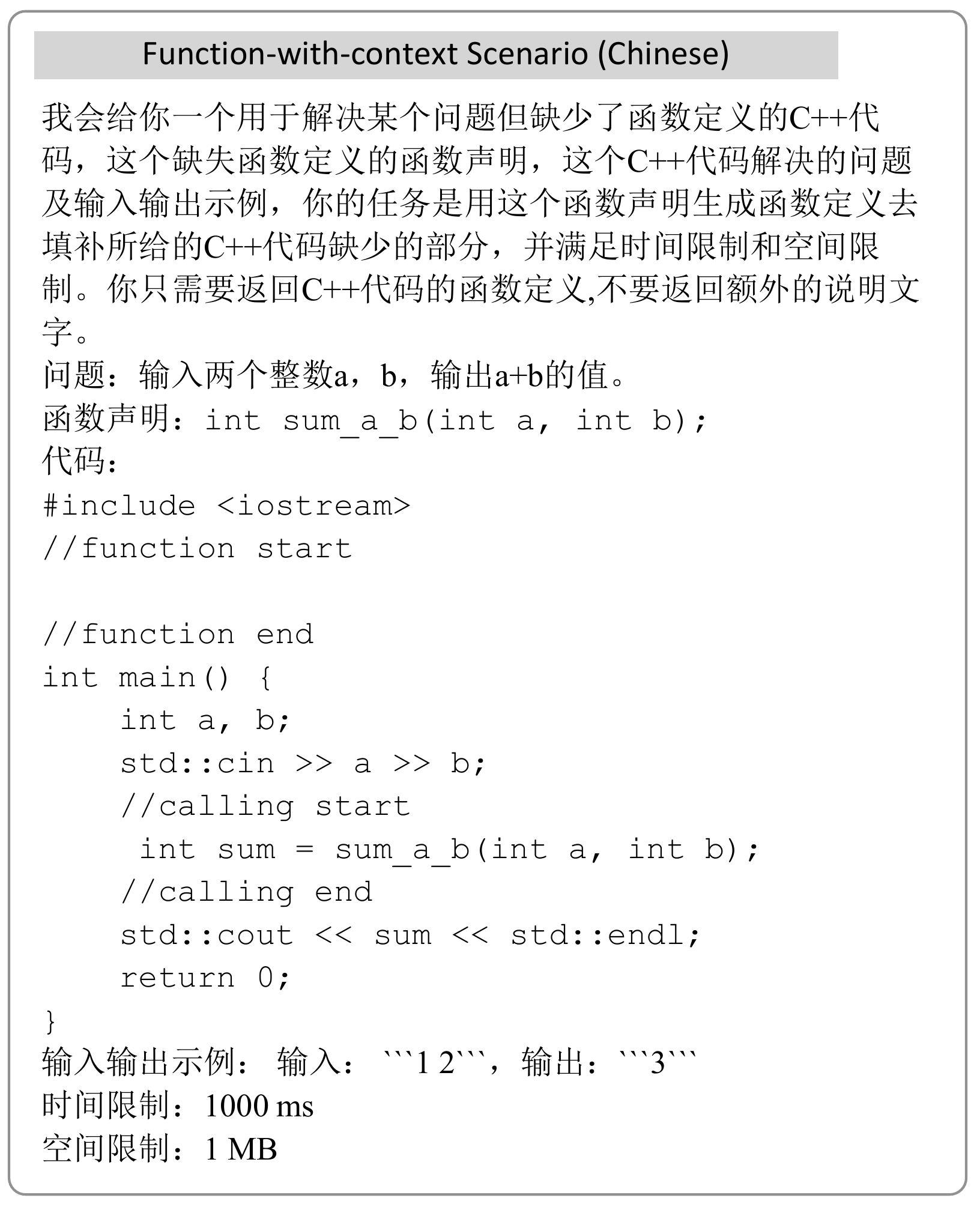}}
    \hfill
    \subfigure[Function-with-Context scenario (English).]{\includegraphics[width=0.48\textwidth]{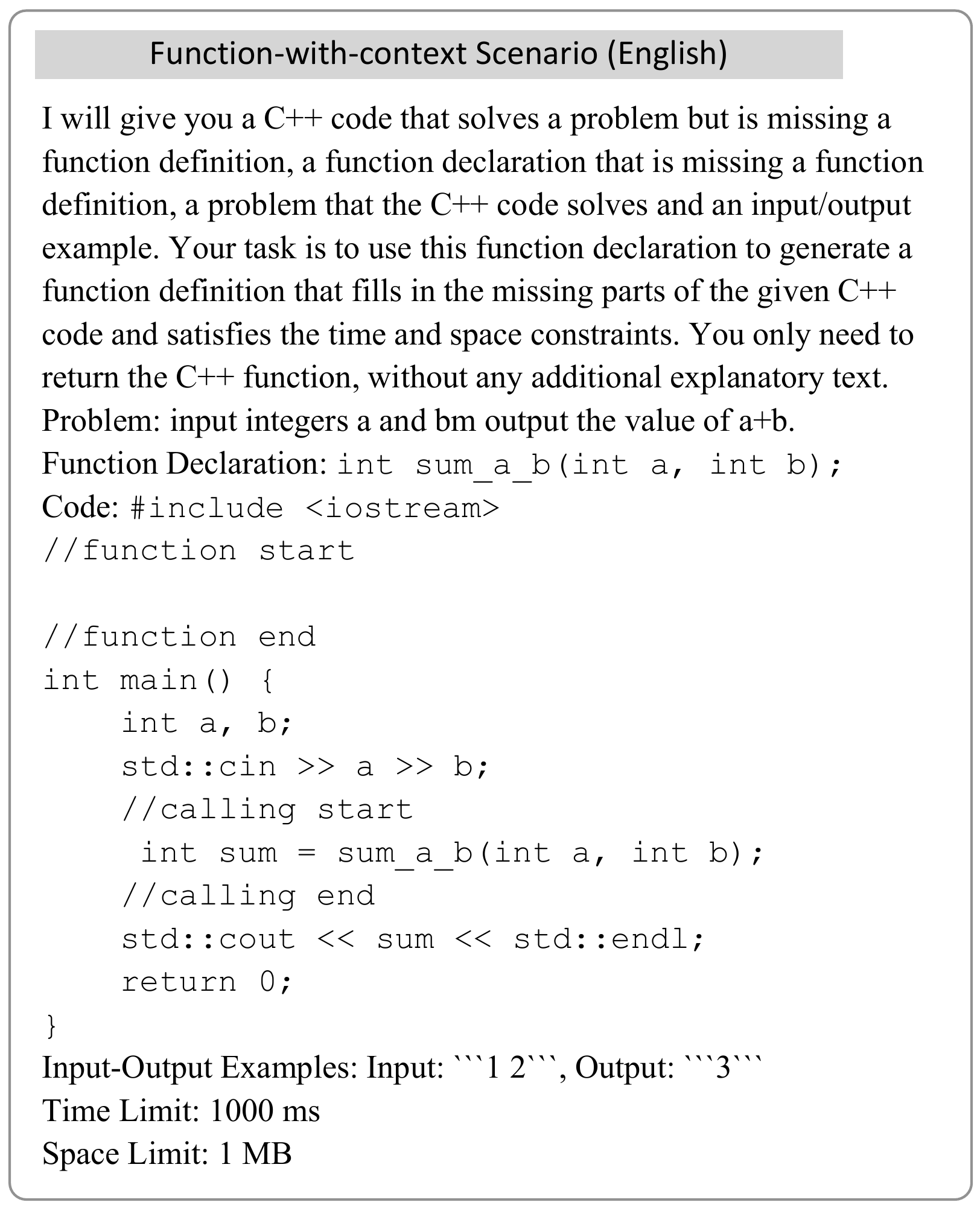}}
    \caption{Examples of specific design of the prompts for code generation task. The model generates the text highlighted in yellow.}
    \label{fig:CG}
\end{figure}

We describe our approach for generating target functions that adhere to function templates by designing prompts. 
These prompts consist of three components: question description, function description, and data range specification. 
In the 'function-only' scenario, no additional information is included in the prompts. In this case, LLMs are asked to generate a runnable function code that conforms to the function prototype and the question description.
In the function-with-context scenario, we incorporate the context of the main function as an additional input to guide the LLM to generate the target function. 
The specific design of the prompts is illustrated in Figure \ref{fig:CG}.

\subsubsection{Evaluation Metric}
To evaluate the effectiveness of the code, we use the original question's test cases to measure its correctness. 
Each question provides 5-10 test cases that cover various possible input scenarios and edge conditions. 
\model benchmark aligns with real-world scenarios of human programming by using the number of passed test cases as the model's score. 
All test case scores are equally weighted, regardless of their difficulty.

\model introduces an evaluation metric called accept rate as the evaluation function for the generated code by the model. 
When the language model generates a response, we extract the runnable C++ code function from it and concatenate it into the main function in the dataset. 
The concatenated code remains executable, and the target input-output pairs are aligned with the test cases in the original dataset. 
By comparing the output of the test cases with the actual output, we can get the number of the test cases the generated code passes, i.e., $\# \text{passes}(q)$. 
We calculate the $AC@k$, which represents the quality of the model-generated code
$$AC@k= \frac{1}{|Q|}\sum_{q\in Q} \left[\mathbb{I}(\# \text{passes}(q)\geq k)\right],$$ 
where $Q$ is the question set, and the indicator function $\mathbb{I}$ presents 
$$\mathbb{I}(\text{condition}) = \left\{\begin{aligned}
    1, & \text{\quad if condition is true,}  \\
    0, & \text{\quad otherwise.}
\end{aligned} \right.$$

 When $k$ is set to $1$, the code can pass at least one test case. $AC@1$ reflects the functional correctness of the model-generated code. In this case, some test cases may not pass, which may result from inappropriate algorithm selection in the generation process or exceeding the time/space limits of the question itself.
When $k$ is set to $all$, $AC@all$ represents the proportion of questions that have passed all test cases. This is a relatively strict metric for evaluating performance.
In addition, we introduce the Accept Rate (AC Rate) metric to measure how well the generated code meets.
The AC Rate is calculated as follows:
$$\text{AC Rate} = \frac{1}{|Q|}\sum_{q\in Q} \Big[ \frac{\#\; \text{passes}(q)}{\#\; \text{testcases}(q)}\Big].$$
It reflects the overall performance of LLMs in the code generation task, including both algorithm selection and implementation.

%% file: debug.tex
\subsection{Code Correction Task}
\subsubsection{Overview}
In code correction task, LLMs need to revise the generated code with the question description, error message and standard code.
Code correction is an essential stage in the programming procedure, the mastering of which could offer practical assistance in software engineering. In addition, utilizing the self-inspiration ability of LLMs to correct codes has emerged as a principle way for improving the accuracy of code generation~\cite{chen2023teaching}. Therefore, we include the evaluation for code correction, which is a fundamental ability of programming by LLMs.

\begin{figure}[htbp]
    \centering
    \subfigure[Code-only scenario (Chinese).]{ \includegraphics[width=0.4\textwidth]{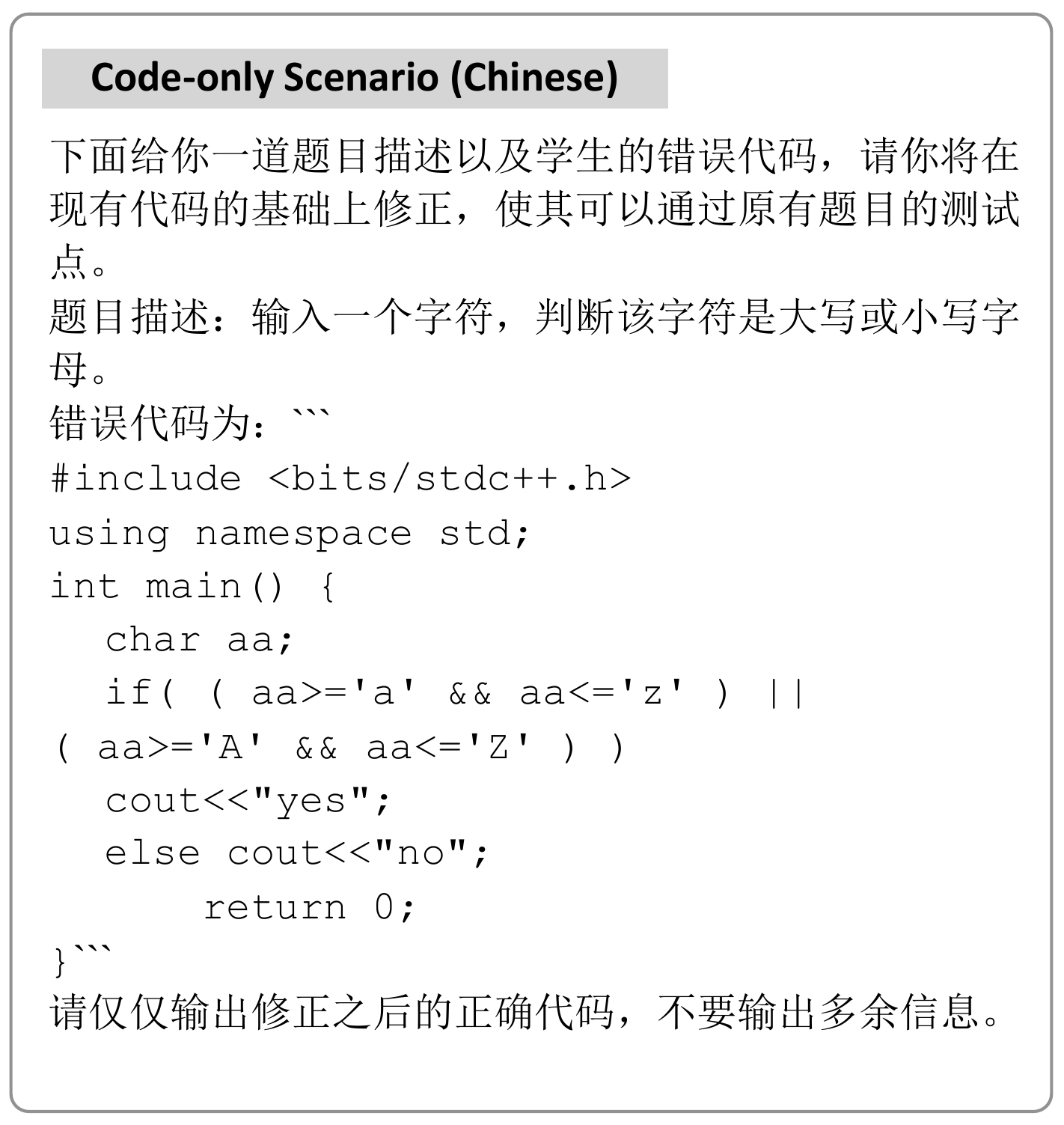}}
    \hspace{0.05cm} 
    \subfigure[Code-only scenario (English).]{ \includegraphics[width=0.4\textwidth]{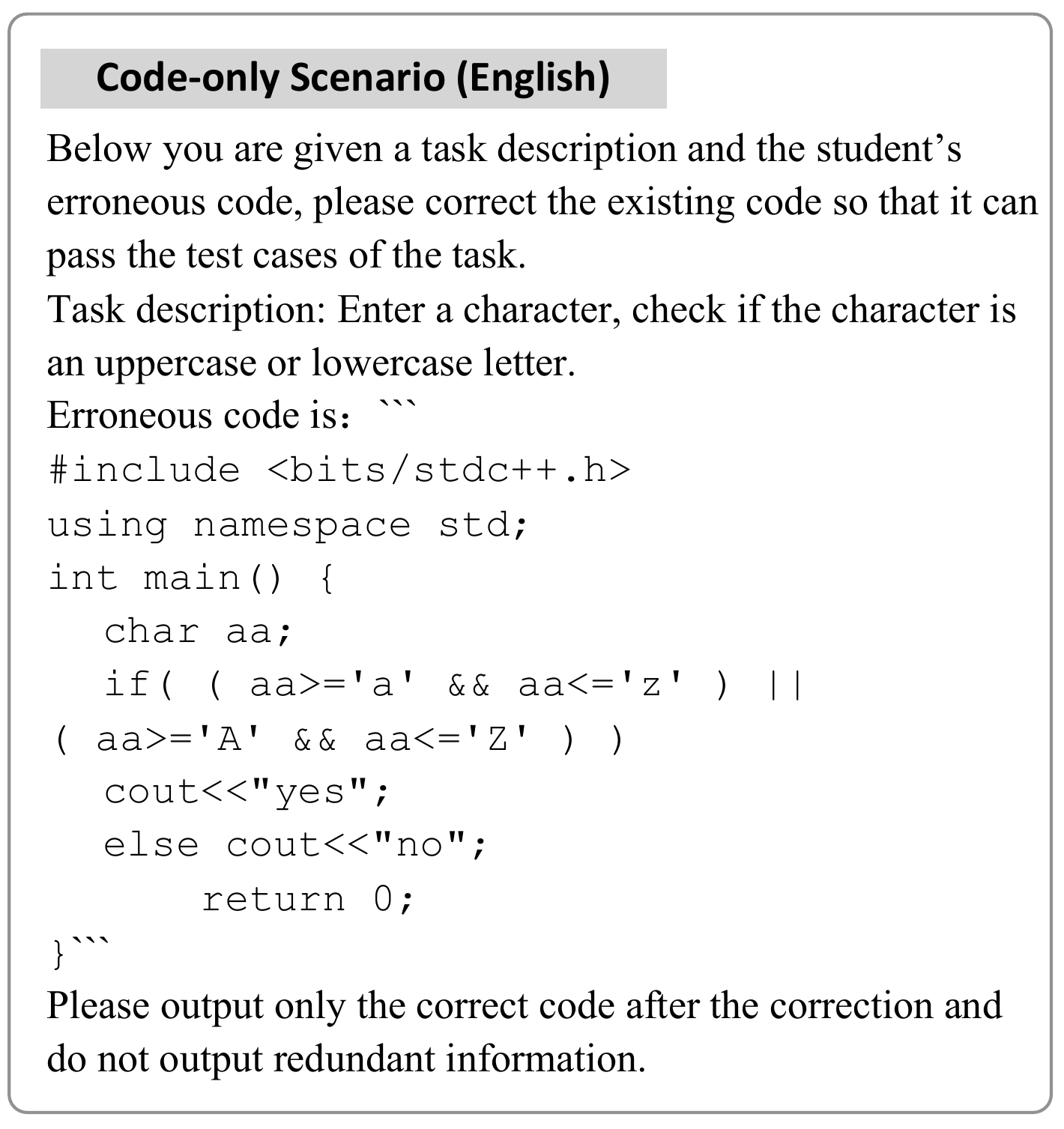}}

    \subfigure[Code-with-Standard Code scenario (Chinese).]{ \includegraphics[width=0.4\textwidth]{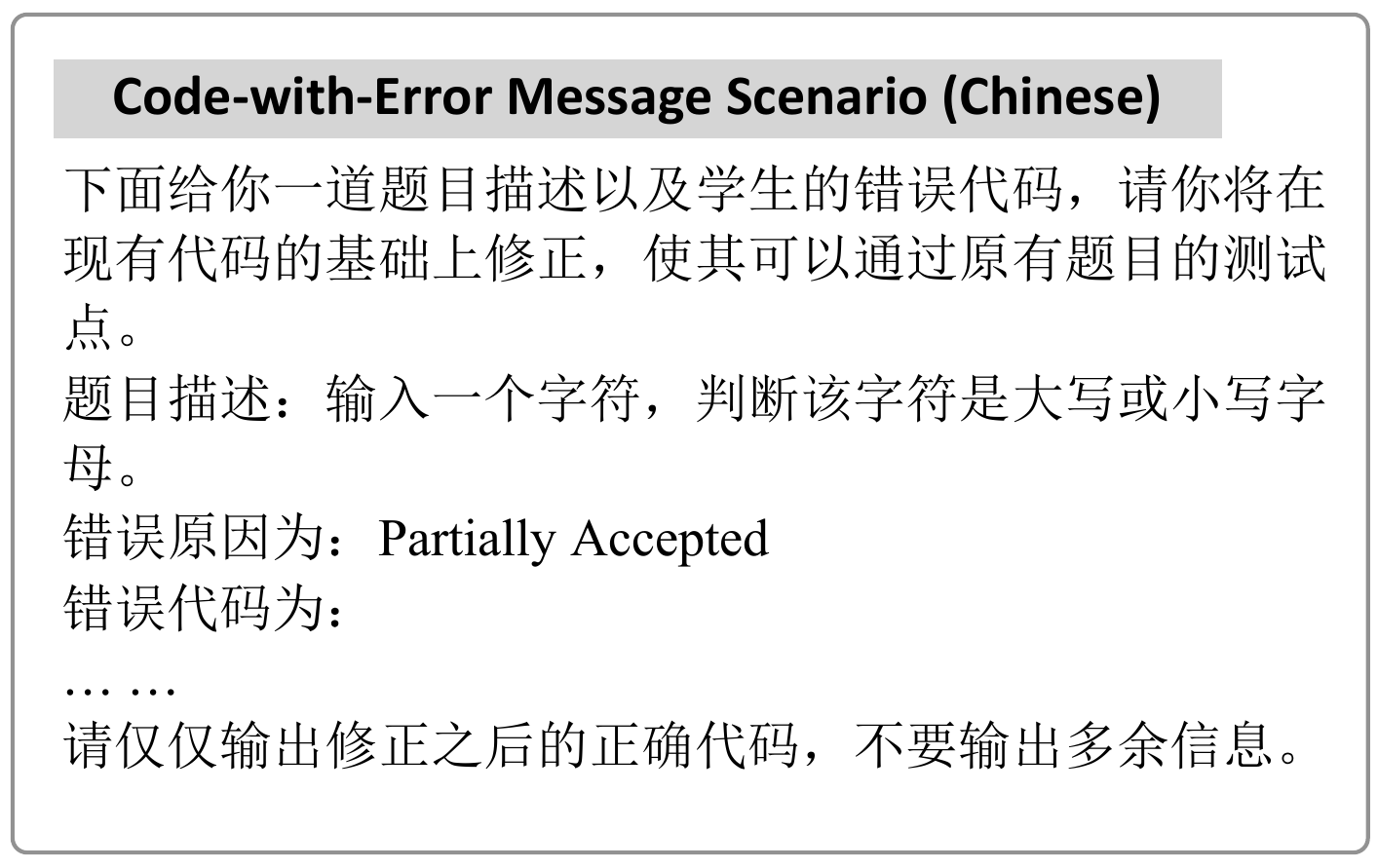}}
    \hspace{0.05cm} 
    \subfigure[Code-with-Error Message scenario (English).]{ \includegraphics[width=0.4\textwidth]{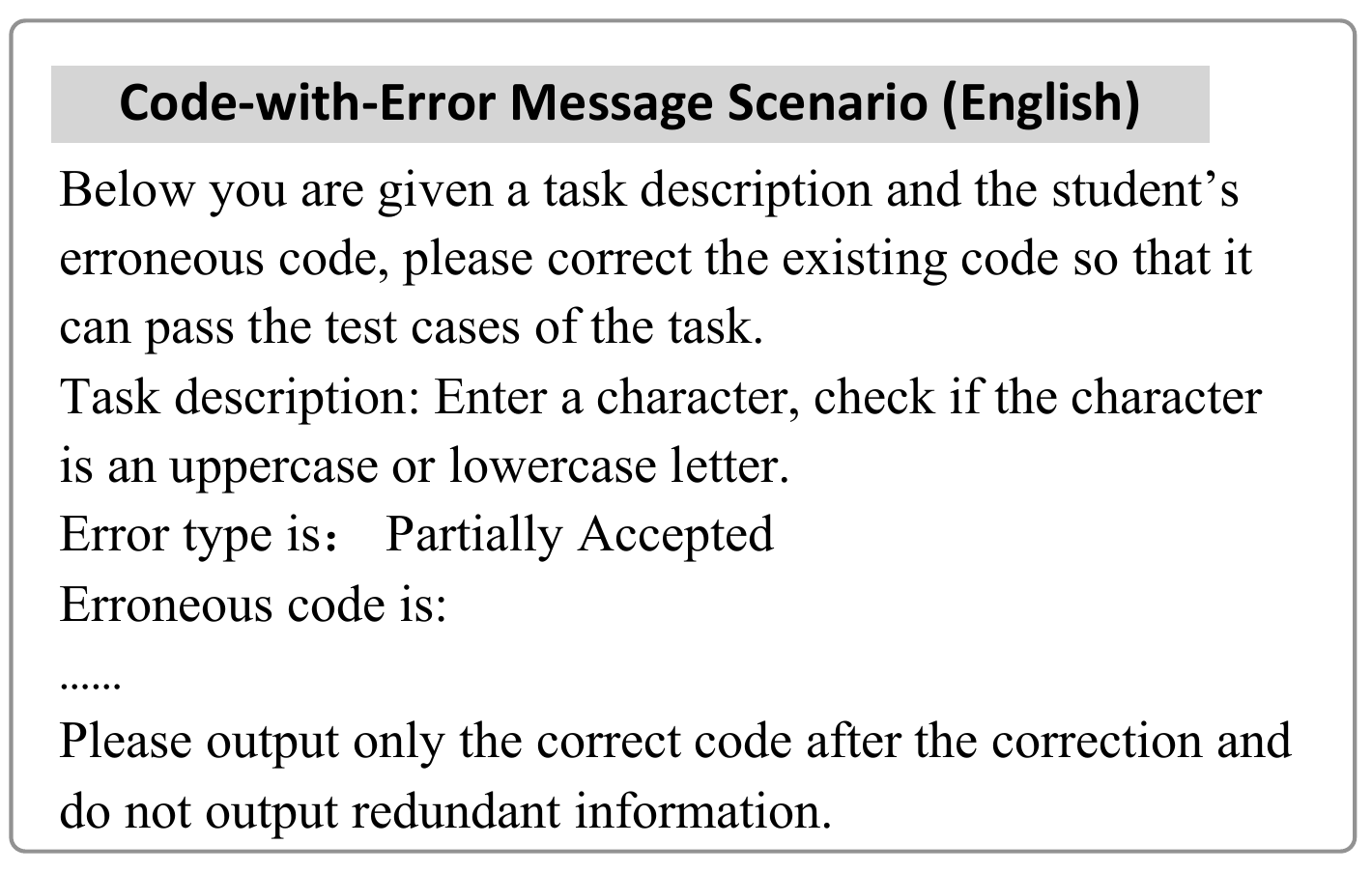}}

    \subfigure[Code-with-Error Message scenario (Chinese).]{ \includegraphics[width=0.4\textwidth]{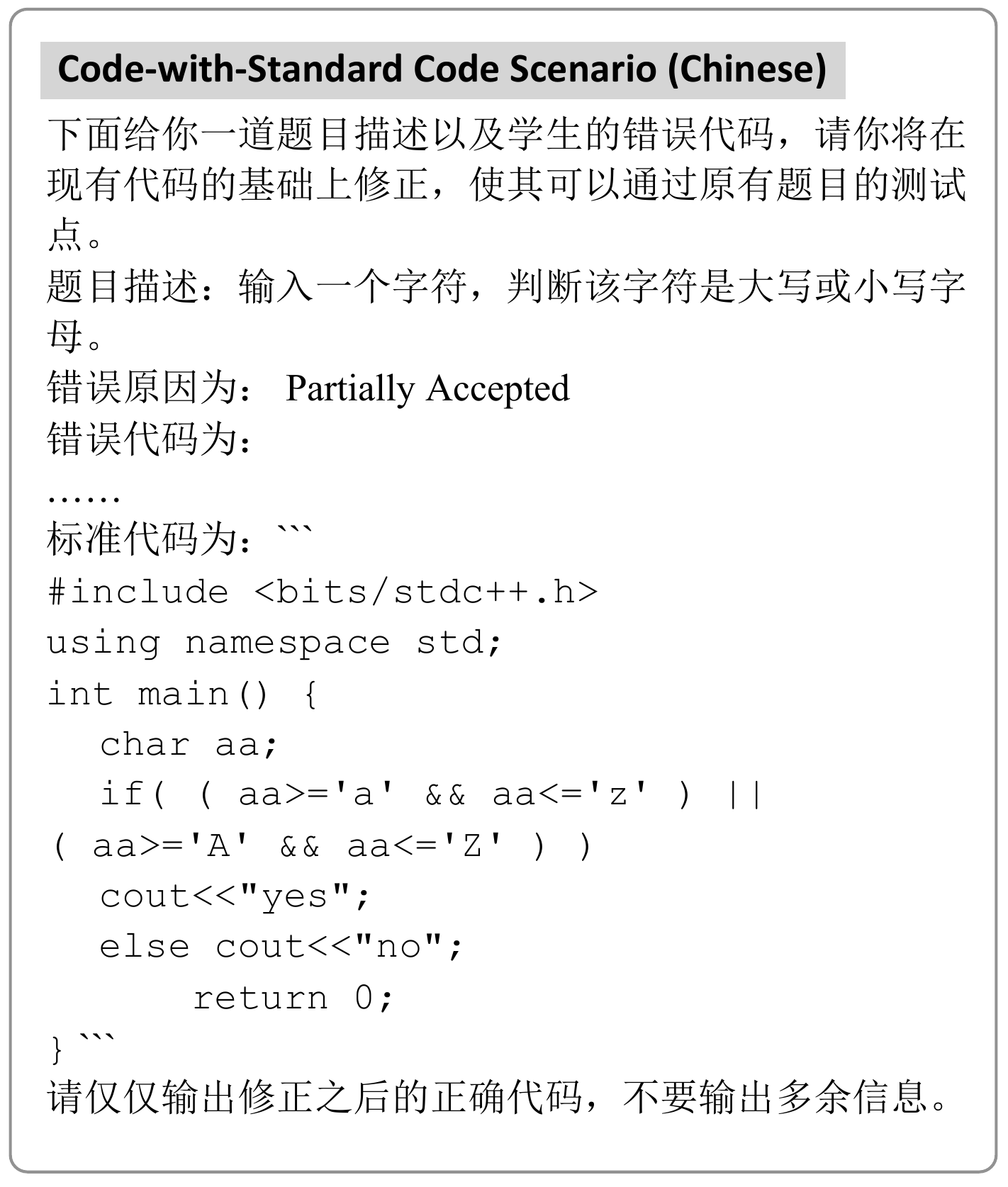}}
    \hspace{0.05cm} 
    \subfigure[Code-with-Standard Code scenario (English).]{ \includegraphics[width=0.4\textwidth]{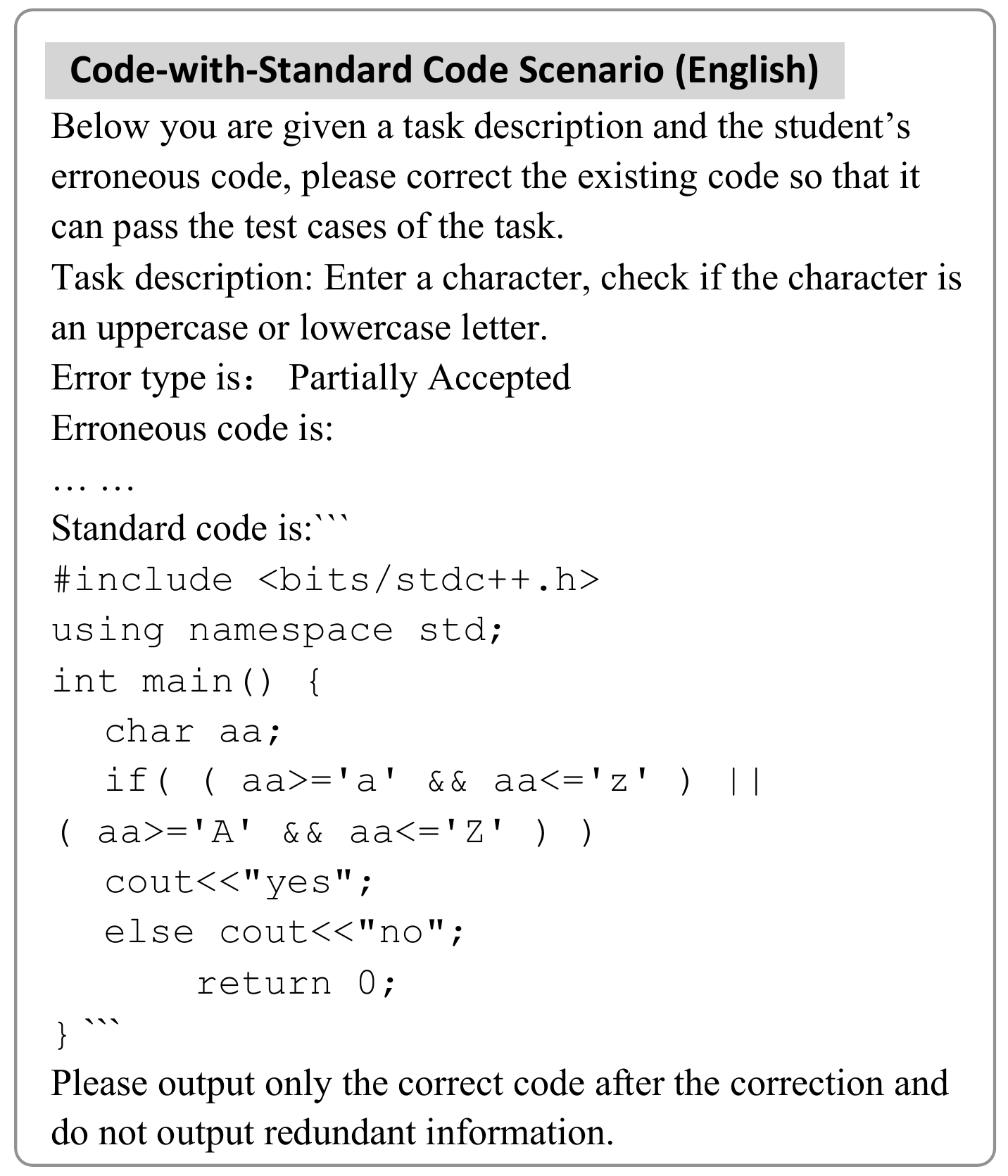}}
    
    \caption{Examples of specific design of the prompts for code correction task.}
    \label{fig:CC}
\end{figure}

\subsubsection{Data Processing}
The test data is originated from real students' submission of exercises on the internal Online Judge platform.
Each submission contains the exercise ID, submission code, error type, and submission time.
The types of error include: Wrong Answer (WA),  Time Limit Exceeded (TLE), and Runtime Error (RE).
If the submission answers all testcases correctly, then the error type is empty.


To ensure that erroneous code can be corrected within a certain number of steps rather than rewritten, we sample examples from submissions of the same users, where both the incorrect version and the correct version of the same problem exist. The incorrect version is then sampled to be the test sample, where we restrict the edit distance~\cite{levenshtein1966binary} between the incorrect version and the correct version to be less or equal to $50$.
From the generated code pairs, we select code pairs for each error type.
Due to varying frequencies of difference error types, the resulting error codes primarily consist of WA error, along with a portion of other error types. 
We also ensure that the number of code pairs for each question is roughly equal.

\subsubsection{Prompting Strategy}
We design three types of prompts for simulating different code correction scenarios.
The first type of prompt, Code-only, contains only the erroneous code and the question description without any additional information.
This prompt is used to evaluate code correction capability of LLMs, simulating humans directly reading code and identifying errors.
The second type of prompt, Code-with-Error Message, includes the erroneous code, task description, and the type of error. 
This prompt simulates engineers debugging their code based on actual feedback from the compiler, aligning with real-world development scenarios.
The third type of prompt is called Code-with-Standard Code. 
It builds upon the Code-with-Error Message prompt by adding standard code as additional input.
In the development process, engineers can refer to existing code for correction. Evaluating code correction ability of LLMs in this scenario can assess its comprehensive programming capabilities, providing a guidance for LLM-assisted programming.
The specific design of three types of prompts is demonstrated in Figure \ref{fig:CC}.

\subsubsection{Evaluation Metric}
Similar to code generation, we use the same test cases and same metrics (AC@1, AC@all, AC Rate) to measure the correctness of modified code.
LLMs should realize code correction based on the input erroneous code.
If the difference between the input and output code is too large (edit distance greater than 50), the output code is considered invalid. 
The score for invalid code is recorded as 0, regardless of whether it passes the test cases.
In code-with-standard code scenario, in order to prevent the LLM from directly outputting standard code, we validate the generated code in the Code-with-Standard Code scenario.
 Any output code that is identical to the standard code is directly marked as an incorrect answer. 

%% file: experiemnt.tex
\section{Evaluation Results}\label{exp}
\subsection{Experimental Setup}
\subsubsection{Dataset Statistics}
In the programming comprehension task, \model releases a total of 250 multiple-choice questions, which are divided into three categories: conceptual understanding (90 questions), commonsense reasoning (99 questions), and multi-hop reasoning (61 questions). 
In the code generation task, \model releases 476 algorithm questions in total. 
These questions are divided into 258 functional questions and 218 descriptive questions.
In the code correction task, \model releases 1330 segments of erroneous code, corresponding to solving 116 questions in the code generation task.
Each erroneous code has an error message, with a total of three types of error messages: Wrong Answer (1175 code segments), Time Limit Exceeded (69 code segments) and Runtime Error (86 code segments).

The test data is available on GitHub\footnote{\url{https://github.com/APEXLAB/CodeApex.git}} for users to download and run. The standard answers are not publicly disclosed to ensure the appropriate usage of \model, as its data might unintentionally include web-crawled content.  
Users are required to submit their model predictions to the provided webpage\footnote{https://apex.sjtu.edu.cn/codeapex/} to obtain an automated accuracy assessment, and a public leaderboard is maintained on the webpage. 
Users can choose whether to publish their models' scores on the leaderboard.

\subsubsection{LLMs to Evaluate}
CodeApex totally evaluates 12 LLMs, comprising a total of 41 variants. 
The tested LLMs and their variants are listed in Table \ref{tab:baselines}.
The variants of the models encompass the chatbot mode and the completion mode, as well as different parameter sizes and versions within the same base model. 
In the programming comprehension task, we conduct experiments on 10 LLMs, including both open-source models and API-based models.
In the code generation task, we evaluate 11 accessible LLMs, including both general-purpose and code-specialized models.
In the code correction task, due to the inability of some LLMs to output recognizable code, our experiments showcase the results of 7 models.
The largest open-source models have 13 billion parameters, while the largest parameterized model based on API is GPT4.

\paragraph{Parameter Settings}
When LLMs generate responses, we choose two distinct temperatures to cater to different application scenarios. 
For the programming comprehension task, we set temperature=0.01 to generate more conservative results, preventing the model from producing responses that do not align with the desired answers.
For the code generation and code correction task, we opted for a temperature of 0.7 and a top-p value of 0.8 to encourage the model to generate a broader range of possible codes, introducing more randomness in each code generation instance. 

\begin{table}[t]
    \centering
    \caption{CodeApex evaluates 14 LLMs along with their variants.}
    \label{tab:baselines}
    \renewcommand{\arraystretch}{1.0}
\resizebox{\columnwidth}{!}{%
    \begin{tabular}{lccc}
    \toprule\hline
      Model   & Model Size & Form & Creator \\ \hline
      ChatGLM\citep{liu2022chatglm} & 6B &  open, general-purpose &  \\
      ChatGLM2\citep{liu2022chatglm} & 6B & open, general-purpose & \\
      ChatGLM3\citep{liu2022chatglm} & 6B & open, general-purpose &  \multirow{-3}{*}{Tsinghua \& Zhipu}\\\hline
      MOSS\citep{sun2023moss} & 16B & open, general-purpose & Fudan University\\\hline
      Chinese-Alpaca\citep{chinese-llama-alpaca} & 7B & open, general-purpose & \citet{chinese-llama-alpaca}\\\hline
      BELLE\citep{belle2023exploring,BELLE} & 7B & open, general-purpose & Beike\\
      BELLE-LLaMA\citep{belle2023exploring,BELLE} & 7B, 13B & open, general-purpose & Beike\\\hline
      InternLM-Chat\citep{2023internlm} & 7B & open, general-purpose & Shanghai AI Lab \& Sense Time \\\hline
      Baichuan\citep{baichuan}  & 7B & open, general-purpose  & Baichuan Inc. \\\hline
      EduChat-base\citep{educhat2023} & 7B, 13B & open, Edu-Use &  \\
      EduChat-sft\citep{educhat2023} & 7B & open, Edu-Use & \multirow{-2}{*}{East China Normal University} \\\hline
      CodeT5-plus\citep{codet5} & 6B, 16B & open, code-specialized & Salesforce AI\\\hline
      WizardCoder\citep{luo2023wizardcoder} & 15B & open, code-specialized & Microsoft\\\hline
      StarCoder\citep{li2023starcoder} & undisclosed &  api, code-specialized & BigCode\\\hline
      Vicuna\citep{zheng2023vicuna} & 13B & open, general-purpose & LM-SYS \\\hline
      GPT-3.5-turbo\citep{zhang2019dialogpt} & undisclosed & api, general-purpose & \\
      GPT-4 & undisclosed & api, general-purpose &  \multirow{-2}{*}{OpenAI} \\\hline
      \bottomrule
    \end{tabular}
    }
\end{table}

\subsection{Programming Comprehension Results}
\begin{table}[htbp]
\caption{Accuracy results of Programming Comprehension Task in the answer-only setting. C.U., C.R., and M.H.R. indicate conceptual understanding, commonsense reasoning, and multi-hop reasoning questions, respectively. Each column displays the best accuracy among 5-shot, 2-shot, and 0-shot scenarios. The best performance of LLMs  is in bold, the second-best performance is underlined, and the third-best performance is wave underlined.  * represents chat-bot mode of the LLM. ZH and EN represent the results of Chinese and English test versions.}
\label{table:PC-AO}
\centering
\renewcommand{\arraystretch}{1.0}
\resizebox{\columnwidth}{!}{%
\begin{tabular}{l|cccccccc}
\toprule\hline
\multicolumn{1}{c|}{{}} & \multicolumn{8}{c}{{Programming Comprehension (Answer-only)}} \\ \cline{2-9} 
\multicolumn{1}{c|}{{}} & \multicolumn{4}{c|}{{ZH}} & \multicolumn{4}{c}{{EN}} \\ \cline{2-9} 
\multicolumn{1}{c|}{\multirow{-3}{*}{{Model}}} & \multicolumn{1}{c}{{C.U.}} & \multicolumn{1}{c}{{C.R.}} & \multicolumn{1}{c}{{M.H.R.}} & \multicolumn{1}{c|}{{Total}} & \multicolumn{1}{c}{{C.U.}} & \multicolumn{1}{c}{{C.R.}} & \multicolumn{1}{c}{{M.H.R.}} & \multicolumn{1}{c}{{Total}} \\ \hline
{ChatGLM-6B} & {{0.3222}} & {0.2626} & {0.3115} & \multicolumn{1}{c|}{{0.2960}} & {{0.3556}} & {0.2424} & {0.3115} & {0.2960} \\
{ChatGLM-6B*} & {0.3333} & {0.2727} & {{0.3115}} & \multicolumn{1}{c|}{{0.2960}} & {0.3926} & {0.2727} & {\uwave{0.3770}} & {0.3200} \\
{ChatGLM2-6B} & {0.3667} & {{0.2727}} & {0.3388} & \multicolumn{1}{c|}{{0.2987}} & {{0.4111}} & {0.3165} & {0.2951} & {0.2987} \\
{ChatGLM2-6B*} & {0.3444} & {{0.2626}} & {0.3115} & \multicolumn{1}{c|}{{0.3040}} & {{0.3556}} & {0.2828} & {{0.3115}} & {0.3040} \\
{ChatGLM3-6B} & {0.3770} & {0.2626}& {0.3770} & \multicolumn{1}{c|}{{0.3013}} & {0.3667} & {0.3300} & {0.2951} & {0.3240} \\
{ChatGLM3-6B*} & {0.4593} & {0.2727} & {0.2789} & \multicolumn{1}{c|}{{0.3080}} & {0.3407} & {0.2256} & {0.2951} & {0.2733} \\
{MOSS-16B*} & {0.3148} & {0.3401} & {{0.3279}} & \multicolumn{1}{c|}{{0.3120}} & {0.2630} & {0.3468} & {{0.3333}} & {0.3120} \\
{Chinese-Alpaca-7B} & {0.3481} & {0.2929} & {{0.2678}} & \multicolumn{1}{c|}{{0.2880}} & {0.3111} & {{0.2727}} & {0.3279} & {0.2947} \\
{Chinese-Alpaca-7B*} & {0.3000} & {{0.2896}} & {0.2787} & \multicolumn{1}{c|}{{0.2840}} & {{0.2926}} & {0.2896} & {0.2350} & {0.2840} \\
{Chinese-Alpaca-plus-7B} & {{0.2333}} & {0.3367} & {0.2623} & \multicolumn{1}{c|}{{0.2653}} & {0.3407} & {{0.2963}} & {0.3279} & {0.2987} \\
{Chinese-Alpaca-plus-7B*} & {0.3259} & {{0.3030}} & {0.2459} & \multicolumn{1}{c|}{{0.2853}} & {0.2630} & {0.2929} & {{0.2295}} & {0.2573} \\
{Chinese-Alpaca-13B} & {{0.2593}} & {0.2929} & {0.2623} & \multicolumn{1}{c|}{{0.2733}} & {{0.3111}} & {0.2323} & {0.2131} & {0.2733} \\
{Chinese-Alpaca-13B*} & {0.2741} & {0.2929} & {{0.2077}} & \multicolumn{1}{c|}{{0.2627}} & {0.2778} & {0.2795} & {{0.2514}} & {0.2547} \\
{Chinese-Alpaca-plus-13B} & {0.2667} & {0.3401} & {{0.2131}} & \multicolumn{1}{c|}{{0.2827}} & {{0.2444}} & {0.3569} & {0.3060} & {0.2827} \\
{Chinese-Alpaca-plus-13B*} & {0.2741} & {{0.2997}} & {{0.1967}} & \multicolumn{1}{c|}{{0.2573}} & {0.2889} & {{0.3569}} & {{0.2459}} & {0.2973} \\
{BELLE-7B-1M} & {{0.3333}} & {0.3199} & {0.2787} & \multicolumn{1}{c|}{{0.2947}} & {0.3333} & {0.3199} & {{0.2787}} & {0.3080} \\
{BELLE-7B-1M*} & {{0.3481}} & {0.3131} & {0.2459} & \multicolumn{1}{c|}{{0.3013}} & {{0.3889}} & {0.3131} & {0.2404} & {0.3040} \\
{BELLE-7B-2M} & {0.2444} & {0.3367} & {{0.2842}} & \multicolumn{1}{c|}{{0.2760}} & {0.2556} & {{0.3165}} & {0.2842} & {0.2613} \\
{BELLE-7B-2M*} & {0.2111} & {{0.3165}} & {0.2787} & \multicolumn{1}{c|}{{0.2413}} & {0.2667} & {{0.3165}} & {0.2459} & {0.2400} \\
{BELLE-LLaMA-7B-0.6M} & {{0.2926}} & {0.3401} & {0.3169} & \multicolumn{1}{c|}{{0.3053}} & {{0.3222}} & {0.3333} & {0.3279} & {0.2880} \\
{BELLE-LLaMA-7B-0.6M*} & {0.2778} & {{0.3131}} & {0.3115} & \multicolumn{1}{c|}{{0.3000}} & {0.3444} & {{0.2963}} & {0.3115} & {0.3000} \\
{BELLE-LLaMA-7B-2M} & {{0.2222}} & {0.2963} & {0.1967} & \multicolumn{1}{c|}{{0.2387}} & {{0.3000}} & {0.3333} & {0.2623} & {0.2680} \\
{BELLE-LLaMA-7B-2M*} & {0.2778} & {{0.3064}} & {0.2623} & \multicolumn{1}{c|}{{0.2840}} & {{0.2778}} & {0.3232} & {0.2623} & {0.2840} \\
{BELLE-LLaMA-13B-2M} & {0.2889} & {0.3333} & {{0.1967}} & \multicolumn{1}{c|}{{0.2840}} & {{0.2222}} & {0.3131} & {0.2131} & {0.2840} \\
{BELLE-LLaMA-13B-2M*} & {0.3444} & {0.2929} & {{0.2186}} & \multicolumn{1}{c|}{{0.2827}} & {0.2667} & {0.2660} & {{0.2186}} & {0.2693} \\
{InternLM-Chat-7B} & {{0.4556}} & {0.3434} & {0.3443} & \multicolumn{1}{c|}{{0.3720}} & {0.4444} & {{0.3636}} & {{0.3443}} & {0.3733} \\
{Baichuan-7B} & {0.3407} & {{0.3367}} & {0.2623} & \multicolumn{1}{c|}{{0.3147}} & {0.2926} & {{0.3064}} & {0.2951} & {0.3147} \\
{EduChat-base-002-7B*} & {{0.2259}} & {0.3131} & {0.2295} & \multicolumn{1}{c|}{{0.2480}} & {0.3296} & {0.3434} & {{0.3279}} & {0.3147} \\
{EduChat-base-002-13B*} & {0.2815} & {0.3535} & {0.2459} & \multicolumn{1}{c|}{{0.3013}} & {0.3667} & \uwave{0.3636} & {0.2459} & {0.3267} \\
{EduChat-sft-002-7B*} & {0.2444} & {0.2424} & {0.2951} & \multicolumn{1}{c|}{{0.2560}} & {0.2889} & {0.3434} & {0.2459} & {0.2920} \\
CodeT5-plus-16B & {0.2556} & {0.3030} & {0.3224} & \multicolumn{1}{c|}{{0.2640}} & {0.2074} & {0.2997} & {0.3224} & {0.2640} \\
CodeT5-plus-16B* & {0.2815} & \underline{0.3737} & {0.2732} & \multicolumn{1}{c|}{{0.3160}} & {0.1963} & \underline{0.3737} & {0.2787} & {0.2467} \\
CodeT5-plus-6B & {0.2333} & {{0.3535}} & {0.2404} & \multicolumn{1}{c|}{{0.2693}} & {0.2778} & {0.3535} & {0.2896} & {0.3173} \\
CodeT5-plus-6B* & {0.2296} & {0.2828} & {0.2568} & \multicolumn{1}{c|}{{0.2573}} & {0.2704} & {0.2828} & {0.2568} & {0.3040} \\
GPT-3.5-turbo & \uwave{0.6074} & {0.3300} & {0.6066} & \multicolumn{1}{c|}{\uwave{0.4893}} & {0.5222} & {0.3468} & \underline{0.6066} & {0.4893} \\
GPT-3.5-turbo* & {0.5963} & \uwave{0.3603} & \uwave{0.6393} & \multicolumn{1}{c|}{\uwave{0.5053}} & \uwave{0.5333} & {0.3603} & \underline{0.6066} & {0.4413}\\
GPT4 & \underline{0.6741} & \textbf{0.5455} & \textbf{0.8360} & \multicolumn{1}{c|}{\underline{0.6600}} & \textbf{0.6444} & \
\textbf{0.4848} & \textbf{0.8033} & \textbf{0.6160} \\
GPT4* & \textbf{0.7000} & \textbf{0.5455} & \underline{0.8197} & \multicolumn{1}{c|}{\textbf{0.6680}} & \underline{0.6222} & \textbf{0.4848} & \textbf{0.8033} & \underline{0.6120}\\
\hline
Human (closed-book) & 0.5818 & 0.6599 & 0.6683 & \multicolumn{1}{c|}{0.6343} & - & - & - & -\\
Human (open-book) & 0.8596 & 0.6821 & 0.8878 & \multicolumn{1}{c|}{0.7960} & - & - & - & -\\\hline
\bottomrule
\end{tabular}}
\end{table}

\begin{table}[htbp]
\caption{Accuracy results of Programming Comprehension Task in CoT setting. C.U., C.R., and M.H.R. indicate conceptual understanding, commonsense reasoning, and multi-hop reasoning questions, respectively. Each column displays the best accuracy among 5-shot, 2-shot, and 0-shot scenarios. The best performance of LLMs  is in bold, the second-best performance is underlined, and the third-best performance is wave underlined. * represents chat-bot mode of the LLM. ZH and EN represent the results of Chinese and English test versions. Since CodeT5 fails to generate a response with chain-of-thought, we exclude it in this setting.}
\label{table:PC-COT}
\centering
\renewcommand{\arraystretch}{1.0}
\resizebox{\columnwidth}{!}{%
\begin{tabular}{l|cccc|cccc}
\toprule
\hline
\multicolumn{1}{c|}{{}} & \multicolumn{8}{c}{{Programming Comprehension (Chain-of-Thought)}} \\ \cline{2-9} 
\multicolumn{1}{c|}{{}} & \multicolumn{4}{c|}{{ZH}} & \multicolumn{4}{c}{{EN}} \\ \cline{2-9} 
\multicolumn{1}{c|}{\multirow{-3}{*}{{Model}}} & \multicolumn{1}{c}{{C.U.}} & \multicolumn{1}{c}{{C.R.}} & \multicolumn{1}{c}{{M.H.R.}} & \multicolumn{1}{c|}{{Total}} & \multicolumn{1}{c}{{C.U.}} & \multicolumn{1}{c}{{C.R.}} & \multicolumn{1}{c}{{M.H.R.}} & \multicolumn{1}{c}{{Total}} \\ \hline
{ChatGLM-6B} & {0.3111} & {0.2121} & {0.3770} & \multicolumn{1}{c|}{{0.2880}} & {{0.3444}} & {0.1818} & {0.3115} & {0.2640} \\
{ChatGLM-6B*} & {0.3333} & {0.2222} & {{0.2787}} & \multicolumn{1}{c|}{{0.2480}} & {0.2889} & {0.2525} & {{0.3115}} & {0.2760} \\
{ChatGLM2-6B} & {0.4778} & {{0.2525}} & {0.3115} & \multicolumn{1}{c|}{{0.3240}} & {{0.3667}} & {0.2525} & {0.2951} & {0.2960} \\
{ChatGLM2-6B*} & {0.4111} & {{0.2929}} & {0.3770} & \multicolumn{1}{c|}{{0.3360}} & {{0.4000}} & {0.2525} & {{0.2787}} & {0.2960} \\
{ChatGLM3-6B} &  0.3630 & 	0.2626& 0.2514& 0.2827 &0.3963 &	0.2963	& 0.2841 & 0.2867  \\
{ChatGLM3-6B*} & 0.3519 &	0.2256 &	0.3497& 0.2733  & 0.3370 & 0.2054 & 0.2842 & 0.2693 \\
{MOSS-16B*} & {0.2667} & {0.1717} & {{0.3279}} & \multicolumn{1}{c|}{{0.2240}} & {0.2667} & {0.2020} & {{0.2787}} & {0.2360} \\
{Chinese-Alpaca-7B} & {0.2889} & {0.3030} & {{0.2131}} & \multicolumn{1}{c|}{{0.2680}} & {0.2556} & {{0.3333}} & {0.2131} & {0.2640} \\
{Chinese-Alpaca-7B*} & {0.2778} & {{0.3232}} & {0.2623} & \multicolumn{1}{c|}{{0.2640}} & {{0.2667}} & {0.3232} & {0.2295} & {0.2800} \\
{Chinese-Alpaca-plus-7B} & {{0.3667}} & {0.3131} & {0.2787} & \multicolumn{1}{c|}{{0.3240}} & {0.2778} & {{0.3030}} & {0.2951} & {0.2880} \\
{Chinese-Alpaca-plus-7B*} & {0.2667} & {{0.2626}} & {0.2295} & \multicolumn{1}{c|}{{0.2400}} & {0.2778} & {0.2626} & {{0.2623}} & {0.2480} \\
{Chinese-Alpaca-13B} & {{0.2889}} & {0.2929} & {0.2459} & \multicolumn{1}{c|}{{0.2640}} & {{0.3111}} & {0.2323} & {0.2295} & {0.2520} \\
{Chinese-Alpaca-13B*} & {0.3111} & {0.2727} & {{0.2295}} & \multicolumn{1}{c|}{{0.2640}} & {0.2444} & {0.2121} & {{0.2623}} & {0.2160} \\
{Chinese-Alpaca-plus-13B} & {0.3111} & {0.2929} & {{0.2131}} & \multicolumn{1}{c|}{{0.2600}} & {{0.3111}} & {0.2525} & {0.2951} & {0.2560} \\
{Chinese-Alpaca-plus-13B*} & {0.3556} & {{0.2727}} & {{0.2295}} & \multicolumn{1}{c|}{{0.2920}} & {0.3000} & {{0.2222}} & {{0.2951}} & {0.2560} \\
{BELLE-7B-1M} & {{0.3333}} & {0.2929} & {0.2459} & \multicolumn{1}{c|}{{0.2720}} & {0.2444} & {0.2222} & {{0.2131}} & {0.2120} \\
{BELLE-7B-1M*} & {{0.3333}} & {0.2525} & {0.2131} & \multicolumn{1}{c|}{{0.2600}} & {{0.2111}} & {0.2020} & {0.1475} & {0.1840} \\
{BELLE-7B-2M} & {0.2889} & {0.2626} & {{0.2295}} & \multicolumn{1}{c|}{{0.2240}} & {0.2556} & {{0.2222}} & {0.1311} & {0.2040} \\
{BELLE-7B-2M*} & {0.2667} & {{0.2424}} & {0.2131} & \multicolumn{1}{c|}{{0.2400}} & {0.1889} & {{0.2222}} & {0.1311} & {0.1880} \\
{BELLE-LLaMA-7B-0.6M} & {{0.2667}} & {0.3030} & {0.2951} & \multicolumn{1}{c|}{{0.2760}} & {{0.2222}} & {0.2323} & {0.2623} & {0.2320} \\
{BELLE-LLaMA-7B-0.6M*} & {0.3444} & {{0.3636}} & {0.3115} & \multicolumn{1}{c|}{{0.3200}} & {0.3333} & {{0.2424}} & {0.2623} & {0.2600} \\
{BELLE-LLaMA-7B-2M} & {{0.2444}} & {0.3030} & {0.2295} & \multicolumn{1}{c|}{{0.2640}} & {{0.2222}} & {0.2121} & {0.1311} & {0.1880} \\
{BELLE-LLaMA-7B-2M*} & {0.2778} & {{0.3030}} & {0.2459} & \multicolumn{1}{c|}{{0.2800}} & {{0.2111}} & {0.2020} & {0.1967} & {0.1880} \\
{BELLE-LLaMA-13B-2M} & {0.2889} & {0.2727} & {{0.1967}} & \multicolumn{1}{c|}{{0.2560}} & {{0.2556}} & {0.2222} & {0.1967} & {0.2120} \\
{BELLE-LLaMA-13B-2M*} & {0.2667} & {0.2828} & {{0.2131}} & \multicolumn{1}{c|}{{0.2600}} & {0.2778} & {0.1717} & {{0.1803}} & {0.2120} \\
{InternLM-Chat-7B} & {{0.3556}} & {0.1818} & {0.3607} & \multicolumn{1}{c|}{{0.2880}} & {0.4222} & {{0.2525}} & {{0.2623}} & {0.3160} \\
{Baichuan-7B} & {0.0667} & {{0.1111}} & {0.0656} & \multicolumn{1}{c|}{{0.0720}} & {0.1222} & {{0.1111}} & {0.2960} & {0.1000} \\
{EduChat-base-002-7B*} & {{0.2778}} & {0.2323} & {0.2623} & \multicolumn{1}{c|}{{0.2480}} & {0.2556} & {0.2424} & {{0.2623}} & {0.2360} \\
{EduChat-base-002-13B*} & {0.3000} & {0.2424} & {0.3279} & \multicolumn{1}{c|}{{0.2800}} & {0.3111} & {0.2424} & {0.2787} & {0.2680} \\
{EduChat-sft-002-7B*} & {0.3111} & {0.2121} & {0.2623} & \multicolumn{1}{c|}{{0.2520}} & {0.3111} & {0.2222} & {0.2623} & {0.2560} \\
GPT-3.5-turbo & {\uwave{0.5944}} & {{0.4141}} & {\uwave{0.6803}} & \multicolumn{1}{c|}{{\uwave{0.5260}}} & {\uwave{0.5278}} & {{0.3687}} & {\uwave{0.6393}} & {{0.4740}} \\
GPT-3.5-turbo* & {{0.5667}} & {\uwave{0.4242}} & {{0.6503}} & \multicolumn{1}{c|}{{{0.5187}}} & {{0.5074}} & {\uwave{0.3737}} & {{0.6339}} & {\uwave{0.4853}}\\
GPT4 & \underline{0.7111} & \underline{0.5152} & \textbf{0.8361} & \underline{0.6640} & \underline{0.6444} & \underline{0.5051} & \textbf{0.8689} &  \textbf{0.6320} \\
GPT4* & \textbf{0.7222} & \textbf{0.5859} & \underline{0.7869} & \multicolumn{1}{c|}{\textbf{0.6960}}  & \textbf{0.6556} & \textbf{0.5152} & \underline{0.8361} & \underline{0.6280} \\
\hline
Human (closed-book) & 0.5818 & 0.6599 & 0.6683 & \multicolumn{1}{c|}{0.6343} & - & - & - & -\\\hline
Human (open-book) & 0.8596 & 0.6821 & 0.8878 & \multicolumn{1}{c|}{0.7960} & - & - & - & -\\\hline
\bottomrule
\end{tabular}%
}
\end{table}

The overall performance of the programming comprehension task in the answer-only and CoT scenarios are presented in Table \ref{table:PC-AO} and Table \ref{table:PC-COT}.
The tables display the best results achieved in the 0-shot, 2-shot, and 5-shot settings. 
Detailed results for the N-shot ($N=0,2,5$) settings are provided in the Appendix \ref{app:x-shot}.
GPT4 obtains the highest average accuracy among all the models, followed by GPT3.5-turbo.
The third best model in Answer-only scenario is InternLM-Chat-7B, which has an accuracy of 37\%, which still have a huge gap between GPT.
It is noteworthy that the accuracy of most LLMs is below 50\%, highlighting that \model Benchmark is challenging in the programming comprehension task.

\paragraph{Bilingual Accuracy Comparison.}
\textbf{The Chinese version scored higher than the English version.}
There are two main reasons:
(1) The source question descriptions are obtained from final exams in Chinese universities, and thus the exam questions are initially presented in Chinese.
Even after translating them into English version, they still contain some language habits specific to Chinese speakers.
As a result, when inputting these English questions with biases into the LLM, some noise might be introduced into the model's encoding results.
(2) Most evaluated models are primarily trained on Chinese data, which leads to a poor understanding of English.
In Table \ref{table:PC-AO} and \ref{table:PC-COT}, LLM trained primarily on English corpus, such as codeT5 and GPT3.5-turbo, tend to have approximate performance on Chinese and English versions. 
 
\paragraph{Accuracy Comparison on Different Question Categories.} 
Among the three question categories, \textbf{approximately half of the models perform best in conceptual comprehension, followed by commonsense reasoning, with the lowest accuracy on multi-hop reasoning.}
This result indicates that LLMs are likely to incorporate knowledge of programming concepts during training. 
Most models scored higher in commonsense reasoning compared to multi-hop reasoning, suggesting that the capability of LLMs significantly decreases with an increase in the number of reasoning steps.

\paragraph{Effects of Chain-of-Thought Prompting.} 
\textbf{Most of the models achieve approximate or lower accuracy than the answer-only setting.}
The accuracy results of the CoT setting are depicted in Table \ref{table:PC-COT}.
The reasons for this observation are two folds:
(1) Models we evaluate do not reach model sizes that have the emergent ability of CoT.
According to \cite{wei2022emergent}, the emergence of CoT requires LLM to have at least 60B parameters. 
When the parameter number is not enough, the CoT setting might introduce additional noise, and the generated response of LLM would be unstable. That's why GPT3.5-turbo, which has reached the emergence point, achieves higher accuracy in the CoT setting.
(2) When answering conceptual understanding and commonsense reasoning questions, we do not require multi-step reasoning.
Thus, the CoT ability of LLMs does not provide assistance for these categories of questions.
However, for multi-hop reasoning questions, there is a noticeable improvement in accuracy in the CoT scenario for some models (such as ChatGLM2, educhat, and GPT3.5-turbo).
Since CodeT5 fails to generate a response with chain-of-thought, we exclude it in the CoT setting.

\paragraph{Human Performance}
\textbf{Novice programmers perform similarly to GPT-4 in closed-book tests after learning, while human performance in open-book exams is significantly better than all LLMs.}
The performance of human testers is shown in Table \ref{table:PC-AO}. 
Note that programming comprehension tasks in \model is considered a semi-open-book exam for LLMs, i.e., they have limited offline knowledge base.

\subsection{Code Generation Results}
\begin{table}[t]
\caption{Performance on AC@1, AC@all, and AC Rate metrics of code generation task in Chinese version. The best performance of LLMs  is in bold, the second-best performance is underlined, and the third-best performance is wave underlined. General-purpose LLMs that are not able to generate code are removed from the table.}
\label{tab:CG-ZH}
\renewcommand{\arraystretch}{1.0}
\centering
\begin{tabular}{l|ccc|ccc}
\toprule
\hline
\multicolumn{1}{c|}{{}} & \multicolumn{6}{c}{{Code Generation (Chinese)}} \\ \cline{2-7} 
\multicolumn{1}{c|}{{}} & \multicolumn{3}{c|}{{Function-only}} & \multicolumn{3}{c}{{Function-with-Context}} \\ \cline{2-7} 
\multicolumn{1}{c|}{\multirow{-3}{*}{{Model}}} & \multicolumn{1}{c}{{AC@1}} & \multicolumn{1}{c}{{AC@all}} & \multicolumn{1}{c|}{{AC Rate}} & \multicolumn{1}{c}{{AC@1}} & \multicolumn{1}{c}{{AC@all}} & \multicolumn{1}{c}{{AC Rate}} \\ \hline
{ChatGLM-6B} & {0.2143} & {0.0924} & \multicolumn{1}{l|}{{0.1371}} & {0.1576} & {0.0693} & {0.1063} \\
{ChatGLM2-6B} & {0.2143} & {0.1197} & \multicolumn{1}{l|}{{0.1560}} & {0.1996} & {0.0777} & {0.1274} \\
{ChatGLM3-6B} & 0.1387&	0.1092&	0.1330 & 0.0903 &	0.0651	& 0.0854 \\
{MOSS-16B*} & {0.2374} & {0.1492} & \multicolumn{1}{l|}{{0.1879}} & {0.2458} & {0.1282} & {0.1770} \\
{Chinese-Alpaca-plus-13B} & {0.2773} & {0.1387} & \multicolumn{1}{l|}{{0.1867}} & {0.2710} & {0.1366} & {0.1886} \\
{BELLE-7B-1M} & {0.1639} & {0.0588} & \multicolumn{1}{l|}{{0.0954}} & {0.1513} & {0.0651} & {0.0941} \\
{BELLE-LLaMA-13B-2M} & {0.1912} & {0.0903} & \multicolumn{1}{l|}{{0.1283}} & {0.1996} & {0.0840} & {0.1274} \\
{InternLM-Chat-7B} & {0.1450} & {0.0735} & \multicolumn{1}{l|}{{0.1039}} & {0.1513} & {0.0924} & {0.1128} \\
{Baichuan-Chat-13B} & {0.3130} & {0.1786} & \multicolumn{1}{l|}{{0.2303}} & {0.1723} & {0.0987} & {0.1263} \\
{WizardCoder-15B} & \uwave{0.4496} & \uwave{
 0.2773} & \multicolumn{1}{l|}{\uwave{0.3468}} & \uwave{0.4328} &\uwave{0.2668} & \uwave{0.3329} \\
{StarCoder} & {0.2227} & {0.1366} & \multicolumn{1}{l|}{0.1679} & {0.1870} & {0.0924} & {0.1340} \\
{Vicuna-13B} & {0.2689} & {0.1261} & \multicolumn{1}{l|}{{0.1790}} & {0.3046} & {0.1492} & {0.2045} \\
{GPT3.5-turbo} & {\textbf{0.6429}} & \underline{0.4265} & \multicolumn{1}{l|}{\underline{0.5240}} & \underline{0.6660} & \underline{0.4853} & \underline{0.5644} \\
 GPT4 & \underline{0.6216} & {\textbf{0.4968}} & {\textbf{0.5411}} & {\textbf{0.7672}} & {\textbf{0.6336}} & {\textbf{0.6659}} \\\hline
Human (one-submission) & - &-  & - & 0.7099 & 0.7099 & 0.7099 \\
Human (three-submission) & - &-  & - & 0.9288& 0.9288 & 0.9288 \\
Human (five-submission) & - &-  & - & 0.9766 & 0.9766 & 0.9766 \\\hline
\bottomrule
\end{tabular}
\end{table}

\begin{table}[t]
\caption{Performance on AC@1, AC@all, and AC Rate metrics of code generation task in English version. The best performance of LLMs is in bold, the second-best performance is underlined, and the third-best performance is wave underlined. General-purpose LLMs that are not able to generate code are removed from the table.}
\label{tab:CG-EN}
\centering
\renewcommand{\arraystretch}{0.95}
\begin{tabular}{l|ccc|ccc}
\toprule\hline
\multicolumn{1}{c|}{{}} & \multicolumn{6}{c}{{Code Generation (English)}} \\ \cline{2-7} 
\multicolumn{1}{c|}{{}} & \multicolumn{3}{c|}{{Function-only}} & \multicolumn{3}{c}{{Function-with-Context}} \\ \cline{2-7} 
\multicolumn{1}{c|}{\multirow{-3}{*}{{Model}}} & \multicolumn{1}{c}{{AC@1}} & \multicolumn{1}{c}{{AC@all}} & \multicolumn{1}{c|}{{AC Rate}} & \multicolumn{1}{c}{{AC@1}} & \multicolumn{1}{c}{{AC@all}} & \multicolumn{1}{c}{{AC Rate}} \\ \hline
{ChatGLM-6B} & {0.2080} & {0.0693} & \multicolumn{1}{l|}{{0.1203}} & {0.1828} & {0.0672} & {0.1031} \\
{ChatGLM2-6B} & {0.1891} & {0.0819} & \multicolumn{1}{l|}{{0.1243}} & {0.1029} & {0.0462} & {0.0668} \\
{ChatGLM3-6B} &0.1450	& 0.1176 &	0.1363 & 0.1324 &	0.0987	& 0.1232 \\
{MOSS-16B*} & {0.2626} & {0.1513} & \multicolumn{1}{l|}{{0.2002}} & {0.1534} & {0.0882} & {0.1178} \\
{Chinese-Alpaca-plus-13B} & {0.2878} & {0.1345} & \multicolumn{1}{l|}{{0.1963}} & {0.2542} & {0.1092} & {0.1682} \\
{BELLE-7B-1M} & {0.2038} & {0.0651} & \multicolumn{1}{l|}{{0.1161}} & {0.1618} & {0.0525} & {0.0863} \\
{BELLE-LLaMA-13B-2M} & {0.1870} & {0.0777} & \multicolumn{1}{l|}{{0.1197}} & {0.2227} & {0.0861} & {0.1434} \\
{InternLM-Chat-7B} & {0.0945} & {0.0525} & \multicolumn{1}{l|}{{0.0707}} & {0.3025} & {0.1597} & {0.2126} \\
{Baichuan-Chat-13B} & {0.3298} & {0.1618} & \multicolumn{1}{l|}{{0.2310}} & {0.3319} & {0.1681} & {0.2273} \\
{WizardCoder-15B} & \uwave{0.4244} & \uwave{
 0.2500} & \multicolumn{1}{l|}{\uwave{0.3248}} & \uwave{0.4391} & \uwave{0.2752} & \uwave{0.3444} \\
{StarCoder} & {0.3382} & {0.1891} & \multicolumn{1}{l|}{{0.2494}} & {0.2794} & {0.1723} & {0.2116} \\
{Vicuna-13B} & {0.1870} & {0.0861} & \multicolumn{1}{l|}{{0.1236}} & {0.2983} & {0.1218} & {0.1861} \\
{GPT3.5-turbo} & \textbf{0.6134} & \underline{0.4286} & \multicolumn{1}{l|}{\underline{0.5105}} & \underline{0.6597} & \underline{0.4832} & \underline{0.5606} \\ 
GPT4 & \underline{0.5839} & \textbf{0.4623} & \textbf{0.5176} & \textbf{0.7032} & \textbf{0.5705} & \textbf{0.5992}\\\hline
Human (one-submission) & - &-  & - & 0.7099 & 0.7099 & 0.7099 \\
Human (three-submission) & - &-  & - & 0.9288& 0.9288 & 0.9288 \\
Human (five-submission) & - &-  & - & 0.9766 & 0.9766 & 0.9766 \\\hline
\bottomrule
\end{tabular}
\end{table}

\begin{table}[t]
\caption{Compilable code proportion on each scenario. All models except MOSS-16B* use the completion pattern to generate code. The best performance of LLMs is in bold, the second-best performance is underlined, and the third-best performance is wave underlined.}
\label{tab:CompCode}
\centering
\renewcommand{\arraystretch}{1.0}
\resizebox{\columnwidth}{!}{%
\begin{tabular}{l|c|c|c|c}
\toprule\hline
\multicolumn{1}{c|}{{}} & \multicolumn{4}{c}{{Compilable Code Proportion}} \\ \cline{2-5} 
\multicolumn{1}{c|}{{}} & \multicolumn{2}{c|}{{ZH}} & \multicolumn{2}{c}{{EN}} \\ \cline{2-5} 
\multicolumn{1}{c|}{\multirow{-3}{*}{{Model}}} & \multicolumn{1}{c|}{{Function-only}} & \multicolumn{1}{c|}{{Function-with-Context}} & \multicolumn{1}{c|}{{Function-only}} & \multicolumn{1}{c}{{Function-with-Context}} \\ \hline
{ChatGLM-6B} & {0.5693} & \multicolumn{1}{c|}{{0.5231}} & {0.6429} & {0.5336} \\
{ChatGLM2-6B} & {0.5105} & \multicolumn{1}{c|}{{0.5399}} & {0.5399} & {0.4034} \\
{ChatGLM3-6B} &  {0.6828} & 0.4601 & 0.7353 & 0.6261\\
{MOSS-16B*} & {0.5231} & \multicolumn{1}{c|}{{0.5147}} & {0.6092} & {0.3739} \\
{Chinese-Alpaca-plus-13B} & {0.7164} & \multicolumn{1}{c|}{{0.6744}} & {0.7017} & {0.6723} \\
{BELLE-7B-1M} & {0.4244} & \multicolumn{1}{c|}{{0.4244}} & {0.5273} & {0.4307} \\
{BELLE-LLaMA-13B-2M} & {0.5105} & \multicolumn{1}{c|}{{0.4601}} & {0.4580} & {0.5357} \\
{InternLM-Chat-7B} & {0.3466} & \multicolumn{1}{c|}{{0.4265}} & {0.2017} & {0.7626} \\
{Baichuan-Chat-13B} & {0.6218} & \multicolumn{1}{c|}{{0.3908}} & {0.7605} & {0.7437} \\
{WizardCoder-15B} & \underline{0.8529} & \multicolumn{1}{c|}{\uwave{
   0.8634}} & \underline{0.8361} & \uwave{0.7899} \\
{StarCoder} & {0.4685} & \multicolumn{1}{c|}{{
  0.4853}} & {0.6765} & {0.5945} \\
{Vicuna-13B} & {0.7122} & \multicolumn{1}{c|}{{
   0.7983}} & {0.4496} & {0.7815} \\
{GPT3.5-turbo} & \textbf{0.9118} & \multicolumn{1}{c|}{\underline{
   0.8950}} & \textbf{0.8929} & \textbf{0.8929} \\ 
   GPT4 & \uwave{0.7865} & \textbf{0.9159} & \uwave{0.7883} & \underline{0.8674} \\\hline
   \bottomrule
\end{tabular}}
\end{table}

Code generation task results of all the models are shown in Table \ref{tab:CG-ZH} and Table \ref{tab:CG-EN}.
Two prompt strategies (function-only and function-with-context) are employed for each language version.
The evaluation metrics include AC@1, AC@all, and AC Rate.
GPT outperforms the other LLMs, with the best accepted rate over 66\% (GPT4).
WizardCoder and StarCoder ranked second and third, highlighting the significant improvement in code generation capability through code-based fine-tuning.
There is no noticeable performance difference between the Chinese and English versions.

\paragraph{Effects of Contexts in Prompt.}
As shown in Table~\ref{tab:CG-ZH} and Table~\ref{tab:CG-EN}, \textbf{providing the context of function calls for LLM can effectively enhance the accuracy of generating target function code.}
Meanwhile, Table \ref{tab:CompCode} shows the proportion of compilable code in each scenario.
The majority of models are capable of generating over 50\% of compilable code, which demonstrates the ability of LLM to adhere to the function prototype. 
After concatenating the generated function and the main function, the code that could be compiled can be checked by testcases.
Generally, \textbf{providing contextual information about the function can assist LLMs in generating compilable code}.

\paragraph{Question Category Comparison.}
Table \ref{tab:DNCompare} shows the performance on explicit and narrative questions.
The data is selected from function-only and function-with-context scenarios, with the better-performing data chosen. 
The table demonstrates that \textbf{LLMs perform better on explicit questions than narrative questions}.
This is because explicit questions only require atomic code generation ability of LLMs, while narrative questions require LLM to first understand the question description, convert natural language into program logic, and finally generate code.
It is indeed a significant challenge to extract program logic from natural language, requiring a deep understanding of language semantics and context. 
Meanwhile, three metrics of LLM on narrative questions are mostly consistent, indicating that the generated code is either entirely correct or entirely incorrect.
This may be due to the fact that narrative questions in the dataset are designed for real-world problem-solving, with relatively low requirements for code time and space complexity.

\paragraph{Human Performance}
We extract real student learning records from online platforms to evaluate human performance in code generation tasks. We calculated the pass rates for human submissions in n times (n=1, 3, 5), and the average results are shown in the Table \ref{tab:CG-ZH}, Table~\ref{tab:CG-EN} and Table~\ref{tab:DNCompare}.
Experimental results suggest that \textbf{human performance can outperform all LLMs, and they have a better performance on narrative questions.} This is mainly due to the ease with which human programmers can translate between natural language and code languages.

\begin{table}[t]
\centering
\renewcommand{\arraystretch}{0.95}
\caption{Best performance comparison between explicit and narrative questions on AC@1, AC@all and AC Rate metric of code generation in English version. The best performance of LLMs is in bold, the second-best performance is underlined, and the third-best performance is wave underlined.}
\label{tab:DNCompare}
\begin{tabular}{l|ccc|ccc}
\toprule\hline
\multicolumn{1}{c|}{}   & \multicolumn{6}{c}{Code Generation (English)}       \\ \cline{2-7} 
\multicolumn{1}{c|}{}   & \multicolumn{3}{c|}{Explicit Questions}   & \multicolumn{3}{c}{Narrative Questions}                                       \\ \cline{2-7} 
\multicolumn{1}{c|}{\multirow{-3}{*}{Model}} &  AC@1 &  AC@all & \multicolumn{1}{c|}{AC Rate} & AC@1 &  AC@all & AC Rate \\ \hline 
\multicolumn{1}{l|}{ChatGLM-6B}  & 0.2558 & 0.1008 & \multicolumn{1}{c|}{0.1240} & 0.1560 & 0.1560 & 0.1560  \\
\multicolumn{1}{l|}{ChatGLM2-6B} & 0.2326 & 0.1202 & \multicolumn{1}{c|}{0.1153} & 0.1376 & 0.1376 & 0.1376  \\
\multicolumn{1}{l|}{ChatGLM3-6B} &0.1387 &	0.1092	& 0.1330   & 0.0903	& 0.0651	& 0.0854   \\
\multicolumn{1}{l|}{MOSS-16B*} & 0.3488 & 0.2403 & \multicolumn{1}{c|}{0.2562} & 0.1606 & 0.1606 & 0.1606  \\
\multicolumn{1}{l|}{Chinese-Alpaca-plus-13B} & 0.3876 & 0.2054 & \multicolumn{1}{c|}{0.2384} & 0.1835 & 0.1835 & 0.1835  \\
\multicolumn{1}{l|}{BELLE-7B-1M} & 0.2674 & 0.0930 & \multicolumn{1}{c|}{0.1335} & 0.1330 & 0.1330 & 0.1330  \\
\multicolumn{1}{l|}{BELLE-LLaMA-13B-2M} & 0.2946 & 0.1395 & \multicolumn{1}{c|}{0.1933} & 0.1422 & 0.1422 & 0.1422  \\

\multicolumn{1}{l|}{InternLM-Chat-7B}   & 0.3992 & 0.2287 & \multicolumn{1}{c|}{0.2502} & 0.1927 & 0.1927 & 0.1927  \\

\multicolumn{1}{l|}{Baichuan-Chat-13B} & 0.4147 & 0.2481 & \multicolumn{1}{c|}{0.2680} & 0.2523 & 0.2523 & 0.2523  \\

\multicolumn{1}{l|}{Wizard-Coder-15B}   & \uwave{0.5388} & \uwave{0.3488} & \multicolumn{1}{c|}{\uwave{0.3943}} & \uwave{0.3440} & \uwave{0.3440} & \uwave{0.3440}  \\ 
\multicolumn{1}{l|}{StarCoder}   & 0.4264 & 0.2597 & \multicolumn{1}{c|}{0.2744} & 0.2385 & 0.2385 & 0.2385  \\

\multicolumn{1}{l|}{Vicuna-13B}   & 0.3953 & 0.1822 & \multicolumn{1}{c|}{0.2393} & 0.1881 & 0.1881 & 0.1881  \\

\multicolumn{1}{l|}{GPT3.5-turbo}  &  \underline{0.7054} & \underline{0.5271} & \multicolumn{1}{c|}{\underline{0.5770}} & \underline{0.6193} & \textbf{0.6193} & \textbf{0.6193}  \\
\multicolumn{1}{l|}{GPT4} & \textbf{0.7276}  & \textbf{0.5837}   & \textbf{0.6025}  & \textbf{0.6743}  & \underline{0.5550}  & \underline{0.5959}   \\\hline
  \multicolumn{1}{l|}{ Human (one-submission)} &0.6970 &0.6970 &0.6970 & 0.7457 & 0.7457 & 0.7457\\
  \multicolumn{1}{l|}{ Human (three-submissions)} & 0.9169 & 0.9169& 0.9169&0.9618&0.9618&0.9618 \\
  \multicolumn{1}{l|}{ Human (five-submissions)}& 0.9731& 0.9731& 0.9731&0.9865 &0.9865  &0.9865\\\hline
\bottomrule
\end{tabular}
\end{table}

\begin{table}[t]
\centering
\caption{Performance on AC@1, AC@all, and AC Rate metrics of code generation task in Chinese version. The best performance of LLMs is in bold, the second-best performance is underlined, and the third-best performance is wave underlined. }
\label{tab:CC-ZH}
\renewcommand{\arraystretch}{1.0}
\resizebox{\columnwidth}{!}{%
\begin{tabular}{l|ccc|ccc|ccc}
\toprule
\hline
\multicolumn{1}{c|}{}   & \multicolumn{9}{c}{Code Correction (Chinese)}\\ \cline{2-10} 
\multicolumn{1}{c|}{}   & \multicolumn{3}{c|}{Code-only}  & \multicolumn{3}{c|}{Code-with-Error Message}   & \multicolumn{3}{l}{Code-with-Standard Code}                                                                                                 \\ \cline{2-10} 
\multicolumn{1}{c|}{\multirow{-3}{*}{Model}} & { AC@1} & { AC@n} & \multicolumn{1}{c|}{{ AC Rate}} & { AC@1} & { AC@n} & \multicolumn{1}{c|}{AC Rate} & \multicolumn{1}{c}{{ { AC@1}}} & \multicolumn{1}{c}{{ { AC@n}}} & \multicolumn{1}{c}{AC Rate} \\ \hline
ChatGLM-6B  & 0.1323 & 0.0519 & 0.0899 & 0.1286 & 0.0511 & 0.0851 & 0.1699   & 0.1566 & 0.1086  \\
ChatGLM2-6B & 0.2549 & 0.0902 & 0.1574 & 0.2316 & 0.0887 & 0.1402 & 0.2022   & 0.1654 & 0.1476  \\
ChatGLM3-6B & 0.4293	& 0.2436	& 0.2944 & 0.4414 & 0.2376 & 0.2918 & 0.2917 & 0.2805 & 0.2220  \\
BELLE-7B-1M & 0.0526 & 0.0226 & 0.0345 & 0.0256 & 0.0060 & 0.0156 & 0.0880 & 0.0496 & 0.0724   \\
BELLE-LLaMA-13B-2M   & 0.0647 & 0.0218 & 0.0386 & 0.0496	& 0.0165 & 0.0276 & 0.0429 & 0.0398 & 0.0241   \\

InternLM-Chat-7B  & 0.1150 & 0.0677 & 0.0785 & 0.1271 & 0.0662 & 0.0831 & 0.1474 & 0.1338 & 0.1334   \\
Baichuan-Chat-13B & 0.3654 & 0.2534 & 0.2887 & 0.3782 & 0.2556 & 0.2762 & 0.2376 & 0.2083 & 0.1803 \\
WizardCoder-15B   & 0.5767 & \uwave{0.4331} & 0.4568 & 0.5556 & \uwave{0.4120} & 0.4490 & 0.2556 & 0.2429	& 0.1963   \\
StarCoder   & \underline{0.6639} & 0.4090 & \uwave{0.4869} & \underline{0.6421}	& 0.3767 & \uwave{0.4638} & \underline{0.4737} & \uwave{0.3030} & \underline{0.3452}   \\
GPT3.5-turbo   & \uwave{0.5699} & \underline{0.5466} & \underline{0.4909} & \uwave{0.5722} & \underline{0.5481} & \underline{0.5015} & \uwave{0.3436} & \underline{0.3316} & \uwave{0.2908}   \\ 
GPT4  & \textbf{0.6970}	& \textbf{0.6594}& \textbf{0.6453} & \textbf{0.7143}	& \textbf{0.6767}	& \textbf{0.6677} & \textbf{0.6038} & \textbf{0.5940}	& \textbf{0.5269}  \\ \hline\bottomrule
\end{tabular}
\vspace{-10pt}
}
\end{table}

\begin{table}[t]
\centering
\caption{Performance on AC@1, AC@all, and AC Rate metrics of code generation task in English version. The best performance of LLMs is in bold, the second-best performance is underlined, and the third-best performance is wave underlined. }
\label{tab:CC-EN}
\renewcommand{\arraystretch}{1.0}
\resizebox{\columnwidth}{!}{%
\begin{tabular}{l|ccc|ccc|ccc}
\toprule
\hline
\multicolumn{1}{c|}{}   & \multicolumn{9}{c}{Code Correction (English)}                                                                                                                                                                                                                                                                                                  \\ \cline{2-10} 
\multicolumn{1}{c|}{}   & \multicolumn{3}{c|}{Code-only}  & \multicolumn{3}{c|}{Code-with-Error Message}   & \multicolumn{3}{l}{Code-with-Standard Code}                                                                                                 \\ \cline{2-10} 
\multicolumn{1}{c|}{\multirow{-3}{*}{Model}} & { AC@1} & { AC@n} & \multicolumn{1}{c|}{{ AC Rate}} & { AC@1} & { AC@n} & \multicolumn{1}{c|}{AC Rate} & \multicolumn{1}{c}{{ { AC@1}}} & \multicolumn{1}{c}{{ { AC@n}}} & \multicolumn{1}{c}{AC Rate} \\ \hline
ChatGLM-6B  & 0.1489 & 0.0579 & 0.0950 & 0.0128 & 0.0030 & 0.0088 & 0.1241 & 0.1113 & 0.1056   \\
ChatGLM2-6B & 0.2323 & 0.0850 & 0.1284 & 0.1835 & 0.0797 & 0.1023 & 0.2504 & 0.2113 & 0.1943   \\
ChatGLM3-6B & 0.3692 & 0.2248 & 0.2769 & 0.3022 & 0.1609 & 0.2163 & 0.2105 & 0.2075 & 0.1612     \\
BELLE-7B-1M & 0.1609 & 0.0549 & 0.0959 & 0.0023 & 0.0008 & 0.0007   & 0.1158 & 0.1008 & 0.0914    \\
BELLE-LLaMA-13B-2M   & 0.1105 & 0.0571 & 0.0795 & 0.2150 & 0.0579 & 0.1307 & 0.1692 & 0.1466 & 0.1142  \\

InternLM-Chat-7B  & 0.1850 & 0.1113 & 0.1289 & 0.1150 & 0.0609 & 0.0757 & \uwave{0.2632} & 0.2436 & 0.1978   \\
Baichuan-Chat-13B & 0.2414	& 0.1737 & 0.1963 & 0.0955	& 0.0549 & 0.0582 & 0.1842 & 0.1805 & 0.1398  \\
WizardCoder-15B   & 0.4504 & 0.3331 & 0.3347 & 0.2917 & 0.2316 & 0.2443 & 0.1571	& 0.1391	& 0.1052  \\
StarCoder   & \underline{0.5992} & \uwave{0.3602} & \uwave{0.4326} & \underline{0.5586} & \uwave{0.3113} & \underline{0.4017} & \underline{0.3602} & \underline{0.2782} & \underline{0.2694}  \\
GPT3.5-turbo   & \uwave{0.5083} & \underline{0.4925} & \underline{0.4332} & \uwave{0.3632} & \underline{0.3459} & \uwave{0.3161} & 0.2617  & \uwave{0.2594} & \uwave{0.2003}   \\ 
GPT4  & \textbf{0.6436}	& \textbf{0.6143} & \textbf{0.5775} & \textbf{0.6617} & \textbf{0.6323} & \textbf{0.6022} & \textbf{0.4812} & \textbf{0.4752} & \textbf{0.4032}  \\ \hline\bottomrule
\end{tabular}
}
\vspace{-5pt}
\end{table}

\begin{table}[t]
\caption{Best performance comparison between correcting three types of code error (WA, TLE, RE) on AC@1, AC@all and AC Rate metric of code correction task in Chinese version. The best performance of LLMs is in bold, the second-best performance is underlined, and the third-best performance is wave underlined.}
\label{tab:ErrorType-ZH}
\centering
\renewcommand{\arraystretch}{1.0}
\resizebox{\columnwidth}{!}{%
\begin{tabular}{l|ccccccccc}
\toprule
\hline
\multirow{3}{*}{Models} & \multicolumn{9}{c}{Code Correction (Chinese)}                                                                               \\ \cline{2-10} 
      & \multicolumn{3}{c|}{WA}      & \multicolumn{3}{c|}{TLE}     & \multicolumn{3}{c}{RE}    \\ \cline{2-10} 
      & AC@1   & AC@all & \multicolumn{1}{c|}{AC Rate} & AC@1   & AC@all & \multicolumn{1}{c|}{AC Rate} & AC@1   & AC@all & AC Rate \\ \hline
ChatGLM-6B  & 0.1736 & 0.1591 & \multicolumn{1}{c|}{0.1148}  & 0.2609 & 0.2609 & \multicolumn{1}{c|}{0.1988}  & 0.1512 & 0.1163 & 0.0941  \\
ChatGLM2-6B & 0.2493 & 0.1711 & \multicolumn{1}{c|}{0.1525}  & 0.3623 & 0.1884 & \multicolumn{1}{c|}{0.3854}  & 0.2442 & 0.1744 & 0.1487  \\
ChatGLM3-6B & 0.4468 & 0.2805 & \multicolumn{1}{c|}{0.2826}  & 0.4348 & 0.3768 & \multicolumn{1}{c|}{0.5030}  & 0.3837 & 0.3372 & 0.3383  \\
BELLE-7B-1M & 0.0894 & 0.0519 & \multicolumn{1}{c|}{0.0742}  & 0.1159 & 0.0435 & \multicolumn{1}{c|}{0.1055}  & 0.0465 & 0.0348 & 0.0218  \\
BELLE-LLaMA-13B-2M   & 0.0655 & 0.0357 & \multicolumn{1}{c|}{0.0379}  & 0.0870 & 0.0580 & \multicolumn{1}{c|}{0.1034}  & 0.1279 & 0.1279 & 0.0750  \\
InternLM-Chat-7B  & 0.1498 & 0.1362 & \multicolumn{1}{c|}{0.1371}  & 0.1594 & 0.1594 & \multicolumn{1}{c|}{0.1116}  & 0.1046 & 0.0814 & 0.1214  \\
Baichuan-13B-Chat & 0.3728 & 0.2477 & \multicolumn{1}{c|}{0.2837}  & \underline{0.5362} & \underline{0.4348} & \multicolumn{1}{c|}{\underline{0.5639}}  & 0.3256 & 0.2674 & 0.2606  \\
WizardCoder-15B   & \uwave{0.5864} & \uwave{0.4264} & \multicolumn{1}{c|}{0.4487}  & 0.4493 & \uwave{0.4058} & \multicolumn{1}{c|}{\uwave{0.5497}}  & \underline{0.5698} & \uwave{0.5465} & \uwave{0.5498}  \\
StarCoder   & \underline{0.6843} & 0.4162 & \multicolumn{1}{c|}{\underline{0.4934}}  & \uwave{0.4783} & 0.2319 & \multicolumn{1}{c|}{0.4361}  & \uwave{0.5349} & 0.4535 & 0.4188  \\
GPT-3.5-turbo  & 0.5668 & \underline{0.5405} & \multicolumn{1}{c|}{\uwave{0.4932}}  & 0.4493 & \underline{0.4348} & \multicolumn{1}{c|}{0.4077}  & \underline{0.7791} & \underline{0.7791} & \underline{0.7162}  \\
GPT-4 & \textbf{0.6996} & \textbf{0.6612} & \multicolumn{1}{c|}{\textbf{0.6546}}  & \textbf{0.8261} & 
\textbf{0.7391} & \multicolumn{1}{c|}{\textbf{0.7667}}  & \textbf{0.8605} & \textbf{0.8488} & \textbf{0.8240}  \\ \hline\bottomrule
\end{tabular}
}
\end{table}

\subsection{Code Correction Results}
The overall performance of LLMs in the code correction task is shown in Table \ref{tab:CC-ZH} and \ref{tab:CC-EN}.
There is a significant difference in debugging ability among different models.
GPT-4 performs the best among all LLMs with the best accuracy of 66\%, followed by GPT3.5-turbo, StarCoder and WizardCoder. 
Baichuan and ChatGLM3 could correct over 25\% erroneous code, which is equipped with code correction capability.
Other models perform poorly in code correction tasks, with a success rate of less than 20\%.
The code correction capability of LLMs still has a lot of room for improvement.

\paragraph{Feedback Analysis}
\textbf{Providing error types does not significantly help LLMs in the code correction task.}
This may be because the online judge platform only provides three types of error without specific line numbers or other detailed error messages, which is not sufficient to assist LLMs in code correction.
\textbf{LLMs that are unable to complete code correction tasks can exhibit code correction capability after being provided with standard code.}
This is because the standard code serves as a reference point or template, guiding the model to make corrections based on the provided code structure and logic.


\paragraph{Error Type Analysis.}
We compare the performance on three types of error (WA, TLE, and RE) in Table~\ref{tab:ErrorType-ZH} and Table~\ref{tab:ErrorType-EN}.
Results shows correcting these three types of errors has different emergence ability of LLMs.
Correcting wrong answer codes is the hardest task for LLMs since errors might occur everywhere.
TLE usually occurs in closed loops or too many loops, which is relatively easy to locate.
RE is also easy to locate, usually because the divisor is 0, the subscript is out of bounds, and so on.
Thus when parameter size is limited (StarCoder, GPT3.5), RE and TLE are easy to find, while the  ability to find WA error emergents until the parameter size of GPT4.

\begin{table}[thb]
\caption{Best performance comparison between correcting three types of code error (WA, TLE, RE) on AC@1, AC@all and AC Rate metric of code correction task in English version. The best performance of LLMs is in bold, the second-best performance is underlined, and the third-best performance is wave underlined. }
\label{tab:ErrorType-EN}
\centering
\renewcommand{\arraystretch}{1.0}
\resizebox{\columnwidth}{!}{%
\begin{tabular}{l|ccc|ccc|ccc}
\toprule
\hline
\multirow{3}{*}{Models} & \multicolumn{9}{c}{Code Correction (English)}                                                                               \\ \cline{2-10} 
      & \multicolumn{3}{c|}{WA}      & \multicolumn{3}{c|}{TLE}     & \multicolumn{3}{c}{RE}    \\ \cline{2-10} 
      & AC@1   & AC@all & \multicolumn{1}{l|}{AC Rate} & AC@1   & AC@all & \multicolumn{1}{l|}{AC Rate} & AC@1   & AC@all & AC Rate \\\hline
ChatGLM-6B  & 0.1506 & 0.1038 & \multicolumn{1}{l|}{0.1078}  & 0.2319 & 0.2319 & \multicolumn{1}{l|}{0.1379}  & 0.1395 & 0.1163 & 0.0941  \\
ChatGLM2-6B & 0.2647 & 0.2213 & \multicolumn{1}{l|}{0.2011}  & 0.2609 & 0.1159 & \multicolumn{1}{l|}{0.2840}  & 0.2093 & 0.2093 & 0.1910  \\
ChatGLM3-6B & 0.3745 & 0.2187 & \multicolumn{1}{l|}{0.2726}  & 0.3188 & 0.2464 & \multicolumn{1}{l|}{0.3773}  & 0.3372 & 0.2907 & 0.2783  \\
BELLE-7B-1M & 0.1591 & 0.1089 & \multicolumn{1}{l|}{0.0968}  & 0.1739 & 0.0725 & \multicolumn{1}{l|}{0.1318}  & 0.1744 & 0.1512 & 0.1596  \\
BELLE-LLaMA-13B-2M   & 0.2204 & 0.1404 & \multicolumn{1}{l|}{0.1300}  & 0.2754 & 0.2464 & \multicolumn{1}{l|}{0.2738}  & 0.1511 & 0.1511 & 0.1091  \\
InternLM-Chat-7B  & 0.2570 & 0.2383 & \multicolumn{1}{l|}{0.2010}  &  0.3623 & \uwave{0.3623} & \multicolumn{1}{l|}{0.1318}  & 0.2674 & 0.2209 & 0.1924  \\
Baichuan-13B-Chat & 0.2289 & 0.1821 & \multicolumn{1}{l|}{0.1844}  & 0.3913 & 0.2754 & \multicolumn{1}{l|}{\uwave{0.4118}}  & 0.2907 & 0.2326 & 0.2374  \\
WizardCoder & 0.4553 & 0.3277 & \multicolumn{1}{l|}{0.3241}  & 0.3913 & 0.2754 & \multicolumn{1}{l|}{\uwave{0.4118}}  & 0.4302 & 0.4070 & 0.3861  \\
StarCoder   & \underline{0.6136} & \uwave{0.3617} & \multicolumn{1}{l|}{\underline{0.4349}}  & \underline{0.5217} & 0.2899 & \multicolumn{1}{l|}{\underline{0.4178}}  & \uwave{0.5233} & \uwave{0.4535} & \uwave{0.4052}  \\
GPT-3.5-turbo  & \uwave{0.4945} & \underline{0.4783} & \multicolumn{1}{l|}{\uwave{0.4171}}  & \uwave{0.4638} & \underline{0.4348} & \multicolumn{1}{l|}{\underline{0.4178}}  & \underline{0.7326} & \underline{0.7326} & \underline{0.6958} \\
GPT-4 & \textbf{0.6460} & \textbf{0.6170} & \multicolumn{1}{l|}{\textbf{0.5837}}  & \textbf{0.7391} & \textbf{0.6667} & \multicolumn{1}{l|}{\textbf{0.7079}}  & \textbf{0.8372} & \textbf{0.8372} & \textbf{0.8213}  \\ \hline\bottomrule
\end{tabular}
}
\vspace{-10pt}
\end{table}

%% file: conclusion.tex
\section{Conclusion}\label{conclusion}
We present \model, a bilingual benchmark focusing on the programming capabilities of LLMs.
First, we evaluate LLMs in the programming comprehension task by assessing their accuracy in three categories of multiple-choice questions. 
We design two types of prompts, answer-only and chain-of-thought, with three in-context learning scenarios, to compare LLMs' performance.
Second, we evaluate LLMs' code generation capabilities by testing the acceptance rate of test cases.
The test set includes 476 C++-based coding questions, categorized as either explicit questions or narrative questions.
We also provide a comparison on function-only and function-with-context scenarios, in order to compare the impact of context information of the target code.
Third, we introduce 1332 real-world erroneous code segments to evaluate LLMs' code correction capability.
We compare the performance in code-only, code-with-error message, and code-with-standard code scenarios, and analysis the result of different code error types.
We organize experiments on 12 famous LLMs, including general-purpose LLMs and specialized models based on code fine-tuning.
Currently, GPT achieves top-tier performance in programming abilities, but human tests indicate that there is still significant room for improvement. 
Furthermore, specialized LLMs demonstrate competitiveness and show their potential for further enhancement.

We also incorporate human performance into the leaderboard, comparing the coding abilities of the \model with human-level performance. 
The experiments indicate that the programming capabilities of LLMs still have some distance from human performance.
In the future, the test set within \model will undergo continuous updates, encompassing an expanded range of question types and additional programming languages.
We hope that \model could provide a reference for later study on the programming capabilities of LLMs, fostering their development and prosperity together. 

%% file: appendix.tex
\section{Breakdown Performance in Programming Comprehension}\label{app:x-shot}
Table~\ref{table:shot-Ao-ZH}, Table~\ref{table:shot-Ao-EN}, Table~\ref{table:shot-CoT-ZH}, and Table~\ref{table:shot-CoT-EN} present the breakdown of performance of LLMs in programming comprehension task in the 5-shot, 2-shot, and 0-shot scenarios. 

\begin{table}[ht]
\centering
\caption{5-shot, 2-shot, and 0-shot accuracy results of Chinese version Programming Comprehension Task in the answer-only setting.  C.U., C.R,
and M.H.R indicate conceptual understanding, commonsense reasoning, and multi-hop reasoning
questions, respectively. * represents chat-bot mode of the LLM. - indicates that the model does not have the ability to complete the test in that setting.}
\label{table:shot-Ao-ZH}
\renewcommand{\arraystretch}{1}
\resizebox{\columnwidth}{!}{%
\begin{tabular}{l|cccccccccccc}
\toprule
\hline
 & \multicolumn{12}{c}{{Programming Comprehension (Answer-only)-ZH}} \\ \cline{2-13} 
 & \multicolumn{3}{c|}{{C.U.}} & \multicolumn{3}{c|}{{C.R.}} & \multicolumn{3}{c|}{{M.H.R.}} & \multicolumn{3}{c}{{Total}} \\ \cline{2-13} 
\multirow{-3}{*}{Model} & {{5-shot}} & {{2-shot}} & \multicolumn{1}{c|}{{{0-shot}}} & {{5-shot}} & {{2-shot}} & \multicolumn{1}{c|}{{{0-shot}}} & {{5-shot}} & {{2-shot}} & \multicolumn{1}{c|}{{{0-shot}}} & {{5-shot}} & {{2-shot}} & {{0-shot}} \\ \hline
{ChatGLM-6B} & {0.32} & {0.30} & \multicolumn{1}{c|}{{0.32}} & {0.16} & 0.12 & \multicolumn{1}{c|}{0.26} & 0.30 & 0.26 & \multicolumn{1}{c|}{0.31} & 0.25 & 0.22 & 0.30 \\
{ChatGLM-6B*} & {0.31} & {0.33} & \multicolumn{1}{c|}{{0.31}} & {0.27} & 0.25 & \multicolumn{1}{c|}{0.16} & 0.25 & 0.31 & \multicolumn{1}{c|}{0.23} & 0.28 & 0.30 & 0.23 \\
{ChatGLM2-6B} & {0.32} & {0.33} & \multicolumn{1}{c|}{{0.37}} & {0.25} & 0.27 & \multicolumn{1}{c|}{0.21} & 0.30 & 0.28 & \multicolumn{1}{c|}{0.34} & 0.29 & 0.30 & 0.30 \\
{ChatGLM2-6B*} & {0.34} & {0.31} & \multicolumn{1}{c|}{{0.32}} & {0.26} & 0.26 & \multicolumn{1}{c|}{0.19} & 0.31 & 0.31 & \multicolumn{1}{c|}{0.23} & 0.30 & 0.29 & 0.25 \\

{ChatGLM3-6B} & {0.29} & {0.30} & \multicolumn{1}{c|}{{0.34}} & {0.24} & 0.26 & \multicolumn{1}{c|}{0.24} & 0.38 & 0.34 & \multicolumn{1}{c|}{0.34} & 0.29 & 0.30 & 0.30 \\

{ChatGLM3-6B*} & {0.23} & {0.35} & \multicolumn{1}{c|}{{0.46}} & {0.21} & 0.27 & \multicolumn{1}{c|}{0.19} & 0.25 & 0.17 & \multicolumn{1}{c|}{0.28} & 0.23 & 0.28 & 0.31 \\

{MOSS-16B*} & {0.31} & {0.30} & \multicolumn{1}{c|}{{0.28}} & {0.24} & 0.28 & \multicolumn{1}{c|}{{0.34}} & 0.33 & 0.25 & \multicolumn{1}{c|}{0.32} & 0.29 & 0.28 & 0.31 \\
{Chinese-Alpaca-7B} & {0.21} & {0.26} & \multicolumn{1}{c|}{{0.35}} & {0.27} & 0.29 & \multicolumn{1}{c|}{0.25} & 0.18 & 0.27 & \multicolumn{1}{c|}{0.26} & 0.23 & 0.27 & 0.29 \\
{Chinese-Alpaca-7B*} & {0.24} & {0.24} & \multicolumn{1}{c|}{{0.30}} & {0.29} & 0.29 & \multicolumn{1}{c|}{0.27} & 0.21 & 0.15 & \multicolumn{1}{c|}{0.28} & 0.25 & 0.24 & 0.28 \\
{Chinese-Alpaca-plus-7B} & {0.21} & {0.19} & \multicolumn{1}{c|}{{0.23}} & {0.30} & {0.34} & \multicolumn{1}{c|}{0.30} & 0.14 & 0.16 & \multicolumn{1}{c|}{0.26} & 0.23 & 0.24 & 0.27 \\
Chinese-Alpaca-plus-7B* & 0.33 & 0.20 & \multicolumn{1}{c|}{0.28} & 0.29 & 0.30 & \multicolumn{1}{c|}{0.16} & 0.21 & 0.25 & \multicolumn{1}{c|}{0.25} & 0.29 & 0.25 & 0.22 \\
Chinese-Alpaca-13B & 0.24 & 0.21 & \multicolumn{1}{c|}{0.26} & 0.23 & 0.20 & \multicolumn{1}{c|}{0.29} & 0.15 & 0.20 & \multicolumn{1}{c|}{0.26} & 0.22 & 0.20 & 0.27 \\
Chinese-Alpaca-13B* & 0.23 & 0.27 & \multicolumn{1}{c|}{0.27} & 0.23 & 0.29 & \multicolumn{1}{c|}{0.24} & 0.21 & 0.20 & \multicolumn{1}{c|}{0.18} & 0.23 & 0.26 & 0.24 \\
Chinese-Alpaca-plus-13B & 0.27 & 0.24 & \multicolumn{1}{c|}{0.24} & 0.34 & 0.28 & \multicolumn{1}{c|}{0.20} & 0.21 & 0.20 & \multicolumn{1}{c|}{0.19} & 0.28 & 0.25 & 0.21 \\
Chinese-Alpaca-plus-13B* & 0.27 & 0.23 & \multicolumn{1}{c|}{0.27} & 0.27 & 0.30 & \multicolumn{1}{c|}{0.28} & 0.13 & 0.20 & \multicolumn{1}{c|}{0.20} & 0.23 & 0.25 & 0.26 \\
BELLE-7B-1M & 0.27 & 0.26 & \multicolumn{1}{c|}{0.33} & 0.32 & 0.25 & \multicolumn{1}{c|}{0.28} & 0.28 & 0.25 & \multicolumn{1}{c|}{0.26} & 0.29 & 0.25 & 0.29 \\
BELLE-7B-1M* & 0.35 & 0.31 & \multicolumn{1}{c|}{0.26} & 0.31 & 0.29 & \multicolumn{1}{c|}{0.26} & 0.21 & 0.21 & \multicolumn{1}{c|}{0.25} & 0.30 & 0.28 & 0.26 \\
BELLE-7B-2M & 0.20 & 0.22 & \multicolumn{1}{c|}{0.24} & 0.32 & {0.34} & \multicolumn{1}{c|}{0.26} & 0.28 & 0.26 & \multicolumn{1}{c|}{0.19} & 0.27 & 0.28 & 0.24 \\
BELLE-7B-2M* & 0.20 & 0.21 & \multicolumn{1}{c|}{0.20} & 0.32 & 0.30 & \multicolumn{1}{c|}{0.25} & 0.18 & 0.15 & \multicolumn{1}{c|}{0.28} & 0.24 & 0.23 & 0.24 \\
BELLE-LLaMA-7B-0.6M & 0.29 & 0.24 & \multicolumn{1}{c|}{0.14} & 0.33 & {0.34} & \multicolumn{1}{c|}{0.27} & 0.28 & 0.32 & \multicolumn{1}{c|}{0.21} & 0.31 & 0.30 & 0.21 \\
BELLE-LLaMA-7B-0.6M* & 0.28 & 0.28 & \multicolumn{1}{c|}{0.28} & 0.30 & 0.31 & \multicolumn{1}{c|}{0.31} & 0.30 & 0.31 & \multicolumn{1}{c|}{0.28} & 0.29 & 0.30 & 0.29 \\
BELLE-LLaMA-7B-2M & 0.17 & 0.22 & \multicolumn{1}{c|}{0.20} & 0.30 & 0.29 & \multicolumn{1}{c|}{0.28} & 0.16 & 0.17 & \multicolumn{1}{c|}{0.20} & 0.22 & 0.24 & 0.23 \\
BELLE-LLaMA-7B-2M* & 0.19 & 0.28 & \multicolumn{1}{c|}{0.23} & 0.21 & 0.30 & \multicolumn{1}{c|}{0.31} & 0.09 & 0.26 & \multicolumn{1}{c|}{0.20} & 0.17 & 0.28 & 0.25 \\
BELLE-LLaMA-13B-2M & 0.19 & 0.21 & \multicolumn{1}{c|}{0.29} & 0.31 & 0.31 & \multicolumn{1}{c|}{{0.33}} & 0.18 & 0.19 & \multicolumn{1}{c|}{0.20} & 0.24 & 0.25 & 0.28 \\
BELLE-LLaMA-13B-2M* & 0.34 & 0.24 & \multicolumn{1}{c|}{0.19} & 0.27 & 0.29 & \multicolumn{1}{c|}{0.27} & 0.22 & 0.21 & \multicolumn{1}{c|}{0.21} & 0.28 & 0.25 & 0.23 \\
InternLM-Chat-7B & 0.46 & 0.42 & \multicolumn{1}{c|}{0.44} & 0.22 & 0.27 & \multicolumn{1}{c|}{{0.34}} & {0.34} & 0.33 & \multicolumn{1}{c|}{0.31} & 0.34 & 0.34 & 0.37 \\
Baichuan-7B & 0.34 & 0.24 & \multicolumn{1}{c|}{0.34} & 0.26 & 0.24 & \multicolumn{1}{c|}{{0.34}} & 0.26 & 0.25 & \multicolumn{1}{c|}{0.24} & 0.29 & 0.25 & 0.31 \\
EduChat-base-002-7B* & 0.19 & 0.23 & \multicolumn{1}{c|}{0.18} & 0.31 & 0.22 & \multicolumn{1}{c|}{0.14} & 0.23 & 0.20 & \multicolumn{1}{c|}{0.16} & 0.25 & 0.22 & 0.16 \\
EduChat-base-002-13B* & 0.28 & 0.25 & \multicolumn{1}{c|}{0.17} & 0.35 & {0.34} & \multicolumn{1}{c|}{0.03} & 0.25 & 0.22 & \multicolumn{1}{c|}{0.13} & 0.30 & 0.28 & 0.10 \\
EduChat-sft-002-7B* & 0.22 & 0.24 & \multicolumn{1}{c|}{0.17} & 0.19 & 0.24 & \multicolumn{1}{c|}{0.15} & 0.25 & 0.30 & \multicolumn{1}{c|}{0.27} & 0.21 & 0.26 & 0.19 \\
CodeT5-plus-16B & 0.16 & 0.17 & \multicolumn{1}{c|}{0.26} & 0.28 & 0.29 & \multicolumn{1}{c|}{0.30} & 0.32 & 0.26 & \multicolumn{1}{c|}{0.21} & 0.25 & 0.24 & 0.26 \\
CodeT5-plus-16B* & 0.28 & 0.18 & \multicolumn{1}{c|}{-} & {0.37} & 0.26 & \multicolumn{1}{c|}{-} & 0.27 & 0.23 & \multicolumn{1}{c|}{-} & 0.32 & 0.22 & - \\
CodeT5-plus-6B & 0.20 & 0.19 & \multicolumn{1}{c|}{0.23} & 0.35 & {0.35} & \multicolumn{1}{c|}{0.19} & 0.24 & 0.23 & \multicolumn{1}{c|}{0.18} & 0.27 & 0.26 & 0.20 \\
CodeT5-plus-6B* & 0.23 & 0.21 & \multicolumn{1}{c|}{-} & 0.28 & 0.23 & \multicolumn{1}{c|}{-} & 0.26 & 0.18 & \multicolumn{1}{c|}{-} & 0.26 & 0.21 & - \\
GPT-3.5-turbo & {0.58} & {0.56} & \multicolumn{1}{c|}{{0.61}} & 0.30 & 0.31 & \multicolumn{1}{c|}{{0.33}} & {0.61} & {0.59} & \multicolumn{1}{c|}{{0.57}} & {0.47} & {0.47} & {0.49} \\
GPT-3.5-turbo* & {0.60} & {0.57} & \multicolumn{1}{c|}{{0.57}} & {0.36} & {0.35} & \multicolumn{1}{c|}{0.27} & {0.61} & {0.64} & \multicolumn{1}{c|}{{0.54}} & {0.51} & {0.50} & {0.44} \\

GPT-4   & 0.67  &  0.64 & \multicolumn{1}{c|}{ 0.63 } & 0.55  & 0.54  & \multicolumn{1}{c|}{ 0.49 } & 0.84  & 0.82  & \multicolumn{1}{c|}{ 0.80 } & 0.66  & 0.64  & 0.62  \\
GPT-4*  & 0.70  & 0.67  & \multicolumn{1}{c|}{ 0.66 } &  0.55 &  0.48 & \multicolumn{1}{c|}{ 0.55 } & 0.82  & 0.82  & \multicolumn{1}{c|}{ 0.77 } & 0.67  &  0.63 & 0.64  \\ \hline

\bottomrule
\end{tabular}}
\end{table}

\begin{table}[ht]
\centering
\caption{5-shot, 2-shot, and 0-shot accuracy results of Chinese version Programming Comprehension Task in CoT setting.  C.U., C.R,
and M.H.R indicate conceptual understanding, commonsense reasoning, and multi-hop reasoning
questions, respectively.  * represents chat-bot mode of the LLM. - indicates that the model does not have the ability to complete the test in that setting.}
\label{table:shot-Ao-EN}
\renewcommand{\arraystretch}{1.0}
\resizebox{\columnwidth}{!}{%
\begin{tabular}{l|cccccccccccc}
\toprule
\hline
 & \multicolumn{12}{c}{{Programming Comprehension (Chain-of-Thought)-ZH}} \\ \cline{2-13} 
 & \multicolumn{3}{c|}{{C.U.}} & \multicolumn{3}{c|}{{C.R.}} & \multicolumn{3}{c|}{{M.H.R.}} & \multicolumn{3}{c}{{Total}} \\ \cline{2-13} 
\multirow{-3}{*}{Model} & {{5-shot}} & {{2-shot}} & \multicolumn{1}{c|}{{{0-shot}}} & {{5-shot}} & {{2-shot}} & \multicolumn{1}{c|}{{{0-shot}}} & {{5-shot}} & {{2-shot}} & \multicolumn{1}{c|}{{{0-shot}}} & {{5-shot}} & {{2-shot}} & {{0-shot}} \\ \hline
{ChatGLM-6B} & {0.31} & {0.27} & \multicolumn{1}{c|}{{0.22}} & {0.21} & 0.16 & \multicolumn{1}{c|}{0.18} & 0.38 & 0.26 & \multicolumn{1}{c|}{0.23} & 0.29 & 0.22 & 0.21 \\
{ChatGLM-6B*} & {0.33} & {0.26} & \multicolumn{1}{c|}{{0.27}} & {0.20} & 0.22 & \multicolumn{1}{c|}{0.17} & 0.18 & 0.28 & \multicolumn{1}{c|}{0.28} & 0.24 & 0.25 & 0.23 \\
{ChatGLM2-6B} & {0.48} & {0.36} & \multicolumn{1}{c|}{{0.30}} & {0.23} & 0.25 & \multicolumn{1}{c|}{0.20} & 0.25 & 0.31 & \multicolumn{1}{c|}{0.28} & 0.32 & 0.30 & 0.26 \\
{ChatGLM2-6B*} & {0.41} & {0.32} & \multicolumn{1}{c|}{{0.30}} & {0.24} & 0.29 & \multicolumn{1}{c|}{0.26} & 0.38 & 0.23 & \multicolumn{1}{c|}{0.21} & 0.34 & 0.29 & 0.26 \\

{ChatGLM3-6B} & {0.33} & {0.34} & \multicolumn{1}{c|}{{0.36}} & {0.26} & 0.34 & \multicolumn{1}{c|}{0.24} & 0.25 & 0.24 & \multicolumn{1}{c|}{0.22} & 0.28 & 0.28 & 0.28 \\

{ChatGLM3-6B*} & {0.34} & {0.26} & \multicolumn{1}{c|}{{0.35}} & {0.22} & 0.26 & \multicolumn{1}{c|}{0.23} & 0.27 & 0.35 & \multicolumn{1}{c|}{0.22} & 0.27 & 0.25 & 0.27 \\

{MOSS-16B*} & {0.09} & {0.27} & \multicolumn{1}{c|}{{0.27}} & {0.09} & 0.12 & \multicolumn{1}{c|}{0.17} & 0.23 & 0.33 & \multicolumn{1}{c|}{0.20} & 0.12 & 0.22 & 0.21 \\
{Chinese-Alpaca-7B} & {0.28} & {0.29} & \multicolumn{1}{c|}{{0.21}} & {0.25} & 0.30 & \multicolumn{1}{c|}{0.10} & 0.20 & 0.18 & \multicolumn{1}{c|}{0.21} & 0.25 & 0.27 & 0.17 \\
{Chinese-Alpaca-7B*} & {0.28} & {0.24} & \multicolumn{1}{c|}{{0.27}} & {0.24} & 0.32 & \multicolumn{1}{c|}{0.12} & 0.16 & 0.20 & \multicolumn{1}{c|}{0.26} & 0.24 & 0.26 & 0.21 \\
{Chinese-Alpaca-plus-7B} & {0.20} & {0.37} & \multicolumn{1}{c|}{{0.23}} & {0.27} & 0.31 & \multicolumn{1}{c|}{0.20} & 0.18 & 0.28 & \multicolumn{1}{c|}{0.18} & 0.22 & 0.32 & 0.21 \\
Chinese-Alpaca-plus-7B* & 0.27 & 0.27 & \multicolumn{1}{c|}{0.22} & 0.26 & 0.24 & \multicolumn{1}{c|}{0.24} & 0.15 & 0.20 & \multicolumn{1}{c|}{0.23} & 0.24 & 0.24 & 0.23 \\
Chinese-Alpaca-13B & 0.29 & 0.23 & \multicolumn{1}{c|}{0.27} & 0.29 & 0.20 & \multicolumn{1}{c|}{0.20} & 0.18 & 0.23 & \multicolumn{1}{c|}{0.25} & 0.26 & 0.22 & 0.24 \\
Chinese-Alpaca-13B* & 0.28 & 0.27 & \multicolumn{1}{c|}{0.31} & 0.27 & 0.23 & \multicolumn{1}{c|}{0.17} & 0.23 & 0.16 & \multicolumn{1}{c|}{0.13} & 0.26 & 0.23 & 0.21 \\
Chinese-Alpaca-plus-13B & 0.24 & 0.26 & \multicolumn{1}{c|}{0.31} & 0.25 & 0.29 & \multicolumn{1}{c|}{0.18} & 0.13 & 0.21 & \multicolumn{1}{c|}{0.18} & 0.22 & 0.26 & 0.23 \\
Chinese-Alpaca-plus-13B* & 0.29 & 0.24 & \multicolumn{1}{c|}{0.36} & 0.25 & 0.24 & \multicolumn{1}{c|}{0.27} & 0.15 & 0.23 & \multicolumn{1}{c|}{0.23} & 0.24 & 0.24 & 0.29 \\
BELLE-7B-1M & 0.33 & 0.30 & \multicolumn{1}{c|}{0.26} & 0.23 & 0.29 & \multicolumn{1}{c|}{0.19} & 0.25 & 0.20 & \multicolumn{1}{c|}{0.18} & 0.27 & 0.27 & 0.21 \\
BELLE-7B-1M* & 0.33 & 0.28 & \multicolumn{1}{c|}{0.18} & 0.22 & 0.25 & \multicolumn{1}{c|}{0.10} & 0.21 & 0.16 & \multicolumn{1}{c|}{0.15} & 0.26 & 0.24 & 0.14 \\
BELLE-7B-2M & 0.20 & 0.20 & \multicolumn{1}{c|}{0.29} & 0.24 & 0.26 & \multicolumn{1}{c|}{0.18} & 0.23 & 0.18 & \multicolumn{1}{c|}{0.11} & 0.22 & 0.22 & 0.20 \\
BELLE-7B-2M* & 0.26 & 0.20 & \multicolumn{1}{c|}{0.27} & 0.23 & 0.24 & \multicolumn{1}{c|}{0.23} & 0.15 & 0.21 & \multicolumn{1}{c|}{0.21} & 0.22 & 0.22 & 0.24 \\
BELLE-LLaMA-7B-0.6M & 0.27 & 0.23 & \multicolumn{1}{c|}{0.23} & 0.21 & 0.30 & \multicolumn{1}{c|}{0.28} & 0.16 & 0.30 & \multicolumn{1}{c|}{0.18} & 0.22 & 0.28 & 0.24 \\
BELLE-LLaMA-7B-0.6M* & 0.28 & 0.28 & \multicolumn{1}{c|}{0.34} & 0.20 & 0.36 & \multicolumn{1}{c|}{0.28} & 0.21 & 0.31 & \multicolumn{1}{c|}{0.20} & 0.23 & 0.32 & 0.28 \\
BELLE-LLaMA-7B-2M & 0.24 & 0.19 & \multicolumn{1}{c|}{0.12} & 0.30 & 0.19 & \multicolumn{1}{c|}{0.14} & 0.23 & 0.21 & \multicolumn{1}{c|}{0.15} & 0.26 & 0.20 & 0.14 \\
BELLE-LLaMA-7B-2M* & 0.28 & 0.28 & \multicolumn{1}{c|}{0.17} & 0.26 & 0.30 & \multicolumn{1}{c|}{0.12} & 0.25 & 0.25 & \multicolumn{1}{c|}{0.13} & 0.26 & 0.28 & 0.14 \\
BELLE-LLaMA-13B-2M & 0.22 & 0.29 & \multicolumn{1}{c|}{0.20} & 0.19 & 0.27 & \multicolumn{1}{c|}{0.19} & 0.20 & 0.18 & \multicolumn{1}{c|}{0.16} & 0.20 & 0.26 & 0.19 \\
BELLE-LLaMA-13B-2M* & 0.21 & 0.27 & \multicolumn{1}{c|}{0.17} & 0.22 & 0.28 & \multicolumn{1}{c|}{0.15} & 0.18 & 0.21 & \multicolumn{1}{c|}{0.11} & 0.21 & 0.26 & 0.15 \\
InternLM-Chat-7B & 0.36 & 0.32 & \multicolumn{1}{c|}{0.34} & 0.18 & 0.18 & \multicolumn{1}{c|}{0.14} & 0.36 & 0.20 & \multicolumn{1}{c|}{0.23} & 0.29 & 0.24 & 0.24 \\
Baichuan-7B & 0.04 & 0.06 & \multicolumn{1}{c|}{0.07} & 0.11 & 0.02 & \multicolumn{1}{c|}{0.04} & 0.05 & 0.02 & \multicolumn{1}{c|}{0.07} & 0.07 & 0.03 & 0.06 \\
EduChat-base-002-7B* & 0.28 & 0.27 & \multicolumn{1}{c|}{0.11} & 0.23 & 0.22 & \multicolumn{1}{c|}{0.21} & 0.18 & 0.26 & \multicolumn{1}{c|}{0.15} & 0.24 & 0.25 & 0.16 \\
EduChat-base-002-13B* & 0.30 & 0.29 & \multicolumn{1}{c|}{0.18} & 0.23 & 0.24 & \multicolumn{1}{c|}{0.18} & 0.18 & 0.33 & \multicolumn{1}{c|}{0.16} & 0.24 & 0.28 & 0.18 \\
EduChat-sft-002-7B* & 0.31 & 0.24 & \multicolumn{1}{c|}{0.27} & 0.20 & 0.21 & \multicolumn{1}{c|}{0.19} & 0.25 & 0.23 & \multicolumn{1}{c|}{0.26} & 0.25 & 0.23 & 0.24 \\
CodeT5-plus-16B & - & - & \multicolumn{1}{c|}{-} & - & - & \multicolumn{1}{c|}{-} & - & - & \multicolumn{1}{c|}{-} & - & - & - \\
CodeT5-plus-16B* & - & - & \multicolumn{1}{c|}{-} & - & - & \multicolumn{1}{c|}{-} & - & - & \multicolumn{1}{c|}{-} & - & - & - \\
CodeT5-plus-6B & - & - & \multicolumn{1}{c|}{-} & - & - & \multicolumn{1}{c|}{-} & - & - & \multicolumn{1}{c|}{-} & - & - & - \\
CodeT5-plus-6B* & - & - & \multicolumn{1}{c|}{-} & - & - & \multicolumn{1}{c|}{-} & - & - & \multicolumn{1}{c|}{-} & - & - & - \\
GPT-3.5-turbo & {0.59} & {0.55} & \multicolumn{1}{c|}{{0.50}} & {0.37} & {0.41} & \multicolumn{1}{c|}{{0.35}} & {0.68} & {0.63} & \multicolumn{1}{c|}{{0.52}} & {0.53} & {0.52} & {0.45} \\
GPT-3.5-turbo* & {0.57} & {0.56} & \multicolumn{1}{c|}{{0.54}} & {0.39} & {0.42} & \multicolumn{1}{c|}{{0.33}} & {0.65} & {0.57} & \multicolumn{1}{c|}{{0.56}} & {0.52} & {0.51} & {0.46} \\ 

GPT-4   & 0.71   & 0.71    & \multicolumn{1}{c|}{ 0.63  } & 0.52   & 0.52   & \multicolumn{1}{c|}{ 0.51 } &   &   & \multicolumn{1}{c|}{  } & 0.66  & 0.66 & 0.62 \\

GPT-4*  & 0.72  & 0.68  & \multicolumn{1}{c|}{ 0.57 } &  0.59 &  0.52 & \multicolumn{1}{c|}{ 0.45 } & 0.84  & 0.85  & \multicolumn{1}{c|}{ 0.79 } & 0.70  &  0.66 & 0.58  \\ \hline

\hline
\bottomrule
\end{tabular}}
\end{table}

\begin{table}[ht]
\centering
\caption{5-shot, 2-shot, and 0-shot accuracy results of English version Programming Comprehension Task in the answer-only setting. C.U., C.R,
and M.H.R indicate conceptual understanding, commonsense reasoning, and multi-hop reasoning
questions, respectively.  * represents chat-bot mode of the LLM. - indicates that the model does not have the ability to complete the test in that setting.}
\label{table:shot-CoT-ZH}
\renewcommand{\arraystretch}{1.0}
\resizebox{\columnwidth}{!}{%
\begin{tabular}{l|cccccccccccc}
\toprule
\hline
 & \multicolumn{12}{c}{{Programming Comprehension (Answer-Only)-EN}} \\ \cline{2-13} 
 & \multicolumn{3}{c|}{{C.U.}} & \multicolumn{3}{c|}{{C.R.}} & \multicolumn{3}{c|}{{M.H.R.}} & \multicolumn{3}{c}{{Total}} \\ \cline{2-13} 
\multirow{-3}{*}{Model} & {{5-shot}} & {{2-shot}} & \multicolumn{1}{c|}{{{0-shot}}} & {{5-shot}} & {{2-shot}} & \multicolumn{1}{c|}{{{0-shot}}} & {{5-shot}} & {{2-shot}} & \multicolumn{1}{c|}{{{0-shot}}} & {{5-shot}} & {{2-shot}} & {{0-shot}} \\ \hline
{ChatGLM-6B} & {0.32} & {0.34} & \multicolumn{1}{c|}{{0.36}} & {0.16} & 0.21 & \multicolumn{1}{c|}{0.24} & 0.30 & 0.31 & \multicolumn{1}{c|}{0.30} & 0.29 & 0.28 & 0.30 \\
{ChatGLM-6B*} & {0.32} & {0.25} & \multicolumn{1}{c|}{{0.39}} & {0.27} & 0.24 & \multicolumn{1}{c|}{0.23} & 0.25 & 0.38 & \multicolumn{1}{c|}{0.30} & 0.32 & 0.28 & 0.23 \\
{ChatGLM2-6B} & {0.34} & {0.33} & \multicolumn{1}{c|}{{0.41}} & {0.25} & 0.26 & \multicolumn{1}{c|}{0.32} & 0.30 & 0.30 & \multicolumn{1}{c|}{0.23} & 0.29 & 0.30 & 0.30 \\
{ChatGLM2-6B*} & {0.36} & {0.32} & \multicolumn{1}{c|}{{0.28}} & {0.26} & 0.28 & \multicolumn{1}{c|}{0.23} & 0.31 & 0.28 & \multicolumn{1}{c|}{0.21} & 0.30 & 0.30 & 0.25 \\

{ChatGLM3-6B} & {0.31} & {0.31} & \multicolumn{1}{c|}{{0.37}} & {0.33} & 0.32 & \multicolumn{1}{c|}{0.30} & 0.28 & 0.30 & \multicolumn{1}{c|}{0.30} & 0.31 & 0.31 & 0.32 \\

{ChatGLM3-6B*} & {0.24} & {0.30} & \multicolumn{1}{c|}{{0.34}} & {0.23} & 0.17 & \multicolumn{1}{c|}{0.20} & 0.23 & 0.19 & \multicolumn{1}{c|}{0.30} & 0.23 & 0.22 & 0.27 \\

{MOSS-16B*} & {0.26} & {0.25} & \multicolumn{1}{c|}{{0.26}} & {0.24} & 0.27 & \multicolumn{1}{c|}{{0.35}} & 0.33 & 0.33 & \multicolumn{1}{c|}{0.15} & 0.28 & 0.28 & 0.31 \\
{Chinese-Alpaca-7B} & {0.31} & {0.30} & \multicolumn{1}{c|}{{0.28}} & {0.27} & 0.26 & \multicolumn{1}{c|}{0.25} & 0.18 & 0.21 & \multicolumn{1}{c|}{0.33} & 0.29 & 0.26 & 0.29 \\
{Chinese-Alpaca-7B*} & {0.17} & {0.26} & \multicolumn{1}{c|}{{0.29}} & {0.29} & 0.29 & \multicolumn{1}{c|}{0.19} & 0.21 & 0.16 & \multicolumn{1}{c|}{0.23} & 0.17 & 0.25 & 0.28 \\
{Chinese-Alpaca-plus-7B} & {0.34} & {0.24} & \multicolumn{1}{c|}{{0.28}} & {0.30} & 0.29 & \multicolumn{1}{c|}{0.28} & 0.14 & 0.22 & \multicolumn{1}{c|}{0.33} & 0.30 & 0.26 & 0.27 \\
Chinese-Alpaca-plus-7B* & 0.24 & 0.26 & \multicolumn{1}{c|}{0.21} & 0.29 & 0.21 & \multicolumn{1}{c|}{0.20} & 0.21 & 0.20 & \multicolumn{1}{c|}{0.23} & 0.26 & 0.23 & 0.22 \\
Chinese-Alpaca-13B & 0.26 & 0.20 & \multicolumn{1}{c|}{0.31} & 0.23 & 0.21 & \multicolumn{1}{c|}{0.23} & 0.15 & 0.16 & \multicolumn{1}{c|}{0.21} & 0.23 & 0.20 & 0.27 \\
Chinese-Alpaca-13B* & 0.10 & 0.23 & \multicolumn{1}{c|}{0.28} & 0.23 & 0.28 & \multicolumn{1}{c|}{0.21} & 0.21 & 0.25 & \multicolumn{1}{c|}{0.16} & 0.10 & 0.25 & 0.24 \\
Chinese-Alpaca-plus-13B & 0.24 & 0.21 & \multicolumn{1}{c|}{0.24} & 0.34 & {0.36} & \multicolumn{1}{c|}{0.27} & 0.21 & 0.26 & \multicolumn{1}{c|}{0.31} & 0.28 & 0.28 & 0.21 \\
Chinese-Alpaca-plus-13B* & 0.21 & 0.27 & \multicolumn{1}{c|}{0.29} & 0.27 & {0.36} & \multicolumn{1}{c|}{0.19} & 0.13 & 0.25 & \multicolumn{1}{c|}{0.22} & 0.26 & 0.30 & 0.26 \\
BELLE-7B-1M & 0.33 & 0.32 & \multicolumn{1}{c|}{0.33} & 0.32 & 0.31 & \multicolumn{1}{c|}{0.26} & 0.28 & 0.28 & \multicolumn{1}{c|}{0.23} & 0.31 & 0.31 & 0.29 \\
BELLE-7B-1M* & 0.39 & 0.36 & \multicolumn{1}{c|}{0.33} & 0.31 & 0.27 & \multicolumn{1}{c|}{0.28} & 0.21 & 0.23 & \multicolumn{1}{c|}{0.24} & 0.30 & 0.29 & 0.26 \\
BELLE-7B-2M & 0.26 & 0.24 & \multicolumn{1}{c|}{0.24} & 0.32 & 0.26 & \multicolumn{1}{c|}{0.30} & 0.28 & 0.20 & \multicolumn{1}{c|}{0.26} & 0.26 & 0.24 & 0.24 \\
BELLE-7B-2M* & 0.23 & 0.21 & \multicolumn{1}{c|}{0.27} & 0.32 & 0.23 & \multicolumn{1}{c|}{0.19} & 0.18 & 0.20 & \multicolumn{1}{c|}{0.25} & 0.23 & 0.22 & 0.24 \\
BELLE-LLaMA-7B-0.6M & 0.24 & 0.32 & \multicolumn{1}{c|}{0.29} & 0.33 & 0.23 & \multicolumn{1}{c|}{0.27} & 0.28 & 0.33 & \multicolumn{1}{c|}{0.20} & 0.26 & 0.29 & 0.21 \\
BELLE-LLaMA-7B-0.6M* & 0.27 & 0.31 & \multicolumn{1}{c|}{0.34} & 0.30 & 0.28 & \multicolumn{1}{c|}{0.27} & 0.30 & 0.31 & \multicolumn{1}{c|}{0.25} & 0.29 & 0.30 & 0.29 \\
BELLE-LLaMA-7B-2M & 0.30 & 0.28 & \multicolumn{1}{c|}{0.21} & 0.30 & 0.27 & \multicolumn{1}{c|}{0.33} & 0.16 & 0.14 & \multicolumn{1}{c|}{0.26} & 0.27 & 0.24 & 0.23 \\
BELLE-LLaMA-7B-2M* & 0.28 & 0.26 & \multicolumn{1}{c|}{0.27} & 0.21 & 0.32 & \multicolumn{1}{c|}{0.29} & 0.09 & 0.26 & \multicolumn{1}{c|}{0.25} & 0.26 & 0.28 & 0.25 \\
BELLE-LLaMA-13B-2M & 0.18 & 0.20 & \multicolumn{1}{c|}{0.22} & 0.31 & 0.22 & \multicolumn{1}{c|}{0.25} & 0.18 & 0.18 & \multicolumn{1}{c|}{0.21} & 0.24 & 0.20 & 0.28 \\
BELLE-LLaMA-13B-2M* & 0.24 & 0.27 & \multicolumn{1}{c|}{0.26} & 0.27 & 0.27 & \multicolumn{1}{c|}{0.23} & 0.22 & 0.20 & \multicolumn{1}{c|}{0.15} & 0.27 & 0.25 & 0.23 \\
InternLM-Chat-7B & 0.43 & 0.41 & \multicolumn{1}{c|}{0.44} & 0.22 & 0.33 & \multicolumn{1}{c|}{{0.36}} & {0.34} & 0.26 & \multicolumn{1}{c|}{0.31} & 0.37 & {0.34} & 0.37 \\
Baichuan-7B & 0.26 & 0.29 & \multicolumn{1}{c|}{0.27} & 0.26 & 0.27 & \multicolumn{1}{c|}{0.31} & 0.26 & 0.27 & \multicolumn{1}{c|}{0.30} & 0.22 & 0.28 & 0.31 \\
EduChat-base-002-7B* & 0.33 & 0.24 & \multicolumn{1}{c|}{0.26} & 0.31 & {0.34} & \multicolumn{1}{c|}{0.25} & 0.23 & 0.31 & \multicolumn{1}{c|}{0.33} & 0.31 & 0.30 & 0.16 \\
EduChat-base-002-13B* & 0.37 & 0.30 & \multicolumn{1}{c|}{0.21} & 0.35 & {0.36} & \multicolumn{1}{c|}{0.12} & 0.25 & 0.25 & \multicolumn{1}{c|}{0.18} & 0.33 & 0.31 & 0.10 \\
EduChat-sft-002-7B* & 0.29 & 0.26 & \multicolumn{1}{c|}{0.23} & 0.19 & {0.34} & \multicolumn{1}{c|}{0.25} & 0.25 & 0.25 & \multicolumn{1}{c|}{0.15} & 0.29 & 0.29 & 0.19 \\
CodeT5-plus-16B & 0.19 & 0.20 & \multicolumn{1}{c|}{0.21} & 0.28 & 0.30 & \multicolumn{1}{c|}{0.29} & 0.32 & 0.24 & \multicolumn{1}{c|}{0.22} & 0.25 & 0.25 & 0.26 \\
CodeT5-plus-16B* & 0.20 & 0.20 & \multicolumn{1}{c|}{-} & {0.37} & 0.27 & \multicolumn{1}{c|}{-} & 0.27 & 0.28 & \multicolumn{1}{c|}{-} & 0.24 & 0.25 & - \\
CodeT5-plus-6B & 0.28 & 0.27 & \multicolumn{1}{c|}{0.24} & 0.35 & 0.29 & \multicolumn{1}{c|}{0.19} & 0.24 & 0.29 & \multicolumn{1}{c|}{0.16} & 0.32 & 0.28 & 0.20 \\
CodeT5-plus-6B* & 0.26 & 0.27 & \multicolumn{1}{c|}{-} & 0.28 & 0.28 & \multicolumn{1}{c|}{-} & 0.26 & 0.25 & \multicolumn{1}{c|}{-} & 0.30 & 0.27 & - \\
GPT3.5-turbo & {0.51} & {0.51} & \multicolumn{1}{c|}{{0.52}} & 0.30 & 0.30 & \multicolumn{1}{c|}{{0.35}} & {0.61} & {0.55} & \multicolumn{1}{c|}{{0.51}} & {0.42} & {0.44} & {0.49} \\
GPT3.5-turbo* & {0.53} & {0.52} & \multicolumn{1}{c|}{{0.53}} & {0.36} & 0.32 & \multicolumn{1}{c|}{0.25} & {0.61} & {0.50} & \multicolumn{1}{c|}{{0.55}} & {0.44} & {0.44} & {0.44} \\ 

GPT-4   & 0.62  &  0.64 & \multicolumn{1}{c|}{ 0.63 } & 0.48  & 0.44  & \multicolumn{1}{c|}{ 0.48 } & 0.75  & 0.75  & \multicolumn{1}{c|}{ 0.80 } & 0.60  & 0.59  & 0.62  \\
GPT-4*  & 0.61  & 0.62  & \multicolumn{1}{c|}{ 0.62 } &  0.48 &  0.42 & \multicolumn{1}{c|}{ 0.48 } & 0.82  & 0.77  & \multicolumn{1}{c|}{ 0.80 } & 0.61  &  0.58 & 0.61  \\ \hline

\bottomrule
\end{tabular}}
\end{table}

\begin{table}[ht]
\centering
\caption{5-shot, 2-shot, and 0-shot accuracy results of English version Programming Comprehension Task in CoT setting.  C.U., C.R,
and M.H.R indicate conceptual understanding, commonsense reasoning, and multi-hop reasoning
questions, respectively.  * represents chat-bot mode of the LLM. - indicates that the model does not have the ability to complete the test in that setting.}
\label{table:shot-CoT-EN}
\renewcommand{\arraystretch}{1.0}
\resizebox{\columnwidth}{!}{%
\begin{tabular}{l|cccccccccccc}
\hline
 & \multicolumn{12}{c}{{Programming Comprehension(Chain-of-Thought)-EN}} \\ \cline{2-13} 
 & \multicolumn{3}{c|}{{C.U.}} & \multicolumn{3}{c|}{{C.R.}} & \multicolumn{3}{c|}{{M.H.R.}} & \multicolumn{3}{c}{{Total}} \\ \cline{2-13} 
\multirow{-3}{*}{Model} & {{5-shot}} & {{2-shot}} & \multicolumn{1}{c|}{{{0-shot}}} & {{5-shot}} & {{2-shot}} & \multicolumn{1}{c|}{{{0-shot}}} & {{5-shot}} & {{2-shot}} & \multicolumn{1}{c|}{{{0-shot}}} & {{5-shot}} & {{2-shot}} & {{0-shot}} \\ \hline
{ChatGLM-6B} & {0.31} & {0.34} & \multicolumn{1}{c|}{{0.23}} & {0.16} & 0.16 & \multicolumn{1}{c|}{0.18} & 0.25 & {0.31} & \multicolumn{1}{c|}{0.26} & 0.24 & 0.26 & 0.22 \\
{ChatGLM-6B*} & {0.28} & {0.29} & \multicolumn{1}{c|}{{0.20}} & {0.25} & 0.17 & \multicolumn{1}{c|}{0.25} & 0.31 & 0.28 & \multicolumn{1}{c|}{0.28} & 0.28 & 0.24 & 0.24 \\
{ChatGLM2-6B} & {0.37} & {0.36} & \multicolumn{1}{c|}{{0.33}} & {0.25} & 0.24 & \multicolumn{1}{c|}{0.15} & 0.23 & 0.30 & \multicolumn{1}{c|}{0.21} & 0.29 & 0.30 & 0.23 \\
{ChatGLM2-6B*} & {0.30} & {0.29} & \multicolumn{1}{c|}{{0.40}} & {0.25} & 0.23 & \multicolumn{1}{c|}{0.25} & 0.23 & 0.28 & \multicolumn{1}{c|}{0.21} & 0.26 & 0.26 & 0.30 \\

{ChatGLM3-6B} & {0.35} & {0.33} & \multicolumn{1}{c|}{{0.40}} & {0.20} & 0.30 & \multicolumn{1}{c|}{0.17} & 0.28 & 0.21 & \multicolumn{1}{c|}{0.27} & 0.27 & 0.29 & 0.27 \\

{ChatGLM3-6B*} & {0.34} & {0.28} & \multicolumn{1}{c|}{{0.31}} & {0.21} & 0.21 & \multicolumn{1}{c|}{0.20} & 0.27 & 0.19 & \multicolumn{1}{c|}{0.28} & 0.27 & 0.23 & 0.26 \\

{MOSS-16B*} & {0.13} & {0.27} & \multicolumn{1}{c|}{{0.26}} & {0.10} & 0.19 & \multicolumn{1}{c|}{0.20} & 0.28 & 0.26 & \multicolumn{1}{c|}{0.18} & 0.16 & 0.24 & 0.22 \\
{Chinese-Alpaca-7B} & {0.23} & {0.26} & \multicolumn{1}{c|}{{0.23}} & {0.33} & 0.30 & \multicolumn{1}{c|}{0.14} & 0.13 & 0.21 & \multicolumn{1}{c|}{0.13} & 0.25 & 0.26 & 0.17 \\
{Chinese-Alpaca-7B*} & {0.23} & {0.27} & \multicolumn{1}{c|}{{0.22}} & {0.30} & 0.32 & \multicolumn{1}{c|}{0.14} & 0.11 & 0.23 & \multicolumn{1}{c|}{0.10} & 0.23 & 0.28 & 0.16 \\
{Chinese-Alpaca-plus-7B} & {0.20} & {0.27} & \multicolumn{1}{c|}{{0.28}} & {0.23} & 0.30 & \multicolumn{1}{c|}{0.09} & 0.28 & 0.30 & \multicolumn{1}{c|}{0.13} & 0.23 & 0.29 & 0.17 \\
Chinese-Alpaca-plus-7B* & 0.22 & 0.26 & \multicolumn{1}{c|}{0.28} & 0.26 & 0.18 & \multicolumn{1}{c|}{0.20} & 0.26 & 0.21 & \multicolumn{1}{c|}{0.16} & 0.25 & 0.22 & 0.22 \\
Chinese-Alpaca-13B & 0.31 & 0.26 & \multicolumn{1}{c|}{0.26} & 0.23 & 0.21 & \multicolumn{1}{c|}{0.14} & 0.20 & 0.23 & \multicolumn{1}{c|}{0.20} & 0.25 & 0.23 & 0.20 \\
Chinese-Alpaca-13B* & 0.20 & 0.24 & \multicolumn{1}{c|}{0.22} & 0.21 & 0.16 & \multicolumn{1}{c|}{0.19} & 0.18 & 0.26 & \multicolumn{1}{c|}{0.21} & 0.20 & 0.22 & 0.21 \\
Chinese-Alpaca-plus-13B & 0.24 & 0.31 & \multicolumn{1}{c|}{0.24} & 0.22 & 0.19 & \multicolumn{1}{c|}{0.25} & 0.30 & 0.28 & \multicolumn{1}{c|}{0.10} & 0.25 & 0.26 & 0.21 \\
Chinese-Alpaca-plus-13B* & 0.30 & 0.30 & \multicolumn{1}{c|}{0.21} & 0.22 & 0.15 & \multicolumn{1}{c|}{0.20} & 0.25 & 0.30 & \multicolumn{1}{c|}{0.18} & 0.26 & 0.24 & 0.20 \\
BELLE-7B-1M & 0.23 & 0.17 & \multicolumn{1}{c|}{0.24} & 0.22 & 0.16 & \multicolumn{1}{c|}{0.18} & 0.16 & 0.08 & \multicolumn{1}{c|}{0.21} & 0.21 & 0.14 & 0.21 \\
BELLE-7B-1M* & 0.21 & 0.19 & \multicolumn{1}{c|}{0.12} & 0.19 & 0.20 & \multicolumn{1}{c|}{0.11} & 0.11 & 0.15 & \multicolumn{1}{c|}{0.07} & 0.18 & 0.18 & 0.10 \\
BELLE-7B-2M & 0.13 & 0.14 & \multicolumn{1}{c|}{0.26} & 0.22 & 0.19 & \multicolumn{1}{c|}{0.20} & 0.11 & 0.13 & \multicolumn{1}{c|}{0.13} & 0.16 & 0.16 & 0.20 \\
BELLE-7B-2M* & 0.19 & 0.19 & \multicolumn{1}{c|}{0.14} & 0.22 & 0.22 & \multicolumn{1}{c|}{0.08} & 0.13 & 0.07 & \multicolumn{1}{c|}{0.07} & 0.19 & 0.17 & 0.10 \\
BELLE-LLaMA-7B-0.6M & 0.22 & 0.20 & \multicolumn{1}{c|}{0.08} & 0.22 & 0.23 & \multicolumn{1}{c|}{0.13} & 0.26 & 0.18 & \multicolumn{1}{c|}{0.03} & 0.23 & 0.21 & 0.09 \\
BELLE-LLaMA-7B-0.6M* & 0.33 & 0.22 & \multicolumn{1}{c|}{0.20} & 0.19 & 0.21 & \multicolumn{1}{c|}{0.24} & 0.26 & 0.21 & \multicolumn{1}{c|}{0.25} & 0.26 & 0.22 & 0.23 \\
BELLE-LLaMA-7B-2M & 0.22 & 0.07 & \multicolumn{1}{c|}{0.09} & 0.21 & 0.08 & \multicolumn{1}{c|}{0.10} & 0.10 & 0.05 & \multicolumn{1}{c|}{0.13} & 0.19 & 0.07 & 0.10 \\
BELLE-LLaMA-7B-2M* & 0.21 & 0.11 & \multicolumn{1}{c|}{0.21} & 0.20 & 0.11 & \multicolumn{1}{c|}{0.15} & 0.13 & 0.03 & \multicolumn{1}{c|}{0.20} & 0.19 & 0.09 & 0.18 \\
BELLE-LLaMA-13B-2M & 0.26 & 0.21 & \multicolumn{1}{c|}{0.11} & 0.21 & 0.22 & \multicolumn{1}{c|}{0.13} & 0.15 & 0.13 & \multicolumn{1}{c|}{0.20} & 0.21 & 0.20 & 0.14 \\
BELLE-LLaMA-13B-2M* & 0.28 & 0.21 & \multicolumn{1}{c|}{0.16} & 0.17 & 0.10 & \multicolumn{1}{c|}{0.09} & 0.18 & 0.11 & \multicolumn{1}{c|}{0.10} & 0.21 & 0.14 & 0.12 \\
InternLM-Chat-7B & 0.42 & 0.32 & \multicolumn{1}{c|}{0.37} & 0.25 & 0.20 & \multicolumn{1}{c|}{0.21} & 0.26 & 0.25 & \multicolumn{1}{c|}{0.25} & 0.32 & 0.26 & 0.28 \\
Baichuan-7B & 0.12 & 0.10 & \multicolumn{1}{c|}{0.08} & 0.09 & 0.11 & \multicolumn{1}{c|}{0.02} & 0.08 & 0.02 & \multicolumn{1}{c|}{0.30} & 0.10 & 0.08 & 0.04 \\
EduChat-base-002-7B* & 0.24 & 0.26 & \multicolumn{1}{c|}{0.13} & 0.19 & 0.24 & \multicolumn{1}{c|}{0.12} & 0.26 & 0.20 & \multicolumn{1}{c|}{0.15} & 0.23 & 0.24 & 0.13 \\
EduChat-base-002-13B* & 0.31 & 0.24 & \multicolumn{1}{c|}{0.12} & 0.24 & 0.22 & \multicolumn{1}{c|}{0.12} & 0.25 & 0.28 & \multicolumn{1}{c|}{0.13} & 0.27 & 0.24 & 0.12 \\
EduChat-sft-002-7B* & 0.29 & 0.31 & \multicolumn{1}{c|}{0.24} & 0.22 & 0.20 & \multicolumn{1}{c|}{0.17} & 0.26 & 0.25 & \multicolumn{1}{c|}{0.15} & 0.26 & 0.25 & 0.19 \\
CodeT5-plus-16B & - & - & \multicolumn{1}{c|}{-} & - & - & \multicolumn{1}{c|}{-} & - & - & \multicolumn{1}{c|}{-} & - & - & - \\
CodeT5-plus-16B* & - & - & \multicolumn{1}{c|}{-} & - & - & \multicolumn{1}{c|}{-} & - & - & \multicolumn{1}{c|}{-} & - & - & - \\
CodeT5-plus-6B & - & - & \multicolumn{1}{c|}{-} & - & - & \multicolumn{1}{c|}{-} & - & - & \multicolumn{1}{c|}{-} & - & - & - \\
CodeT5-plus-6B* & - & - & \multicolumn{1}{c|}{-} & - & - & \multicolumn{1}{c|}{-} & - & - & \multicolumn{1}{c|}{-} & - & - & - \\
GPT3.5-turbo & {0.44} & {0.49} & \multicolumn{1}{c|}{{0.53}} & {0.34} & {0.35} & \multicolumn{1}{c|}{{0.37}} & {0.64} & {0.61} & \multicolumn{1}{c|}{{0.57}} & {0.45} & {0.46} & {0.47} \\
GPT3.5-turbo* & {0.51} & {0.50} & \multicolumn{1}{c|}{{0.47}} & {0.37} & {0.34} & \multicolumn{1}{c|}{{0.33}} & {0.63} & {0.61} & \multicolumn{1}{c|}{{0.50}} & {0.49} & {0.47} & {0.42} \\ 

GPT-4   & 0.60   & 0.64   & \multicolumn{1}{c|}{ 0.62  } & 0.51   & 0.51   & \multicolumn{1}{c|}{ 0.51 } & 0.87  & 0.82  & \multicolumn{1}{c|}{ 0.75 } & 0.63  & 0.63 & 0.61 \\

GPT-4*  & 0.61  & 0.63  & \multicolumn{1}{c|}{ 0.66 } &  0.52 &  0.52 & \multicolumn{1}{c|}{ 0.49 } & 0.84  & 0.79  & \multicolumn{1}{c|}{ 0.75 } & 0.63  &  0.62 & 0.62  \\

\hline
\bottomrule
\end{tabular}}
\end{table}

\section{Breakdown Performance in Code Generation}\label{app:cg}
Table \ref{tab:codeClass1} and Table \ref{tab:codeClass2} present the breakdown of performance of LLMs in code generation task in the code-only and code-with-context scenarios. 
The breakdown performance of explicit questions and narrative questions is also shown in the table.

\begin{table}[t]
\centering
\renewcommand{\arraystretch}{1.0}
\caption{Best performance comparison between explicit questions and narrative questions on AC@1, AC@all and AC Rate metric of code generation in Chinese version.}
\label{tab:codeClass1}
\resizebox{\columnwidth}{!}{%
\begin{tabular}{l|ccc|ccc|ccc|ccc}
\toprule\hline
\multicolumn{1}{c|}{}   & \multicolumn{12}{c}{Code Generation-ZH}       \\ \cline{2-13} 
\multicolumn{1}{c|}{}   & \multicolumn{6}{c|}{Code-only}   & \multicolumn{6}{c}{Code-with-Context}                                       \\ \cline{2-13} 
\multicolumn{1}{c|}{}   & \multicolumn{3}{c|}{Explicit Questions}   & \multicolumn{3}{c|}{Narrative Questions}& \multicolumn{3}{c|}{Explicit Questions}   & \multicolumn{3}{c}{Narrative Questions}                                       \\ \cline{2-13} 
\multicolumn{1}{c|}{\multirow{-3}{*}{Model}} &  AC@1 &  AC@all & \multicolumn{1}{c|}{AC Rate} & AC@1 &  AC@all & AC Rate &  AC@1 &  AC@all & \multicolumn{1}{c|}{AC Rate} & AC@1 &  AC@all & AC Rate \\ \hline 
\multicolumn{1}{l|}{baichuan}  & 0.42 & 0.26 & \multicolumn{1}{c|}{0.29} & 0.18 & 0.18 & \multicolumn{1}{c|}{0.18} & 0.24 & 0.16 & \multicolumn{1}{c|}{0.16}  & 0.09 & 0.09 & \multicolumn{1}{c}{0.09} \\
\multicolumn{1}{l|}{BELLE} & 0.27 & 0.15 & \multicolumn{1}{c|}{0.16} & 0.10 & 0.10 & \multicolumn{1}{c|}{0.10} & 0.27 & 0.14 & \multicolumn{1}{c|}{0.15}  & 0.12 & 0.12 & \multicolumn{1}{c}{0.12}  \\
\multicolumn{1}{l|}{BELLE-7B-1M}   & 0.20 & 0.09 & \multicolumn{1}{c|}{0.10} & 0.12 & 0.12 & \multicolumn{1}{c|}{0.12} & 0.19 & 0.10 & \multicolumn{1}{c|}{0.11}  & 0.10 & 0.10 & \multicolumn{1}{c}{0.10}  \\
\multicolumn{1}{l|}{chatglm} & 0.27 & 0.14 & \multicolumn{1}{c|}{0.17} & 0.15 & 0.15 & \multicolumn{1}{c|}{0.15} & 0.21 & 0.11 & \multicolumn{1}{c|}{0.12}  & 0.10 & 0.10 & \multicolumn{1}{c}{0.10}  \\
\multicolumn{1}{l|}{chatglm2} & 0.29 & 0.17 & \multicolumn{1}{c|}{0.20} & 0.12 & 0.12 & \multicolumn{1}{c|}{0.12} & 0.27 & 0.12 & \multicolumn{1}{c|}{0.20}  & 0.12 & 0.12 & \multicolumn{1}{c}{0.12} \\
\multicolumn{1}{l|}{cn-alpaca-plus-13B} & 0.36 & 0.20 & \multicolumn{1}{c|}{0.22} & 0.19 & 0.19 & \multicolumn{1}{c|}{0.19} & 0.37 & 0.21 & \multicolumn{1}{c|}{0.24}  & 0.16 & 0.16 & \multicolumn{1}{c}{0.16} \\
\multicolumn{1}{l|}{gpt3.5-turbo}    & 0.70 & 0.48 & \multicolumn{1}{c|}{0.57} & 0.58 & 0.58 & \multicolumn{1}{c|}{0.58} & 0.72 & 0.52 & \multicolumn{1}{c|}{0.60}  & 0.61 & 0.61 & \multicolumn{1}{c}{0.61}  \\

\multicolumn{1}{l|}{gpt-4}    & 0.66 & 0.53 & \multicolumn{1}{c|}{0.57} & 0.57 & 0.46 & \multicolumn{1}{c|}{0.51} & 0.77 & 0.64 & \multicolumn{1}{c|}{0.67}  & 0.77 & 0.62 & \multicolumn{1}{c}{0.66}  \\

\multicolumn{1}{l|}{intern}   & 0.19 & 0.11 & \multicolumn{1}{c|}{0.11} & 0.09 & 0.09 & \multicolumn{1}{c|}{0.09} & 0.21 & 0.14 & \multicolumn{1}{c|}{0.13}  & 0.08 & 0.08 & \multicolumn{1}{c}{0.08}  \\

\multicolumn{1}{l|}{moss} & 0.34 & 0.22 & \multicolumn{1}{c|}{0.23} & 0.11 & 0.11 & \multicolumn{1}{c|}{0.11} & 0.34 & 0.20 & \multicolumn{1}{c|}{0.24}  & 0.14 & 0.14 & \multicolumn{1}{c}{0.14}  \\

\multicolumn{1}{l|}{starcoder}   & 0.28 & 0.18 & \multicolumn{1}{c|}{0.21} & 0.16 & 0.16 & \multicolumn{1}{c|}{0.16} & 0.23 & 0.12 & \multicolumn{1}{c|}{0.15}  & 0.15 & 0.15 & \multicolumn{1}{c}{0.15}  \\ 
\multicolumn{1}{l|}{vicuna}   & 0.35 & 0.18 & \multicolumn{1}{c|}{0.23} & 0.17 & 0.17 & \multicolumn{1}{c|}{0.17} & 0.37 & 0.21 & \multicolumn{1}{c|}{0.26}  & 0.23 & 0.23 & \multicolumn{1}{c}{0.23}  \\

\multicolumn{1}{l|}{wizard}   & 0.54 & 0.35 & \multicolumn{1}{c|}{0.37} & 0.34 & 0.34 & \multicolumn{1}{c|}{0.34} & 0.53 & 0.36 & \multicolumn{1}{c|}{0.39}  & 0.32 & 0.32 & \multicolumn{1}{c}{0.32}  \\

\hline\bottomrule
\end{tabular}}
\end{table}

\begin{table}[t]
\centering
\renewcommand{\arraystretch}{1.0}
\caption{Best performance comparison between explicit questions and narrative questions on AC@1, AC@all and AC Rate metric of code generation in English version.}
\label{tab:codeClass2}
\resizebox{\columnwidth}{!}{%
\begin{tabular}{l|ccc|ccc|ccc|ccc}
\toprule\hline
\multicolumn{1}{c|}{}   & \multicolumn{12}{c}{Code Generation-EN}       \\ \cline{2-13} 
\multicolumn{1}{c|}{}   & \multicolumn{6}{c|}{Code-only}   & \multicolumn{6}{c}{Code-with-Context}                                       \\ \cline{2-13} 
\multicolumn{1}{c|}{}   & \multicolumn{3}{c|}{Explicit Questions}   & \multicolumn{3}{c|}{Narrative Questions}& \multicolumn{3}{c|}{Explicit Questions}   & \multicolumn{3}{c}{Narrative Questions}                                       \\ \cline{2-13} 
\multicolumn{1}{c|}{\multirow{-3}{*}{Model}} &  AC@1 &  AC@all & \multicolumn{1}{c|}{AC Rate} & AC@1 &  AC@all & AC Rate &  AC@1 &  AC@all & \multicolumn{1}{c|}{AC Rate} & AC@1 &  AC@all & AC Rate \\ \hline 
\multicolumn{1}{l|}{baichuan}  & 0.41 & 0.23 & \multicolumn{1}{c|}{0.27} & 0.24 & 0.24 & \multicolumn{1}{c|}{0.24} & 0.40 & 0.25 & \multicolumn{1}{c|}{0.25}  & 0.25 & 0.25 & \multicolumn{1}{c}{0.25} \\
\multicolumn{1}{l|}{BELLE} & 0.28 & 0.13 & \multicolumn{1}{c|}{0.16} & 0.08 & 0.08 & \multicolumn{1}{c|}{0.08} & 0.29 & 0.14 & \multicolumn{1}{c|}{0.19}  & 0.14 & 0.14 & \multicolumn{1}{c}{0.14}  \\
\multicolumn{1}{l|}{BELLE-7B-1M}   & 0.27 & 0.09 & \multicolumn{1}{c|}{0.13} & 0.13 & 0.13 & \multicolumn{1}{c|}{0.13} & 0.19 & 0.08 & \multicolumn{1}{c|}{0.10}  & 0.12 & 0.12 & \multicolumn{1}{c}{0.12}  \\
\multicolumn{1}{l|}{chatglm} & 0.26 & 0.10 & \multicolumn{1}{c|}{0.12} & 0.16 & 0.16 & \multicolumn{1}{c|}{0.16} & 0.22 & 0.09 & \multicolumn{1}{c|}{0.11}  & 0.14 & 0.14 & \multicolumn{1}{c}{0.14}  \\
\multicolumn{1}{l|}{chatglm2} & 0.23 & 0.12 & \multicolumn{1}{c|}{0.12} & 0.14 & 0.14 & \multicolumn{1}{c|}{0.14} & 0.14 & 0.07 & \multicolumn{1}{c|}{0.07}  & 0.06 & 0.06 & \multicolumn{1}{c}{0.06} \\
\multicolumn{1}{l|}{cn-alpaca-plus-13B} & 0.39 & 0.21 & \multicolumn{1}{c|}{0.24} & 0.17 & 0.17 & \multicolumn{1}{c|}{0.17} & 0.32 & 0.17 & \multicolumn{1}{c|}{0.20}  & 0.18 & 0.18 & \multicolumn{1}{c}{0.18} \\
\multicolumn{1}{l|}{gpt3.5-turbo}    & 0.69 & 0.50 & \multicolumn{1}{c|}{0.56} & 0.53 & 0.53 & \multicolumn{1}{c|}{0.53} & 0.71 & 0.53 & \multicolumn{1}{c|}{0.58}  & 0.62 & 0.62 & \multicolumn{1}{c}{0.62}  \\

\multicolumn{1}{l|}{gpt-4}    & 0.60 & 0.47 & \multicolumn{1}{c|}{0.52} & 0.56 & 0.45 & \multicolumn{1}{c|}{0.51} & 0.73 & 0.58 & \multicolumn{1}{c|}{0.60}  & 0.67 & 0.56 & \multicolumn{1}{c}{0.60}  \\

\multicolumn{1}{l|}{intern}   & 0.14 & 0.09 & \multicolumn{1}{c|}{0.08} & 0.04 & 0.04 & \multicolumn{1}{c|}{0.04} & 0.40 & 0.23 & \multicolumn{1}{c|}{0.25}  & 0.19 & 0.19 & \multicolumn{1}{c}{0.19}  \\

\multicolumn{1}{l|}{moss} & 0.35 & 0.24 & \multicolumn{1}{c|}{0.26} & 0.16 & 0.16 & \multicolumn{1}{c|}{0.16} & 0.22 & 0.13 & \multicolumn{1}{c|}{0.15}  & 0.08 & 0.08 & \multicolumn{1}{c}{0.08}  \\

\multicolumn{1}{l|}{starcoder}   & 0.43 & 0.26 & \multicolumn{1}{c|}{0.27} & 0.24 & 0.24 & \multicolumn{1}{c|}{0.24} & 0.35 & 0.23 & \multicolumn{1}{c|}{0.24}  & 0.19 & 0.19 & \multicolumn{1}{c}{0.19}  \\ 
\multicolumn{1}{l|}{vicuna}   & 0.25 & 0.12 & \multicolumn{1}{c|}{0.14} & 0.11 & 0.11 & \multicolumn{1}{c|}{0.11} & 0.40 & 0.18 & \multicolumn{1}{c|}{0.24}  & 0.19 & 0.19 & \multicolumn{1}{c}{0.19}  \\

\multicolumn{1}{l|}{wizard}   & 0.54 & 0.34 & \multicolumn{1}{c|}{0.39} & 0.29 & 0.29 & \multicolumn{1}{c|}{0.29} & 0.52 & 0.35 & \multicolumn{1}{c|}{0.39}  & 0.34 & 0.34 & \multicolumn{1}{c}{0.34}  \\

\hline\bottomrule
\end{tabular}}
\end{table}

\section{Breakdown Performance in Code Correction}\label{app:CC}
Table \ref{tab:CC-ZH-wa}, \ref{tab:CC-ZH-CPU_T},  \ref{tab:CC-ZH-re}, \ref{tab:CC-EN-wa}, \ref{tab:CC-EN-CPU_T}, and \ref{tab:CC-EN-re} present the breakdown of performance of LLMs in code correction task with wrong answer error, time limited error and runtime error.
The breakdown performance of code-only, code-with-error message and code-with-standard code scenarios are also shown in the table.

\begin{table}[t]
\centering
\caption{Performance on AC@1, AC@all, and AC Rate metrics of code correction task (Wrong Answer Error) in Chinese version. }
\label{tab:CC-ZH-wa}
\renewcommand{\arraystretch}{1.0}
\resizebox{\columnwidth}{!}{%
\begin{tabular}{l|ccc|ccc|ccc}
\toprule
\hline
\multicolumn{1}{c|}{}   & \multicolumn{9}{c}{Code Correction (Wrong Answer)-ZH}\\ \cline{2-10} 
\multicolumn{1}{c|}{}   & \multicolumn{3}{c|}{Code-only}  & \multicolumn{3}{c|}{Code-with-Error Message}   & \multicolumn{3}{l}{Code-with-Standard Code}                                                                                                 \\ \cline{2-10} 
\multicolumn{1}{c|}{\multirow{-3}{*}{Model}} & { AC@1} & { AC@n} & \multicolumn{1}{c|}{{ AC Rate}} & { AC@1} & { AC@n} & \multicolumn{1}{c|}{AC Rate} & \multicolumn{1}{c}{{ { AC@1}}} & \multicolumn{1}{c}{{ { AC@n}}} & \multicolumn{1}{c}{AC Rate} \\ \hline
ChatGLM-6B              & 0.1362    & 0.0511    & 0.0906    & 0.1243    & 0.0468    & 0.0814    & 0.1736    & 0.1591    &  0.1148  \\
ChatGLM2-6B             & 0.2493    & 0.0791    & 0.1482    & 0.2315    & 0.0851    & 0.1361   & 0.2119 		  & 0.1711 	  &  0.1525  \\
ChatGLM3-6B             & 0.4323    & 0.2298    & 0.2826    & 0.4468    & 0.2230    & 0.2798    & 0.2917    & 0.2805    & 0.2220  \\
BELLE-7B-1M             & 0.0553    & 0.0221    & 0.0346    & 0.0272    & 0.0051    & 0.0155    & 0.0894    & 0.0519    & 0.0742   \\
BELLE-LLaMA-13B-2M      & 0.0655    & 0.0196    & 0.0379    & 0.0494    & 0.0136    & 0.0253    & 0.0391    & 0.0357    & 0.0219    \\
InternLM-Chat-7B        & 0.1234    & 0.0706    & 0.0832    & 0.1362    & 0.0681    & 0.0859    & 0.1498    & 0.1362    & 0.1371   \\
Baichuan-Chat-13B       & 0.3643    & 0.2477    & 0.2837    & 0.3728    & 0.2443    & 0.2649    & 0.2400    & 0.2102    & 0.1890 \\
WizardCoder-15B         & 0.5864    & 0.4264    & 0.4487    & 0.5626    & 0.4034    & 0.4387    & 0.2511    & 0.2383	& 0.2000   \\
StarCoder               & 0.6843    & 0.4162    & 0.4934    & 0.6621 	& 0.3821 	& 0.4691    & 0.4757    & 0.3047    & 0.3447   \\
GPT3.5-turbo            & 0.5643    & 0.5396    & 0.4818    & 0.5668    & 0.5405    & 0.4932    & 0.3404    & 0.3268    & 0.2916   \\ 

GPT4 & 0.6774	& 0.6417    & 0.6287    & 0.6996    & 0.6612    & 0.6546    & 0.5838    & 0.5728	& 0.5152  \\ \hline\bottomrule
\end{tabular}
}
\end{table}

\begin{table}[t]
\centering
\caption{Performance on AC@1, AC@all, and AC Rate metrics of code correction task (Wrong Answer Error) in English version. }
\label{tab:CC-EN-wa}
\renewcommand{\arraystretch}{1.0}
\resizebox{\columnwidth}{!}{%
\begin{tabular}{l|ccc|ccc|ccc}
\toprule
\hline
\multicolumn{1}{c|}{}   & \multicolumn{9}{c}{Code Correction (Wrong Answer)-EN}\\ \cline{2-10} 
\multicolumn{1}{c|}{}   & \multicolumn{3}{c|}{Code-only}  & \multicolumn{3}{c|}{Code-with-Error Message}   & \multicolumn{3}{l}{Code-with-Standard Code}                                                                                                 \\ \cline{2-10} 
\multicolumn{1}{c|}{\multirow{-3}{*}{Model}} & { AC@1} & { AC@n} & \multicolumn{1}{c|}{{ AC Rate}} & { AC@1} & { AC@n} & \multicolumn{1}{c|}{AC Rate} & \multicolumn{1}{c}{{ { AC@1}}} & \multicolumn{1}{c}{{ { AC@n}}} & \multicolumn{1}{c}{AC Rate} \\ \hline
ChatGLM-6B              & 0.1506    & 0.0562    & 0.0950    & 0.0136    & 0.0034    & 0.0093    & 0.1174    & 0.1038    & 0.1078  \\
ChatGLM2-6B             & 0.2332    & 0.0791    & 0.1196    & 0.1872    & 0.0765    & 0.1018    & 0.2647    & 0.2213 	  & 0.2011  \\
ChatGLM3-6B             & 0.3745    & 0.2187    & 0.2726    &0.3123    & 0.1583    & 0.2161    & 0.2145    & 0.2111    & 0.1622  \\
BELLE-7B-1M             & 0.1591    & 0.0468    & 0.0903    & 0.0026 	  & 0.0009 	  & 0.0008    & 0.1260    & 0.1089    & 0.0968   \\
BELLE-LLaMA-13B-2M      & 0.1123    & 0.0553    & 0.0785    & 0.2204    & 0.0553    & 0.1300    & 0.1660    & 0.1404    & 0.1161    \\
InternLM-Chat-7B        & 0.1864    & 0.1038    & 0.1269    & 0.1209    & 0.0604    & 0.0768    & 0.2570    & 0.2383    & 0.2010   \\
Baichuan-Chat-13B       & 0.2289    & 0.1634    & 0.1844    & 0.0970    & 0.0553    & 0.0587    & 0.1864    & 0.1821    & 0.1451 \\
WizardCoder-15B         & 0.4553    & 0.3277    & 0.3241    & 0.3081    & 0.2417    & 0.2535    & 0.1498    & 0.1311	& 0.1018   \\
StarCoder               & 0.6136    & 0.3617    & 0.4349    & 0.5770    & 0.3183    & 0.4079    & 0.3566    & 0.2774    & 0.2660   \\
GPT3.5-turbo            & 0.4945    & 0.4783    & 0.4171    & 0.3609    & 0.3430    & 0.3144    & 0.2536    & 0.2511    & 0.2020   \\ 
GPT4 & 0.6247	& 0.5957    & 0.5581    & 0.6460    & 0.6170    & 0.5837    & 0.4664    & 0.4596	& 0.3949  \\ \hline\bottomrule
\end{tabular}
}
\end{table}

\begin{table}[t]
\centering
\caption{Performance on AC@1, AC@all, and AC Rate metrics of code correction task (Time Limit
Exceeded Error) in Chinese version. }
\label{tab:CC-ZH-CPU_T}
\renewcommand{\arraystretch}{1.0}
\resizebox{\columnwidth}{!}{%
\begin{tabular}{l|ccc|ccc|ccc}
\toprule
\hline
\multicolumn{1}{c|}{}   & \multicolumn{9}{c}{Code Correction (CPU Timeout)-ZH}\\ \cline{2-10} 
\multicolumn{1}{c|}{}   & \multicolumn{3}{c|}{Code-only}  & \multicolumn{3}{c|}{Code-with-Error Message}   & \multicolumn{3}{l}{Code-with-Standard Code}                                                                                                 \\ \cline{2-10} 
\multicolumn{1}{c|}{\multirow{-3}{*}{Model}} & { AC@1} & { AC@n} & \multicolumn{1}{c|}{{ AC Rate}} & { AC@1} & { AC@n} & \multicolumn{1}{c|}{AC Rate} & \multicolumn{1}{c}{{ { AC@1}}} & \multicolumn{1}{c}{{ { AC@n}}} & \multicolumn{1}{c}{AC Rate} \\ \hline
ChatGLM-6B              & 0.1159    & 0.0870    & 0.1359    & 0.1739    & 0.1304    & 0.1988    & 0.2609 	  & 0.2609    & 0.0913  \\
ChatGLM2-6B             & 0.3623    & 0.1884    & 0.3854    & 0.3043    & 0.1449    & 0.2941    & 0.0580    & 0.0580 	  & 0.0487  \\
ChatGLM3-6B             & 0.4348    & 0.3623    & 0.5030    & 0.4203    & 0.3623    & 0.5010    & 0.3768    & 0.3768    & 0.1136 \\
BELLE-7B-1M             & 0.0580    & 0.0435    & 0.0730    & 0.0145    & 0.0145 	  & 0.0203    & 0.1159    & 0.0290    & 0.1055   \\
BELLE-LLaMA-13B-2M      & 0.0870    & 0.0580    & 0.0913    & 0.0870    & 0.0580    & 0.1034   & 0.0000    & 0.00000    & 0.0000    \\

InternLM-Chat-7B        & 0.0725    & 0.0580    & 0.0345    & 0.1014    & 0.0870    & 0.1116    & 0.1594    & 0.1594    & 0.0649   \\
Baichuan-Chat-13B       & 0.4783    & 0.3768    & 0.4665    & 0.5362    & 0.4348    & 0.5639    & 0.3188    & 0.2899    & 0.1075 \\
WizardCoder-15B         & 0.4203    & 0.4058    & 0.5071    & 0.4493    & 0.4058    & 0.5497    & 0.3188    & 0.3043 	& 0.0933   \\
StarCoder               & 0.4783    & 0.2319    & 0.4361    & 0.4638    & 0.2173    & 0.4260    & 0.4783    & 0.1739    & 0.3327   \\
GPT3.5-turbo            & 0.4058    & 0.3768    & 0.3692    & 0.4493    & 0.4348    & 0.4077    & 0.3913    & 0.3913    & 0.2252   \\ 
GPT4 & 0.8261	& 0.7246    & 0.7667    &  0.8116	& 0.7391    & 0.7565    & 0.7391    & 0.7391	& 0.5761  \\ \hline\bottomrule
\end{tabular}
}
\end{table}

\begin{table}[t]
\centering
\caption{Performance on AC@1, AC@all, and AC Rate metrics of code correction task (Time Limit
Exceeded Error) in English version. }
\label{tab:CC-EN-CPU_T}
\renewcommand{\arraystretch}{1.0}
\resizebox{\columnwidth}{!}{%
\begin{tabular}{l|ccc|ccc|ccc}
\toprule
\hline
\multicolumn{1}{c|}{}   & \multicolumn{9}{c}{Code Correction(CPU Timeout)-EN}\\ \cline{2-10} 
\multicolumn{1}{c|}{}   & \multicolumn{3}{c|}{Code-only}  & \multicolumn{3}{c|}{Code-with-Error Message}   & \multicolumn{3}{l}{Code-with-Standard Code}                                                                                                 \\ \cline{2-10} 
\multicolumn{1}{c|}{\multirow{-3}{*}{Model}} & { AC@1} & { AC@n} & \multicolumn{1}{c|}{{ AC Rate}} & { AC@1} & { AC@n} & \multicolumn{1}{c|}{AC Rate} & \multicolumn{1}{c}{{ { AC@1}}} & \multicolumn{1}{c}{{ { AC@n}}} & \multicolumn{1}{c}{AC Rate} \\ \hline
ChatGLM-6B              & 0.1304    & 0.0870    & 0.1379    & 0.0000    & 0.0000    & 0.0000    & 0.2319    & 0.2319    & 0.0730  \\
ChatGLM2-6B             & 0.2609    & 0.1159    & 0.2840    & 0.0870    & 0.0290    & 0.0446    & 0.0580    & 0.0435 	 & 0.0406  \\
ChatGLM3-6B             & 0.3188    & 0.2464    & 0.3773    & 0.2319    & 0.1884    & 0.2515    & 0.1304    & 0.1304    & 0.0933  \\
BELLE-7B-1M             & 0.1739    & 0.0725 	& 0.1318    & 0.0000    & 0.0000    & 0.0000    & 0.0290    & 0.0290    & 0.0406   \\
BELLE-LLaMA-13B-2M      & 0.1449    & 0.1014    & 0.1643    & 0.2754    & 0.1159    & 0.2738    & 0.2464    & 0.2464    & 0.0771   \\
InternLM-Chat-7B        & 0.1304    & 0.1159    & 0.1318    & 0.0435    & 0.0290    & 0.0730    & 0.3623    & 0.3623    & 0.1318   \\
Baichuan-Chat-13B       & 0.3913    & 0.2754    & 0.4118    & 0.1159    & 0.0870    & 0.0994    & 0.1884    & 0.1884    & 0.0325 \\
WizardCoder-15B         & 0.3913    & 0.3333    & 0.5051    & 0.1304    & 0.1304    & 0.1846    & 0.2319    & 0.2319	& 0.0913   \\
StarCoder               & 0.4493    & 0.2174    & 0.4178    & 0.4058    & 0.1304    & 0.3448    & 0.5217    & 0.2899    & 0.3124   \\
GPT3.5-turbo            & 0.4638    & 0.4348    & 0.4178    & 0.4058    & 0.3768    & 0.2860    & 0.3623    & 0.3623    & 0.0933   \\ 
GPT4 & 0.7391	& 0.6667    & 0.6815    & 0.7101    & 0.6377    & 0.7079    & 0.5072    & 0.5072	& 0.3144  \\ \hline\bottomrule
\end{tabular}
}
\end{table}

\begin{table}[t]
\centering
\caption{Performance on AC@1, AC@all, and AC Rate metrics of code correction task (Runtime Error) in Chinese version. }
\label{tab:CC-ZH-re}
\renewcommand{\arraystretch}{1.0}
\resizebox{\columnwidth}{!}{%
\begin{tabular}{l|ccc|ccc|ccc}
\toprule
\hline
\multicolumn{1}{c|}{}   & \multicolumn{9}{c}{Code Correction(Runtime Error)-ZH}\\ \cline{2-10} 
\multicolumn{1}{c|}{}   & \multicolumn{3}{c|}{Code-only}  & \multicolumn{3}{c|}{Code-with-Error Message}   & \multicolumn{3}{l}{Code-with-Standard Code}                                                                                                 \\ \cline{2-10} 
\multicolumn{1}{c|}{\multirow{-3}{*}{Model}} & { AC@1} & { AC@n} & \multicolumn{1}{c|}{{ AC Rate}} & { AC@1} & { AC@n} & \multicolumn{1}{c|}{AC Rate} & \multicolumn{1}{c}{{ { AC@1}}} & \multicolumn{1}{c}{{ { AC@n}}} & \multicolumn{1}{c}{AC Rate} \\ \hline
ChatGLM-6B              & 0.0930    & 0.0347    & 0.0477    & 0.1512    & 0.0465    & 0.0682    & 0.1279    & 0.1163    & 0.0941  \\
ChatGLM2-6B             & 0.2442    & 0.1628    & 0.1487    & 0.1744    & 0.0930    & 0.1010    & 0.1860    & 0.1744    & 0.1378  \\
ChatGLM3-6B             & 0.3837    & 0.3372    & 0.3383    & 0.3837    & 0.3372    & 0.3383    & 0.2907    & 0.2791 	  & 0.2278  \\
BELLE-7B-1M             & 0.0116    & 0.0116    & 0.0068    & 0.0116 	  & 0.0116 	  & 0.0136    & 0.0465    & 0.0348    & 0.0218   \\
BELLE-LLaMA-13B-2M      & 0.0349    & 0.0233    & 0.0150    & 0.0233    & 0.0233    & 0.0136    & 0.1279    & 0.1279    & 0.0750    \\

InternLM-Chat-7B        & 0.0349    & 0.0349    & 0.0341    & 0.0233    & 0.0233    & 0.0205    & 0.1046    & 0.0814    & 0.1214   \\
Baichuan-Chat-13B       & 0.2907    & 0.2326    & 0.2483    & 0.3256    & 0.2674    & 0.2606    & 0.1395    & 0.1163    & 0.0928 \\
WizardCoder-15B         & 0.5698    & 0.5465    & 0.5498    & 0.5465    & 0.5349    & 0.5416    & 0.2674    & 0.2558    & 0.2087   \\
StarCoder               & 0.5349    & 0.4535    & 0.4188    & 0.5116    & 0.4302    & 0.4052    & 0.4419    & 0.3837    & 0.3615   \\
GPT3.5-turbo            & 0.7791    & 0.7791    & 0.7162    & 0.7442    & 0.7442    & 0.6958    & 0.3488    & 0.3488 	& 0.3220   \\ 
GPT4 & 0.8605	& 0.8488    & 0.8240    & 0.8372    & 0.8372    & 0.8131    & 0.7674    & 0.7674    & 0.6767  \\ \hline\bottomrule
\end{tabular}
}
\end{table}

\begin{table}[t]
\centering
\caption{Performance on AC@1, AC@all, and AC Rate metrics of code correction task (Runtime Error) in English version. }
\label{tab:CC-EN-re}
\renewcommand{\arraystretch}{1.0}
\resizebox{\columnwidth}{!}{%
\begin{tabular}{l|ccc|ccc|ccc}
\toprule
\hline
\multicolumn{1}{c|}{}   & \multicolumn{9}{c}{Code Correction(Runtime Error)-EN}\\ \cline{2-10} 
\multicolumn{1}{c|}{}   & \multicolumn{3}{c|}{Code-only}  & \multicolumn{3}{c|}{Code-with-Error Message}   & \multicolumn{3}{l}{Code-with-Standard Code}                                                                                                 \\ \cline{2-10} 
\multicolumn{1}{c|}{\multirow{-3}{*}{Model}} & { AC@1} & { AC@n} & \multicolumn{1}{c|}{{ AC Rate}} & { AC@1} & { AC@n} & \multicolumn{1}{c|}{AC Rate} & \multicolumn{1}{c}{{ { AC@1}}} & \multicolumn{1}{c}{{ { AC@n}}} & \multicolumn{1}{c}{AC Rate} \\ \hline
ChatGLM-6B              & 0.1395    & 0.0581    & 0.0655    & 0.0116    & 0.0000    & 0.0068    & 0.1279    & 0.1163    & 0.0941  \\
ChatGLM2-6B             & 0.1977    & 0.1395    & 0.1610    & 0.2093    & 0.1628    & 0.1487    & 0.2093    & 0.2093 	  & 0.1910  \\
ChatGLM3-6B             & 0.3372    & 0.2907    & 0.2783    & 0.2209    & 0.1744    & 0.1951    & 0.2209    & 0.2209    & 0.1910  \\
BELLE-7B-1M             & 0.1744    & 0.1512    & 0.1596    & 0.0000 	  & 0.0000 	  & 0.0000    & 0.0465    & 0.0465    & 0.0409   \\
BELLE-LLaMA-13B-2M      & 0.0581    & 0.0465    & 0.0382    & 0.0930    & 0.0465    & 0.0450    & 0.1511    & 0.1511    & 0.1091    \\
InternLM-Chat-7B        & 0.2093    & 0.2093    & 0.1569    & 0.0930    & 0.0930    & 0.0614    & 0.2674    & 0.2209    & 0.1924   \\
Baichuan-Chat-13B       & 0.2907    & 0.2326    & 0.2374    & 0.0581    & 0.0233    & 0.0231    & 0.1512    & 0.1512    & 0.1296 \\
WizardCoder-15B         & 0.4302    & 0.4070    & 0.3861    & 0.1977    & 0.1744    & 0.1405    & 0.1977    & 0.1744	& 0.1692   \\
StarCoder               & 0.5233    & 0.4535    & 0.4052    & 0.4302    & 0.3605    & 0.3424    & 0.2791 	& 0.2791    & 0.2933   \\
GPT3.5-turbo            & 0.7326    & 0.7326    & 0.6958    & 0.3605    & 0.3605    & 0.3629    & 0.2907    & 0.2906    & 0.2469   \\ 

GPT4 & 0.8256	& 0.8256    & 0.8117    & 0.8372    & 0.8372    & 0.8213    & 0.6628    & 0.6628    & 0.5935  \\ \hline\bottomrule
\end{tabular}
}
\end{table}